\documentclass{article}

\usepackage{amsmath,amsfonts,bm}

\def\eqref#1{equation~\ref{#1}}

\def\1{\bm{1}}

\DeclareMathAlphabet{\mathsfit}{\encodingdefault}{\sfdefault}{m}{sl}
\SetMathAlphabet{\mathsfit}{bold}{\encodingdefault}{\sfdefault}{bx}{n}

\usepackage{microtype}
\usepackage{graphicx}
\usepackage{subcaption}
\usepackage{booktabs} 
\DeclareMathOperator{\Quant}{Quantile} 

\usepackage{hyperref}

\usepackage[preprint]{icml2026}

\usepackage{amsmath}
\usepackage{amssymb}
\usepackage{mathtools}
\usepackage{amsthm}

\usepackage[dvipsnames]{xcolor}

\definecolor{mycolor}{RGB}{240, 247, 252}
\usepackage{siunitx}
\usepackage{multicol}
\usepackage{colortbl}
\usepackage{tabularx}
\usepackage{makecell}
\usepackage{multirow}
\usepackage{diagbox}
\usepackage{wrapfig}
\usepackage{enumitem}
\usepackage[normalem]{ulem}
\usepackage{tgcursor}
\usepackage{bbm}
\usepackage{longtable}
\usepackage{xcolor}
\renewcommand{\algorithmiccomment}[1]{\hfill$\triangleright$~#1}
\usepackage[most]{tcolorbox}
\usepackage{fontawesome5}

\usepackage[capitalize,noabbrev]{cleveref}

\theoremstyle{plain}

\theoremstyle{definition}

\theoremstyle{remark}

\usepackage[textsize=tiny]{todonotes}

\usepackage{marvosym} 
\definecolor{authorcolor}{RGB}{61, 146, 81}

\icmlshowauthorstrue
\makeatletter
\gdef\icmlcorrespondingauthor@text{}
\makeatother

\newcommand{\customauthorblock}{%
    \small
  {\bf
    Shaobo Wang
    $^{\text{\Letter}}$$^*$$^{{\color{authorcolor}\boldsymbol{e,a}}}$
    Jiaming Wang
    $^*$$^{{\color{authorcolor}\boldsymbol{e,n}}}$
    Jiajun Zhang
    $^*$$^{{\color{authorcolor}\boldsymbol{e,b}}}$
    Cong Wang
    $^{{\color{authorcolor}\boldsymbol{e}}}$    
    Yue Min
    $^{{\color{authorcolor}\boldsymbol{e,h}}}$
    Zichen Wen
    $^{{\color{authorcolor}\boldsymbol{e}}}$
    \vspace{3pt}
  } \\
    \small
  {\bf \hspace{-10pt}
  Xingzhang Ren
    $^{{\color{authorcolor}\boldsymbol{a}}}$
    Fei Huang
    $^{{\color{authorcolor}\boldsymbol{a}}}$
    Huiqiang Jiang
    $^{{\color{authorcolor}\boldsymbol{a}}}$
    Junyang Lin
    $^{{\color{authorcolor}\boldsymbol{a}}}$
    Dayiheng Liu
    $^{{\color{authorcolor}\boldsymbol{a}}}$
    Linfeng Zhang
    $^{\text{\Letter}}$$^{{\color{authorcolor}\boldsymbol{e}}}$
    \vspace{5pt}
  } \\
    \small
    {
    $^{\color{authorcolor}\boldsymbol{e}}$ EPIC Lab, SJTU \quad
    $^{\color{authorcolor}\boldsymbol{a}}$ Qwen Team, Alibaba Group \quad 
    $^{\color{authorcolor}\boldsymbol{n}}$ NJU \quad
    $^{\color{authorcolor}\boldsymbol{b}}$ BJTU \quad
    $^{\color{authorcolor}\boldsymbol{h}}$ HKUST
    \vspace{2pt}
    } \\
    \small
    {
    * Equal contribution $\quad$ 
    \Letter\ Corresponding authors  $\quad$
    \faGithub\ \href{https://github.com/gszfwsb/Q-tuning}{Code} $\quad$
    \Mundus\ \href{https://gszfwsb.github.io/Q-tuning/}{Project Page}
    }%
}

\providecommand{\danqi}[1]{
    {\protect\color{purple}{}}
}
\providecommand{\alex}[1]{
    {\protect\color{blue}{}}
}
\providecommand{\aatmik}[1]{
    {\protect\color{orange}{}}
}
\providecommand{\saumya}[1]{
    {\protect\color{orange}{}}
}

\renewcommand{\paragraph}[1]{\vspace{0.2cm}\noindent\textbf{#1}}

\newtcbox{\hlprimarytab}{on line, rounded corners, box align=base, colback=green!10,colframe=white,size=fbox,arc=3pt, before upper=\strut, top=-2pt, bottom=-4pt, left=-2pt, right=-2pt, boxrule=0pt}
\newtcbox{\hlsecondarytab}{on line, box align=base, colback=red!10,colframe=white,size=fbox,arc=3pt, before upper=\strut, top=-2pt, bottom=-4pt, left=-2pt, right=-2pt, boxrule=0pt}
\newcommand{\dashifted}{\raisebox{0.5\depth}{\tiny$\downarrow$}}
\newcommand{\uashifted}{\raisebox{0.5\depth}{\tiny$\uparrow$}}

\newcommand{\ua}{\raisebox{0.2ex}{\hlprimarytab{$\uparrow$}}}
\newcommand{\da}{\raisebox{0.2ex}{\hlsecondarytab{$\downarrow$}}}

\newcommand{\dar}[1]{{\raisebox{0.6ex}{\tiny\hlsecondarytab{\dashifted{#1}}}}}
\newcommand{\uar}[1]{{\raisebox{0.6ex}{\tiny\hlprimarytab{\uashifted{#1}}}}}
\usepackage{wrapfig}
\usepackage{booktabs}  
\usepackage[most]{tcolorbox}

\icmltitlerunning{Winning the Pruning Gamble: A Unified Approach to Joint Sample and Token
Pruning for Efficient Supervised Fine-Tuning}

\begin{document}

\twocolumn[
  \icmltitle{Winning the Pruning Gamble: A Unified Approach to Joint Sample and Token Pruning for Efficient Supervised Fine-Tuning}



  \begin{center}
    \customauthorblock
  \end{center}

  \icmlkeywords{Machine Learning, ICML}

  \vskip 0.3in
]



\printAffiliationsAndNotice{}  
\label{sec:intro}

\begin{abstract}
As supervised fine-tuning (SFT) evolves from a lightweight post-training step into a compute-intensive phase rivaling mid-training in scale, data efficiency has become critical for aligning large language models (LLMs) under tight budgets. Existing data pruning methods suffer from a fragmented design: they operate either at the sample level or the token level in isolation, failing to jointly optimize both dimensions. This disconnect leads to significant inefficiencies—high-value samples may still contain redundant tokens, while token-level pruning often discards crucial instructional or corrective signals embedded in individual examples. To address this bottleneck, we introduce the \textit{Error–Uncertainty (EU) Plane}, a diagnostic framework that jointly characterizes the heterogeneous utility of training data across samples and tokens. Guided by this insight, we propose \textit{Quadrant-based Tuning (Q-Tuning)}, a unified framework that strategically coordinates sample pruning and token pruning with strong empirical robustness. Q-Tuning employs a two-stage strategy: first, it performs sample-level triage to retain examples rich in informative misconceptions or calibration signals; second, it applies an asymmetric token-pruning policy, using a context-aware scoring mechanism to trim less salient tokens exclusively from misconception samples while preserving calibration samples in their entirety. Our method consistently empirically sets a new state of the art across five diverse benchmarks. Remarkably, on SmolLM2-1.7B, Q-Tuning achieves a +38\% average improvement over the full-data SFT baseline using only 12.5\% of the original training data. 
\end{abstract}

\begin{figure}[tb!]
    \centering
    \vspace{-5pt}
\includegraphics[width=0.99\linewidth]{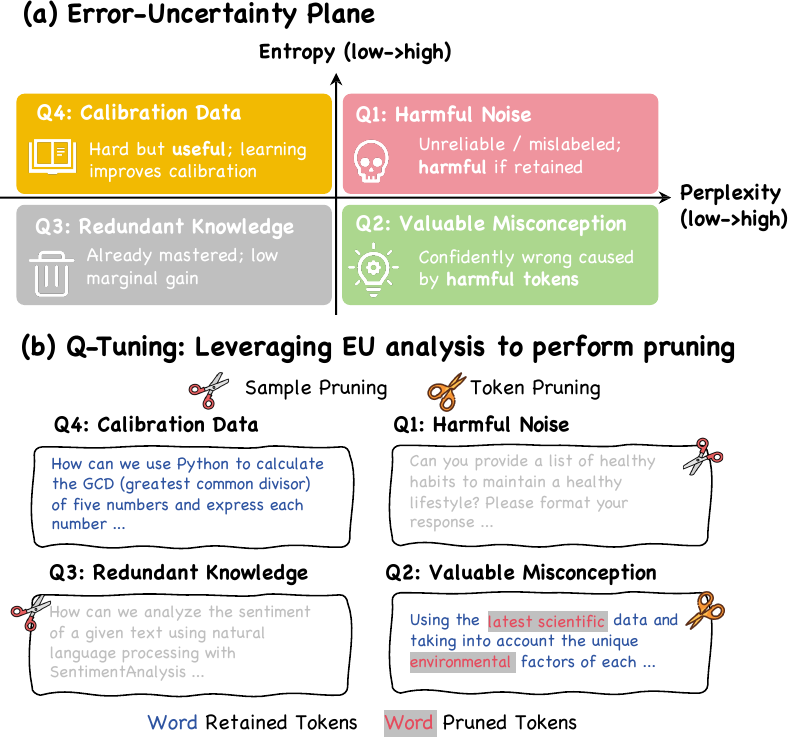}
    \vspace{-5pt}
\caption{
\textbf{(a) Error--Uncertainty (EU) plane.} We partition samples by perplexity and entropy into four regions: Q1 (harmful noise), Q2 (valuable misconceptions), Q3 (redundant knowledge), and Q4 (calibration data).
\textbf{(b) Q-Tuning.} Q-Tuning performs joint pruning guided by the EU plane: it drops Q1 and Q3, selectively prunes tokens in Q2, and retains Q4 in full.
}
    \label{fig:EUplane}
    \vspace{-15pt}
\end{figure}
\section{Introduction}

The explosive growth of alignment datasets—now routinely spanning billions of tokens—has fundamentally transformed Supervised Fine-Tuning (SFT) from a lightweight post-training step into a compute-intensive phase rivaling mid-training in scale~\citep{ouyang2022training,dong2024abilities,yang2025qwen3,achiam2023gpt,team2023gemini}. In this new regime, the primary challenge is no longer simply reducing data volume, but maximizing the utility of every retained token and sample—a task that demands accurate, on-the-fly estimation of data value. Yet despite the emphasis on data efficiency, recent work reveals a troubling paradox: even sophisticated dynamic pruning heuristics often underperform \textit{simple random sampling}~\citep{xia2024rethinking}. This starkly exposes a fundamental disconnect between current strategies and the true utility of alignment data.

At the heart of this challenge lies the fragmented design of existing pruning strategies. \textit{Sample-level pruning} methods~\citep{qin2024infobatch, zhou2023lima, wang2025datawhisperer, yang2024smalltolarge} identify high-potential examples but treat all tokens within them as equally valuable—retaining redundant or even harmful content that dilutes alignment signals. Conversely, \textit{token-level pruning} approaches~\citep{lin2024not, xia2025tokenskip, chen2024image, zhang2024sparsevlm} apply context-agnostic heuristics uniformly across the dataset, blind to the semantic role of each sample. Such a one-size-fits-all policy fails to differentiate, for instance, between a sample containing a correctable misconception—where only specific tokens need refinement—and one serving as a calibration anchor, which must be preserved holistically to maintain model stability. By operating in isolation, neither paradigm captures the interdependent nature of sample and token utility. This raises a central question: \textit{How can we dynamically coordinate sample selection and token pruning within a unified framework to maximize the true learning utility of limited data?}

To address this, we first formalize the problem as \textbf{Generalized Dynamic Data Pruning}, a bilevel optimization framework for jointly optimizing sample and token pruning. We then introduce a novel diagnostic lens, the \textbf{Error-Uncertainty (EU) Plane}, which categorizes training instances by mapping model error (perplexity) against model uncertainty (entropy). 
Specifically, as shown in Figure~\ref{fig:EUplane}(a), training samples are categorized into four quadrants based on perplexity (higher indicating more wrong) and entropy (higher indicating more uncertain):
(i) \textit{Q1 (Harmful Noise)} — unreliable or mislabeled data that actively harms learning and should therefore be removed via sample-level pruning;
(ii) \textit{Q2 (Valuable Misconception)} — confidently wrong responses that, when pruned surgically at the token level, can be transformed into powerful teaching signals, making them ideal candidates for token-level pruning;
(iii) \textit{Q3 (Redundant Knowledge)} — mastered content offering diminishing returns, best eliminated through sample-level pruning to improve efficiency without sacrificing performance; and
(iv) \textit{Q4 (Calibration Data)} — hard but reliable samples essential for improving model confidence and robustness, which should be preserved in full — neither sample nor token pruning should be applied.

This insight motivates our solution: \textbf{Quadrant-based Tuning (Q-Tuning)}, the first principled, integrated method for dynamic data pruning. Guided by EU Plane analysis, Q-Tuning implements a two-stage, context-aware strategy, as shown in Figure~\ref{fig:EUplane}(b). First, at the sample level, it acts as an intelligent triage system: retaining samples that offer clear signals for error correction or calibration (Q2 and Q4), while discarding those classified as harmful noise or redundant knowledge (Q1 and Q3). Crucially, Q-Tuning then applies an asymmetric token policy: for confidently wrong samples (valuable misconceptions), it performs token pruning to isolate the core misconception and amplify the learning signal; for uncertain but correct samples (calibration data), it preserves full token sequences to ensure robust uncertainty modeling. Our contributions are as follows:
\begin{enumerate}[leftmargin=*,topsep=-0.5pt,itemsep=-1ex]
\item We formulate the joint sample--token pruning problem as \textbf{Generalized Dynamic Data Pruning}, a bilevel optimization framework for hybrid pruning.
\item We introduce the \textbf{Error-Uncertainty (EU) Plane}, a tool that reveals and quantifies the heterogeneous value of data across error and uncertainty dimensions. Based on the EU analysis, we propose \textbf{Q-Tuning}, the first integrated, diagnosis-driven algorithm for dynamic pruning that coordinates sample and token decisions.
\item Experiments demonstrate that Q-Tuning exceeds full-data training and all existing pruning baselines across 4 models, 5 benchmarks, and 6 different kinds of data budgets. 
Particularly, with LLaMA3-8B on GSM8K, Q-Tuning reaches 48.07 using only 35\% of the data, outperforming full-data training by 6.0 points and the strongest baseline by 9.9 points.


\end{enumerate}

\section{Generalized Dynamic Data Pruning: A Unified Framework}
\label{sec:problem_formulation}

We first propose \textit{Generalized Dynamic Data Pruning}, a framework for accelerating model training where samples and their constituent tokens are selectively and adaptively omitted at each step, without sacrificing performance. 

Specifically, the framework first considers the coarse-grained \textit{sample level pruning}, which involves identifying and discarding examples from a mini-batch deemed less informative for the model's state. Subsequently, \textit{token-level pruning} operates on the retained samples, performing a finer-grained selection to keep a critical subset of tokens within each. This two-stage process is inherently \textit{dynamic}: pruning criteria can be re-evaluated for each new mini-batch, allowing the data distribution used for gradient updates to evolve alongside the model's parameters $\theta$. The overarching objective is to focus computation on a ``doubly-pruned'' data subset to maximize training efficiency while preserving or enhancing the model's generalization.

To formalize this framework, consider a model $f_\theta$ with parameters $\theta$. At each training step $t$, a mini-batch $\mathcal{B}_t$ is drawn from the training distribution $\mathcal{D}$. The hierarchical pruning process can be modeled by two abstract operators: a sample-level pruner $\Psi$ and a token-level pruner $\Phi$.

\paragraph{Stage 1: Sample-Level Pruning.}
The operator $\Psi$ splits mini-batch $\mathcal{B}_t$ into kept and discarded samples, governed by keep ratio $r_{\text{sample}} \in [0,1]$. The retained set, $\mathcal{B}'_t$, is defined~as:
\begin{equation}
   \mathcal{B}'_t = \mathcal{B}_t \setminus \Psi(\mathcal{B}_t) \subseteq \mathcal{B}_t,
\end{equation}
where $\Psi(\mathcal{B}_t)$ denotes the set of discarded samples, and the size of the retained set $|\mathcal{B}'_t|=\lfloor r_{\text{sample}} \cdot |\mathcal{B}_t| \rfloor$.

\paragraph{Stage 2: Token-Level Pruning.}
For each sample in the retained set $\mathcal{B}'_t$, the operator $\Phi$ determines which tokens to keep, guided by a token keep ratio $r_{\text{token}} \in [0, 1]$. This is modeled by generating a binary mask $m(x) \in \{0,1\}^{L(x)}$ for each sample $x \in \mathcal{B}'_t$, where $L(x)$ is its sequence length. The final, doubly-pruned mini-batch $\tilde{\mathcal{B}}_t$ is constructed by applying these masks:
\begin{equation}
   \tilde{\mathcal{B}}_t = \Phi(\mathcal{B}'_t) = \{m(x)\odot x \mid x\in \mathcal{B}'_t\},
\end{equation}
where $\odot$ denotes element-wise product, and each mask satisfies $\|m(x)\|_1=\lfloor r_{\text{token}}L(x)\rfloor$.

\noindent \textbf{Generalized Dynamic Data Pruning}. We now put all things together into a unified framework. The central problem of this framework is to identify the optimal dynamic pruning operators, $(\Phi, \Psi)$, that guide the training process to a final model $\theta^*$ with the best possible generalization performance. These operators are applied at each step $t$ to transform a mini-batch $\mathcal{B}_t$ into a computationally cheaper, pruned version $\tilde{\mathcal{B}}_t$, while adhering to predefined keep ratios.

This problem is naturally captured as a \textit{bi-level optimization problem}. The outer loop seeks optimal pruners, while the inner loop represents the iterative training procedure that produces the final model parameters under the guidance of these pruners. Formally, the objective is as follows:
\begin{equation} \label{eq:training_revised}
\begin{gathered}
    \min_{\Phi, \Psi} \quad  \mathbb{E}_{(x, y) \sim \mathcal{D}_{\text{test}}} \left[ \mathcal{L}_{\text{test}}(f_{\theta^*}(x), y) \right] \\
    \text{s.t.} \quad  
    \theta^* = \mathop{\arg\min}_{\theta} \sum_{t=1}^{T} 
    \mathbb{E}_{\mathcal{B}_t \sim \mathcal{D}} 
    \mathbb{E}_{(x,y)\sim \tilde{\mathcal{B}}_t} 
    \left[ \mathcal{L}_{\text{train}}(f_{\theta_t}(x),y) \right]
\end{gathered}
\end{equation}
\vspace{-10pt}

where at each step $t$: $\tilde{\mathcal{B}}_t = \Phi(\mathcal{B}_t \setminus \Psi(\mathcal{B}_t))$.

Specifically, the \textit{outer objective} defines the quality metric for any pair of pruners $(\Phi, \Psi)$: the final test performance of the model they produce. The \textit{inner objective} defines the training process itself, where the final parameters $\theta^*$ result from cumulatively minimizing the loss over a sequence of dynamically pruned mini-batches. All existing methods that instantiates these operators $\Phi$ and $\Psi$ can be seen as a specific solution to this alignment problem.

\begin{figure*}[tb!]
    \centering
    \vspace{-5pt}
    \includegraphics[width=0.99\textwidth]{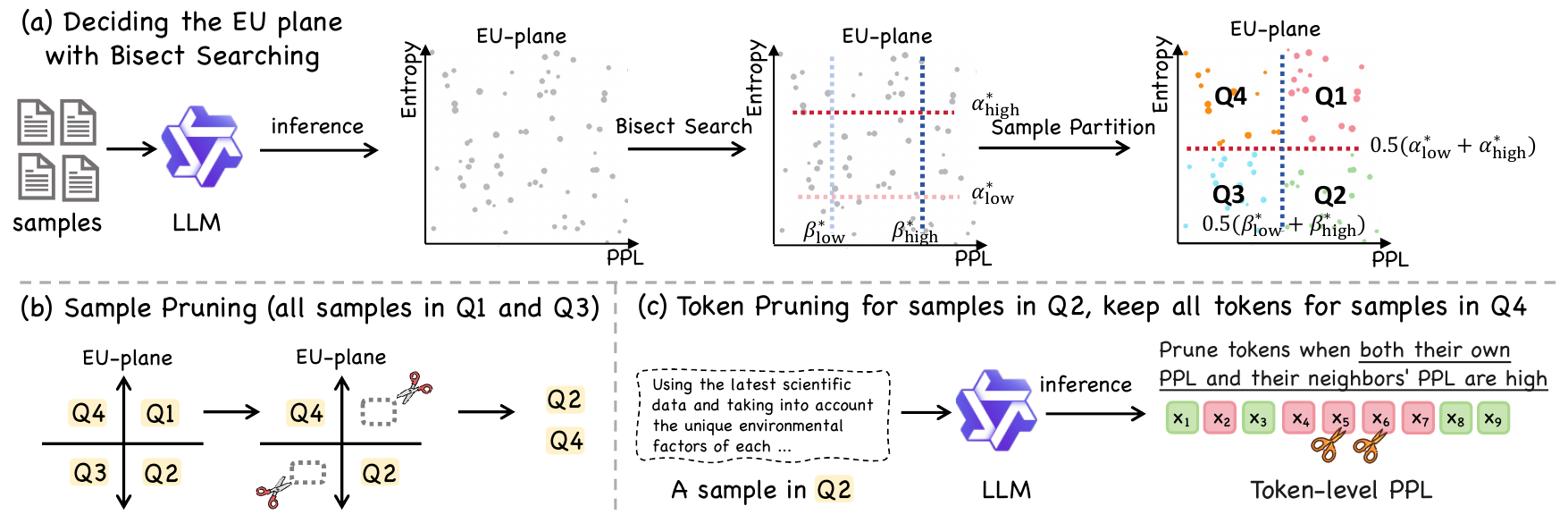} 
    \vspace{-5pt}
    \caption{
    \textbf{(a) Constructing the Error-Uncertainty (EU) Plane via Bisect Search.} We run the base LLM to compute sample-level perplexity (PPL) and entropy, and use bisect search to set thresholds $(\alpha^*_{\text{low}}, \alpha^*_{\text{high}}, \beta^*_{\text{low}}, \beta^*_{\text{high}})$ that partition the EU plane into Q1--Q4.
    \textbf{(b) Sample Pruning for Q1 and Q3.} Samples in Q1 and Q3 are pruned at the sample level, while those in Q2 and Q4 are retained.  
    \textbf{(c) Token-Level Pruning for Q2 Samples.} For retained Q2 samples, token-level pruning is applied based on both the token’s own perplexity and the average perplexity of its neighboring tokens. Tokens with high local PPL are removed, preserving only the most informative ones.
    }
    \label{fig:pipeline}
    \vspace{-10pt}
\end{figure*}

\section{Winning the Pruning Gamble}
\label{sec:method}

Building on the previous analysis, we now introduce \textbf{Q-Tuning}, a dynamic pruning method guided by the Error–Uncertainty (EU) Plane. As shown in Figure~\ref{fig:EUplane}, Q-Tuning proceeds in two coordinated stages: it first prunes harmful noise and redundant knowledge at the sample level, and then applies targeted token-level pruning to valuable misconceptions while preserving calibration data in full. This integrated strategy avoids the failure modes of one-dimensional heuristics and transforms pruning from a gamble into a principled, performance-enhancing process.

\subsection{The Error-Uncertainty Plane: A Diagnostic Lens}
\label{subsec:eu_plane}

We formalize the \textbf{Error–Uncertainty (EU) Plane} introduced in Figure~\ref{fig:EUplane}(a). Each training sample is positioned by two orthogonal axes: \emph{error}, quantified by perplexity (PPL), and \emph{uncertainty}, quantified by predictive entropy (Ent).

\paragraph{Measure Error via Perplexity.}
For a training sample $(x, y)$ with sequence length $L(x)$, let $T(x) \subseteq \{1, \dots, L(x)\}$ denote the set of trainable token positions. We define sample-level error as the perplexity ($\mathrm{PPL}$), derived from the average token-level negative log-likelihood:
\begin{equation}
\mathrm{PPL}(x,y;f_\theta) 
= \exp\  ( \sum_{i \in T(x)} \frac{-\log p\!\left(y_i \mid x, y_{<i}; f_\theta\right)}{|T(x)|} )
\end{equation}

High perplexity indicates the model finds the ground-truth continuation highly surprising, which is a sign of either genuine difficulty or misconception.

\paragraph{Measure Uncertainty via Predictive Entropy.}
Independent of correctness, we quantify the model’s uncertainty using entropy ($\mathrm{Ent}$), computed as the average token-level entropy over trainable positions, where $v \in \mathcal{V}$ denotes a vocabulary token:
\begin{equation}
\mathrm{Ent}(x,y;f_\theta)=
\frac{\sum_{i\in T(x)} H\!\left(p_\theta(\cdot\mid x,y_{<i})\right)}{|T(x)|}.
\end{equation}

High entropy indicates that the model spreads probability mass broadly, reflecting persistent uncertainty even when the top prediction is correct. Taken together, $\mathrm{PPL}$ and $\mathrm{Ent}$ map each sample onto the EU Plane, providing a principled and interpretable basis for partitioning data into the four quadrants illustrated in Figure~\ref{fig:EUplane}(b).

\subsection{Q-Tuning: Generalized Dynamic Data Pruning}
\label{subsec:q_tuning}

Q-Tuning operationalizes the insights of the EU Plane into a two-stage, per-batch dynamic pruning strategy, as illustrated in Figure~\ref{fig:pipeline}. In the first stage, bisect search determines quantile-based thresholds that partition samples into four quadrants, enabling sample-level pruning of uninformative data. In the second stage, token-level pruning is selectively applied within retained misconception samples to isolate informative signals, while calibration samples are preserved in full. The pseudocode of our method is illustrated in Algorithm~\ref{alg:qtuning} in Appendix~\ref{app:pseudo-code}.

\paragraph{Stage 1: Constructing the EU Plane via Binary Search and Pruning Samples Accordingly.}
At each training step, we compute the perplexity ($\mathrm{PPL}$) and entropy ($\mathrm{Ent}$) for every sample $x \in \mathcal{B}_t$ using a gradient-free forward pass of the current model $f_{\theta_t}$. These statistics map each sample to a point on the EU Plane (Figure~\ref{fig:pipeline}(a)).  

Our objective is to determine quantile thresholds $(\alpha^*, \beta^*)$ on the PPL and Entropy axes such that the retained proportion of samples in $Q_2 \cup Q_4$ exactly matches the target sample retention ratio $r_{\text{sample}}$. Here $\Quant_\gamma(X)$ denotes the $\gamma$-quantile of a variable $X$, \emph{i.e.}, the smallest value $q$ such that at least a fraction $\gamma$ of samples satisfy $X \leq q$.  

To locate $(\alpha^*,\beta^*)$, we perform a \textbf{bisect search} on both axes. 
For each axis, the search interval is initialized as $[0,0.5]$. At each iteration we set
$\alpha = \tfrac{1}{2}(\alpha_{\text{low}} + \alpha_{\text{high}}), 
\beta = \tfrac{1}{2}(\beta_{\text{low}} + \beta_{\text{high}})$, and compute the proportion of samples that would fall into the tentative $Q_2 \cup Q_4$ region defined by the current thresholds $(\alpha,\beta)$. 
The intervals are then updated as
\vspace{-5pt}
{\small
\begin{equation}
\begin{aligned}
&(\alpha_{\text{low}}, \alpha_{\text{high}}, \beta_{\text{low}}, \beta_{\text{high}}) \\
&=
\begin{cases}
(\alpha_{\text{low}}, \alpha, \beta_{\text{low}}, \beta), & \text{if $\gamma$} > r_{\text{sample}}, \\[6pt]
(\alpha, \alpha_{\text{high}}, \beta, \beta_{\text{high}}), & \text{otherwise}.
\end{cases}
\end{aligned}
\end{equation}}
The search converges in $O(\log(1/\varepsilon))$ iterations (typically $<10$), incurring negligible overhead. The resulting thresholds $(\alpha,\beta)$ partition the EU Plane (Figure~\ref{fig:pipeline}(b)), from which Q2 and Q4 samples are retained and Q1 and Q3 discarded.

\paragraph{Stage 2: Token-Level Pruning for Confident Errors.}
While Stage~1 removes entire samples deemed unhelpful, not all retained examples are equally homogeneous inside. In particular, samples in Q2 (Valuable Misconceptions) often contain a mix of informative context and locally harmful tokens that mislead the model. To extract the useful signal, we apply token-level pruning that discards only the most detrimental tokens while preserving the surrounding context. In contrast, samples in Q4 (Calibration Data) are challenging yet reliable, and every token contributes to improving model calibration. Therefore, Q4 sequences are kept intact without any token pruning.

For a retained sample $x \in Q_2$ with target sequence $y$, let $T(x)$ denote the set of trainable token positions. 
For each token $i \in T(x)$(Figure~\ref{fig:pipeline}(c)), we compute its token-level perplexity $\mathrm{PPL}_i$, 
which measures how surprising the ground-truth token $y_i$ is to the model.  

To avoid pruning rare but meaningful tokens based on isolated spikes, we compute a \textbf{smoothed importance score} that incorporates local context:
\begin{equation}
\begin{aligned}
s_i(x,y;f_\theta)
&= (1-\lambda)\,\mathrm{PPL}_i(x,y;f_\theta)+ \\
&{\lambda}[\mathrm{PPL}_{i-1}(x,y;f_\theta)+\mathrm{PPL}_{i+1}(x,y;f_\theta)\big],
\label{eq:token_score}
\end{aligned}
\end{equation}

where $\lambda \in [0,1]$ balances the contribution of neighboring tokens (default $\lambda=0.5$). 
This smoothing reduces the risk of mistakenly removing single high-PPL tokens that may still be semantically critical, 
and we analyze the sensitivity of this choice in the ablation study (Section~\ref{ablation:lambda}).

A token is deemed detrimental if both its own $\mathrm{PPL}_i$ and the average $\mathrm{PPL}$ of its immediate neighbors exceed a percentile-based threshold (\emph{e.g.}, the median). 
All tokens in $T(x)$ are subsequently ranked by their smoothed scores $s_i$, 
and only the top-$r_{\textrm{token}}$ fraction are retained to construct a binary mask $m(x)$. 
This mask selectively removes locally noisy tokens while preserving the informative context essential for learning. 
By contrast, no token-level pruning is applied to Q4 samples, as each token therein contributes to reliable calibration and must be preserved in full.

\begin{table*}[tb!]
\centering
\caption{Evaluation on WizardLM with a sample ratio of 12.5\% and a token ratio of 50\%, where \ua~and \da~respectively denote improvements or degradations over the \textit{Random-Random} baseline. Additional results under more sample and token pruning ratios are provided in the Table~\ref{tab:wizard_long3} and Table~\ref{tab:wizard_long2} in Appendix~\ref{appendix:add_instruction} and results on Qwen3-8B trained on the OpenHermes are reported in Appendix~\ref{appendix:instruction_qwen}.}
\vspace{-5pt}
\label{tab:wizard_long}
\resizebox{0.99\textwidth}{!}{%
\begin{tabular}{c|l|cccccc|cccccc}
\toprule 
\multicolumn{1}{c|}{Sample} & \multicolumn{1}{c|}{Token} & \multicolumn{6}{c|}{\textbf{LLaMA2-7B}} 
& \multicolumn{6}{c}{\textbf{Mistral-7B}} \\
\multicolumn{1}{c|}{Pruner} & \multicolumn{1}{c|}{Pruner} & ARC-E & ARC-C & GSM8K & SQuAD & TriviaQA & Avg.
  & ARC-E & ARC-C & GSM8K & SQuAD & TriviaQA & Avg. \\ 
\midrule

\multicolumn{2}{c|}{Zero-Shot} 
  & 53.44 & 38.98 & 5.31 & 12.18 & 43.00 & 30.58 & 66.67 & 46.10 & 18.35 & 10.01 & 43.77 & 36.98 \\
\midrule

\rowcolor{gray!5}
\multicolumn{14}{c}{\textit{$12.5\%$ Samples, $50\%$ Tokens}} \\ \midrule
\multirow{4}{*}{Random}    & Random     & 59.25 & 41.02         & 8.11 & 12.75 & 48.75 & 33.98         & 70.55 & 48.14 & 22.74 & 19.57 & 52.63 & 42.73 \\  
                           & PPL        & 60.49\uar{1.24} & 43.39\uar{2.37} & 7.20\dar{0.91} & 12.20\dar{0.55} & 48.04\dar{0.71} & 34.26\uar{0.28} & 70.72\uar{0.17} & 48.47\uar{0.33} & 25.78\uar{3.04} & 21.36\uar{1.79} & 53.92\uar{1.29} & 44.05\uar{1.32} \\
                           & FastV      & 59.96\uar{0.71} & 42.37\uar{1.35} & 5.76\dar{2.35} & 11.31\dar{1.44} & 46.42\dar{2.33} & 33.17\dar{0.81} & 70.72\uar{0.17} & 46.44\dar{1.70} & 18.80\dar{3.94} & 19.14\dar{0.43} & 51.56\dar{1.07} & 41.33\dar{1.40} \\
                           & SparseVLM  & 54.32\dar{4.93} & 37.97\dar{3.05} & 7.35\dar{0.76} & 12.76\uar{0.01} & 44.65\dar{4.10} & 31.41\dar{2.57} & 67.02\dar{3.53} & 44.75\dar{3.39} & 20.24\dar{2.50} & 10.97\dar{8.60} & 44.61\dar{8.02} & 37.52\dar{5.21} \\
                           \midrule
\multirow{4}{*}{Longest}   & Random     & 59.96\uar{0.71} & \textbf{44.41\uar{3.39}} & 7.51\dar{0.60} & 15.34\uar{2.59} & 48.91\uar{0.16} & 35.22\uar{1.24} & 74.25\uar{3.70} & 48.81\uar{0.67} & 28.73\uar{5.99} & 17.66\dar{1.91} & 55.73\uar{3.10} & 45.04\uar{2.31} \\ 
                           & PPL        & 61.19\uar{1.94} & 43.73\uar{2.71} & 6.82\dar{1.29} & 16.33\uar{3.58} & 48.16\dar{0.59} & 35.24\uar{1.26} & \textbf{75.49\uar{4.94}} & 50.17\uar{2.03} & 27.98\uar{5.24} & 21.49\uar{1.92} & 56.55\uar{3.92} & 46.33\uar{3.60} \\
                           & FastV      & 59.25\uar{0.00} & 43.05\uar{2.03} & 5.69\dar{2.42} & 13.64\uar{0.89} & 46.98\dar{1.77} & 33.72\dar{0.26} & 74.43\uar{3.88} & 49.15\uar{1.01} & 25.70\uar{2.96} & 22.89\uar{3.32} & 54.15\uar{1.52} & 45.26\uar{2.53} \\
                           & SparseVLM  & 54.32\dar{4.93} & 38.31\dar{2.71} & 7.13\dar{0.98} & 10.92\dar{1.83} & 43.77\dar{4.98} & 30.89\dar{3.09} & 69.49\dar{1.06} & 46.10\dar{2.04} & 28.89\uar{6.15} & 8.62\dar{10.95} & 50.30\dar{2.33} & 40.68\dar{2.05} \\ \midrule
\multirow{4}{*}{InfoBatch} & Random     & 60.31\uar{1.06} & 41.36\uar{0.34} & 5.38\dar{2.73} & 15.71\uar{2.96} & 47.74\dar{1.01} & 34.10\uar{0.12} & 69.31\dar{1.24} & 45.76\dar{2.38} & 18.95\dar{3.79} & 21.23\uar{1.66} & 50.39\dar{2.24} & 41.13\dar{1.60} \\ 
                           & PPL        & 59.43\uar{0.18} & 40.34\dar{0.68} & 5.91\dar{2.20} & 13.18\uar{0.43} & 48.31\dar{0.44} & 33.44\dar{0.54} & 70.72\uar{0.17} & 47.12\dar{1.02} & 18.12\dar{4.62} & 24.10\uar{4.53} & 51.26\dar{1.37} & 42.26\dar{0.47} \\
                           & FastV      & 58.90\dar{0.35} & 43.39\uar{2.37} & 3.34\dar{4.77} & 12.37\dar{0.38} & 46.88\dar{1.87} & 32.98\dar{1.00} & 69.14\dar{1.41} & 45.42\dar{2.72} & 14.86\dar{7.88} & 23.19\uar{3.62} & 50.58\dar{2.05} & 40.64\dar{2.09} \\
                           & SparseVLM  & 54.67\dar{4.58} & 40.00\dar{1.02} & 7.73\dar{0.38} & 12.41\dar{0.34} & 45.07\dar{3.68} & 31.98\dar{2.00} & 68.25\dar{2.30} & 45.08\dar{3.06} & 23.43\uar{0.69} & 10.17\dar{9.40} & 45.34\dar{7.29} & 38.46\dar{4.27} \\ \midrule
\multirow{4}{*}{Entropy}   & Random     & 60.31\uar{1.06} & 42.37\uar{1.35} & 6.44\dar{1.67} & 14.10\uar{1.35} & 48.09\dar{0.66} & 34.27\uar{0.29} & 72.13\uar{1.58} & 48.81\uar{0.67} & 20.09\dar{2.65} & 17.55\dar{2.02} & 54.69\uar{2.06} & 42.66\dar{0.07} \\ 
                           & PPL        & 60.49\uar{1.24} & 43.73\uar{2.71} & 6.90\dar{1.21} & 14.53\uar{1.78} & 48.76\uar{0.01} & 34.88\uar{0.90} & 72.84\uar{2.29} & 47.80\dar{0.34} & 24.18\uar{1.44} & 22.80\uar{3.23} & 54.69\uar{2.06} & 44.46\uar{1.73} \\
                           & FastV      & 58.91\dar{0.34} & 43.05\uar{2.03} & 6.37\dar{1.74} & 13.03\uar{0.28} & 47.05\dar{1.70} & 33.68\dar{0.30} & 73.90\uar{3.35} & 47.12\dar{1.02} & 24.56\uar{1.82} & 23.96\uar{4.39} & 54.67\uar{2.04} & 44.84\uar{2.11} \\
                           & SparseVLM  & 55.20\dar{4.05} & 38.98\dar{2.04} & 7.51\dar{0.60} & 12.65\dar{0.10} & 46.14\dar{2.61} & 32.10\dar{1.88} & 68.08\dar{2.47} & 44.07\dar{4.07} & 24.87\uar{2.13} & 10.72\dar{8.85} & 47.00\dar{5.63} & 38.95\dar{3.78} \\ \midrule
\rowcolor{mycolor} \multicolumn{2}{c|}{\textbf{Q-Tuning (Ours)}}
  & \cellcolor{mycolor}\textbf{64.20\uar{4.95}} & \cellcolor{mycolor}42.03\uar{1.01} & \cellcolor{mycolor}\textbf{10.54\uar{2.43}} & \cellcolor{mycolor}\textbf{18.79\uar{6.04}} 
  & \cellcolor{mycolor}\textbf{53.12\uar{4.37}} & \cellcolor{mycolor}\textbf{37.74\uar{3.76}} 
  & \cellcolor{mycolor}71.60\uar{1.05} & \cellcolor{mycolor}48.14\uar{0.00} & \cellcolor{mycolor}\textbf{29.34\uar{6.60}} 
  & \cellcolor{mycolor}\textbf{27.75\uar{8.18}} & \cellcolor{mycolor}\textbf{57.78\uar{5.15}} 
  & \cellcolor{mycolor}\textbf{46.92\uar{4.19}} \\  \midrule
\multicolumn{2}{c|}{Full Dataset} 
  & 61.55 & 42.37 & 8.64 & 13.80 & 50.45 & 35.36 & 71.25 & 45.76 & 26.68 & 31.81 & 53.67 & 45.84 \\
\midrule

\end{tabular}
}
\vspace{-10pt}
\end{table*}

\section{Experiments}
\label{sec:experiments}

We conducted experiments on language models from multiple families and scales, including LLaMA-series (LLaMA2-7B~\citep{touvron2023llama}, LLaMA3-8B~\citep{grattafiori2024llama}), Qwen3-series (Qwen3-8B~\citep{yang2025qwen3}, Qwen3-14B, Qwen3-32B), Mistral-7B~\citep{jiang2023mistral7b}, and SmolLM2-1.7B~\citep{allal2025smollm2}. We fine-tuned models for two settings: (i) alignment on OpenHermes~\citep{OpenHermes25} and WizardLM~\citep{xu2024wizardlm}, and (ii) reasoning on MathInstruct~\citep{yue2023mammoth}. For evaluation, we used ARC-E, ARC-C~\citep{clark2018think}, GSM8K~\citep{cobbe2021training}, SQuAD~\citep{rajpurkar2016squad}, and TriviaQA~\citep{joshi2017triviaqa} for alignment task, and GSM8K and MATH~\citep{hendrycks2021measuring} for reasoning task. We compared against baselines that pair sample-level pruning methods (Random, Longest, Entropy, InfoBatch~\citep{qin2024infobatch}, Alpagasus~\citep{chen2023alpagasus}, Deita~\citep{liu2023makes}, DS2~\citep{pang2024improving}, and LESS~\citep{xia2024less}) with token-level pruning methods (Random, PPL, FastV~\citep{chen2024image}, SparseVLM~\citep{zhang2024sparsevlm}, Rho-1~\citep{lin2024not}, and TokenCleaning~\citep{pang2025token}). More detailed experimental settings are in Appendix~\ref{appendix:exp_setting}.

\begin{table*}[tb!]
\centering
\caption{Evaluation of pruning strategies on GSM8K and MATH under 25\% samples with 50\% tokens. \ua~and \da~respectively indicate improvements or degradations over the \textit{Random-Random} baseline under the same sample and token keep ratio. Additional results under more sample and token pruning ratios are provided in the Table~\ref{tab:gsm8k_math2} in Appendix~\ref{appendix:add_reasoning}.}
\vspace{-5pt}
\label{tab:gsm8k_math} 
\resizebox{0.99\textwidth}{!}{%
\setlength{\tabcolsep}{8pt} 
\begin{tabular}{c|l|ccc|ccc|ccc}
\toprule 
\multicolumn{1}{c|}{Sample} & \multicolumn{1}{c|}{Token} & \multicolumn{3}{c|}{\textbf{LLaMA3-8B}} 
& \multicolumn{3}{c|}{\textbf{Mistral-7B}} & \multicolumn{3}{c}{\textbf{SmolLM2-1.7B}} \\
\multicolumn{1}{c|}{Pruner} & \multicolumn{1}{c|}{Pruner} & GSM8K & MATH & Avg. & GSM8K & MATH & Avg. & GSM8K & MATH & Avg. \\ 
\midrule

\multicolumn{2}{c|}{Zero-Shot} 
  & 27.82 & 2.26 & 15.04 & 19.86 & 3.30 & 11.58 & 15.47 & 2.20 & 8.83 \\
\midrule

\rowcolor{gray!5}
\multicolumn{11}{c}{\textit{$25\%$ Samples, $50\%$ Tokens}} \\ \midrule
\multirow{4}{*}{Random}    & Random     & 23.96 & 2.56 & 13.26 & 23.35 & 1.54 & 12.45 & 14.33 & 2.56 & 8.44 \\
                           & PPL        & 24.18\uar{0.22} & 2.58\uar{0.02} & 13.38\uar{0.12} & 24.94\uar{1.59} & 2.02\uar{0.48} & 13.48\uar{1.03} & 14.18\dar{0.15} & 2.08\dar{0.48} & 8.13\dar{0.31} \\
                           & FastV      & 12.13\dar{11.83}& 2.32\dar{0.24} & 7.23\dar{6.03}  & 12.36\dar{10.99}& 1.24\dar{0.30} & 6.80\dar{5.65}  &  9.86\dar{4.47} & 1.92\dar{0.64} & 5.89\dar{2.55} \\
                           & SparseVLM  & 22.97\dar{0.99} & 4.72\uar{2.16} & 13.85\uar{0.59} & 19.26\dar{4.09} & 4.58\uar{3.04} & 11.92\dar{0.53} & 13.19\dar{1.14} & 3.48\uar{0.92} & 8.34\dar{0.10} \\ \midrule
\multirow{4}{*}{Longest}   & Random     & 22.14\dar{1.82} & 3.18\uar{0.62} & 12.66\dar{0.60} & 21.91\dar{1.44} & 2.18\uar{0.64} & 12.05\dar{0.40} & 12.89\dar{1.44} & 2.06\dar{0.50} & 7.47\dar{0.97} \\
                           & PPL        & 24.94\uar{0.98} & 2.78\uar{0.22} & 13.86\uar{0.60} & 22.90\dar{0.45} & 1.86\uar{0.32} & 12.38\dar{0.07} & 13.19\dar{1.14} & 1.78\dar{0.78} & 7.49\dar{0.95} \\
                           & FastV      & 9.48\dar{14.48} & 2.26\dar{0.30} & 5.87\dar{7.39}  &  7.13\dar{16.22}& 1.46\dar{0.08} & 4.29\dar{8.16}  & 12.36\dar{1.97} & 1.82\dar{0.74} & 7.09\dar{1.35} \\
                           & SparseVLM  & 26.91\uar{2.95} & 4.68\uar{2.12} & 15.80\uar{2.54} & 24.34\uar{0.99} & \textbf{4.84\uar{3.30}} & 14.59\uar{2.14} & 12.43\dar{1.90} & 3.60\uar{1.04} & 8.02\dar{0.42} \\ \midrule
\multirow{4}{*}{InfoBatch} & Random     & 26.23\uar{2.27} & 2.42\dar{0.14} & 14.33\uar{1.07} & 27.14\uar{3.79} & 2.24\uar{0.70} & 14.69\uar{2.24} & 14.33\uar{0.00} & 1.66\dar{0.90} & 7.99\dar{0.45} \\
                           & PPL        & 26.91\uar{2.95} & 2.66\uar{0.10} & 14.79\uar{1.53} & 27.90\uar{4.55} & 2.52\uar{0.98} & 15.21\uar{2.76} & 14.71\uar{0.38} & 1.90\dar{0.66} & 8.30\dar{0.14} \\
                           & FastV      & 7.58\dar{16.38} & 1.88\dar{0.68} & 4.73\dar{8.53}  &  6.44\dar{16.91}& 1.34\dar{0.20} & 3.89\dar{8.56}  &  7.51\dar{6.82} & 1.62\dar{0.94} & 4.56\dar{3.88} \\
                           & SparseVLM  & 14.63\dar{9.33} & 3.26\uar{0.70} & 8.95\dar{4.31}  & 11.90\dar{11.45}& 1.94\uar{0.40} & 6.92\dar{5.53}  & 11.90\dar{2.43} & \textbf{4.36\uar{1.80}} & 8.13\dar{0.31} \\ \midrule
\multirow{4}{*}{Entropy}   & Random     & 30.02\uar{6.06} & 3.66\uar{1.10} & 16.84\uar{3.58} & 26.61\uar{3.26} & 2.08\uar{0.54} & 14.35\uar{1.90} & 14.18\dar{0.15} & 2.24\dar{0.32} & 8.21\dar{0.23} \\
                           & PPL        & 32.98\uar{9.02} & 2.92\uar{0.36} & 17.95\uar{4.69} & 30.17\uar{6.82} & 1.76\uar{0.22} & 15.97\uar{3.52} & 16.38\uar{2.05} & 2.40\dar{0.16} & 9.39\uar{0.95} \\
                           & FastV      & 17.29\dar{6.67} & 2.66\uar{0.10} & 9.97\dar{3.29}  & 14.56\dar{8.79} & 1.12\dar{0.42} & 7.84\dar{4.61}  & 12.59\dar{1.74} & 2.36\dar{0.20} & 7.47\dar{0.97} \\
                           & SparseVLM  & 20.85\dar{3.11} & 5.12\uar{2.56} & 12.98\dar{0.28} & 19.56\dar{3.79} & 4.20\uar{2.66} & 11.88\dar{0.57} & 14.18\dar{0.15} & 2.90\uar{0.34} & 8.54\uar{0.10} \\ \midrule
\rowcolor{mycolor} \multicolumn{2}{c|}{\textbf{Q-Tuning (Ours)}} 
  & \cellcolor{mycolor}\textbf{36.32\uar{12.36}} & \cellcolor{mycolor}\textbf{5.54\uar{2.98}} & \cellcolor{mycolor}\textbf{20.93\uar{7.67}} & \cellcolor{mycolor}\textbf{41.47\uar{18.12}} & \cellcolor{mycolor}4.0\uar{2.46} & \cellcolor{mycolor}\textbf{22.74\uar{10.29}} & \cellcolor{mycolor}\textbf{21.83\uar{7.50}} & \cellcolor{mycolor}3.90\uar{1.34} & \cellcolor{mycolor}\textbf{12.87\uar{4.43}} \\ \midrule
\multicolumn{2}{c|}{Full Dataset} 
  & 32.90 & 3.02 & 17.96 & 42.08 & 3.08 & 22.58 & 16.53 & 2.10 & 9.31 \\
\midrule
\end{tabular}
}
\vspace{-10pt}
\end{table*}


\subsection{Main Results}
\paragraph{Results on alignment datasets.}
Table~\ref{tab:wizard_long} summarizes results in our alignment setting under a fixed low-budget regime (12.5\% samples, 50\% tokens) on WizardLM. Across both LLaMA2-7B and Mistral-7B, Q-Tuning achieves the best overall averages (37.74 and 46.92), outperforming all evaluated combinations of existing sample-pruning and token-pruning methods, and it also exceeds full-data fine-tuning (35.36 and 45.84). Many baseline pairings improve one or two benchmarks but exhibit noticeable trade-offs on others (\emph{e.g.}, degradation on GSM8K or SQuAD for several token-pruning choices), indicating that naively combining independently designed sample- and token-level pruners is insufficient under a tight compute budget. In contrast, Q-Tuning delivers consistent gains across all reported benchmarks, suggesting that jointly coordinating sample triage with targeted token reduction better preserves high-utility supervision in this broad alignment evaluation. We report additional results under more pruning ratios in Appendix~\ref{appendix:add_instruction} (Table~\ref{tab:wizard_long3} and Table~\ref{tab:wizard_long2}), showing similar performance trends, while Appendix~\ref{appendix:instruction_qwen} presents detailed results and analysis on Qwen3-8B with larger performance gains of up to 8.23 points and improvements of up to 5.55 over the strongest baseline. Overall, these results highlight that Q-Tuning is robust to heterogeneous task mixes and remains effective even when both sample and token budgets are simultaneously constrained across diverse model families.

\paragraph{Results on reasoning dataset.}
Table~\ref{tab:gsm8k_math} compares Q-Tuning against diverse baselines on math reasoning datasets under constrained sample and token budgets. Several methods, especially FastV-based token pruning, degrade sharply (e.g., driving LLaMA3-8B average accuracy below 6\% in multiple settings). In contrast, quality-aware sample selection (Entropy, InfoBatch) consistently improves over random sampling: Entropy--Random raises LLaMA3-8B from 13.26 to 16.84 and yields similar gains on Mistral-7B. Across all model scales, Q-Tuning is the strongest and most stable, improving the Random--Random baseline by 7.67 points on LLaMA3-8B and 10.29 on Mistral-7B, reaching 20.93 and 22.74, respectively. Notably, Q-Tuning exceeds full-data GSM8K performance for both models despite using only a quarter of the samples and half of the tokens, indicating strong data efficiency and better training-resource utilization; larger models benefit more from the joint strategy. Additional ratios are reported in Appendix~\ref{appendix:add_reasoning} (Table~\ref{tab:gsm8k_math2}). This suggests that jointly optimized pruning can improve reasoning training signal quality, rather than merely reducing compute. In particular, it preferentially retains error-revealing and correction-rich spans while trimming redundant verbiage. As a result, the model receives cleaner gradient signals and learns more effectively.



\begin{figure}[tb!]
  \centering
  \includegraphics[width=\columnwidth]{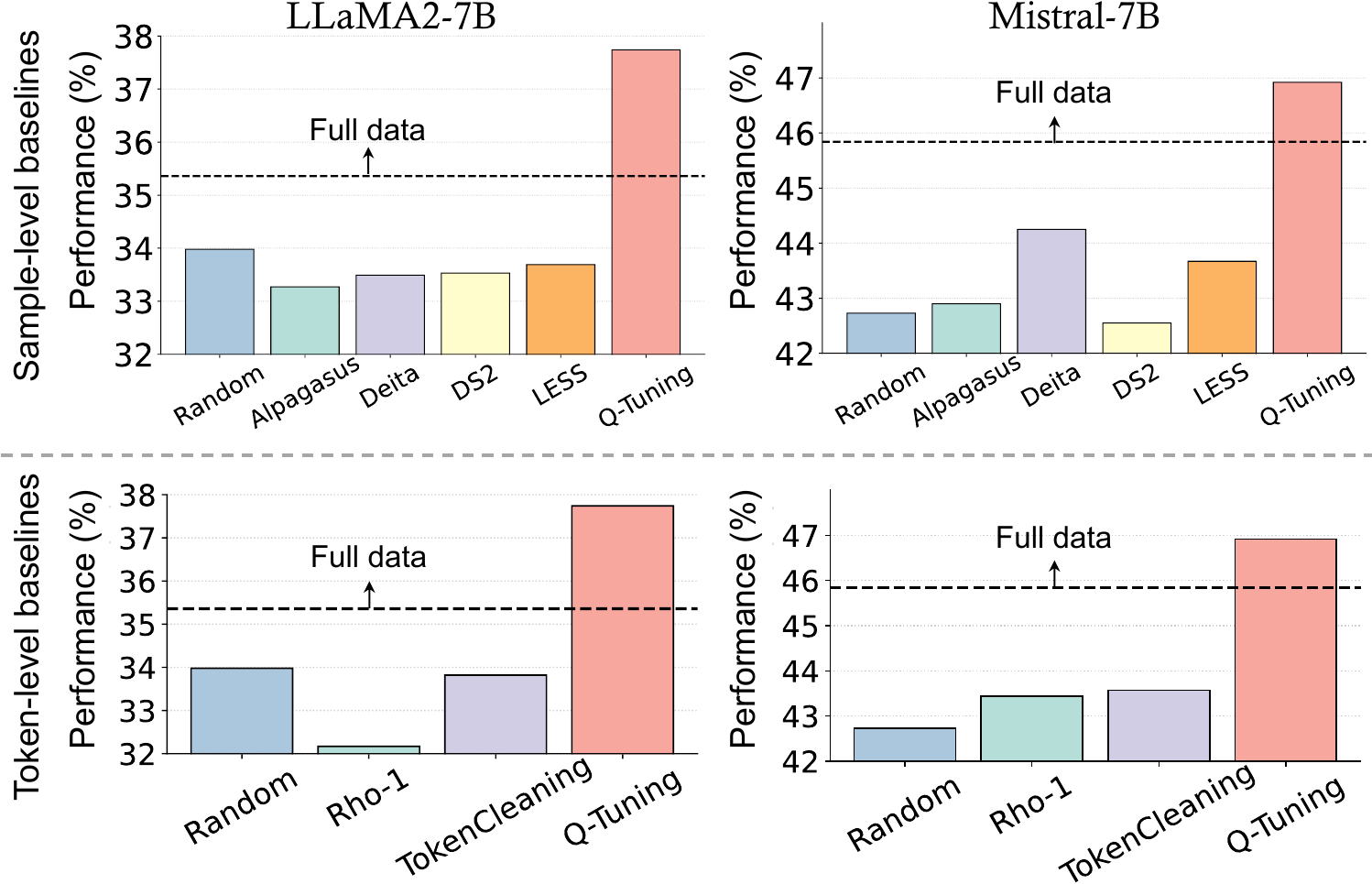}
  \caption{Comparison with stronger task-relevant baselines under matched low-budget settings. Top: Sample-pruner baselines with a sample ratio of 12.5\% and a token ratio of 50\%. Bottom: Sample-pruner baselines with a sample ratio of 12.5\% and a token ratio of 50\%; the dashed line marks the full-data upper bound. Additional results under more pruning ratios are reported in Appendix~\ref{appendix:add_baseline}.}
  \label{fig:stronger_baselines_main}
  \vspace{-10pt}
\end{figure}

\paragraph{Comparison with stronger baselines.}
Figure~\ref{fig:stronger_baselines_main} compares Q-Tuning with stronger, task-relevant baselines under matched low-budget. In top figures with 12.5\% samples and 50\% tokens, Q-Tuning achieves the best average on both models, scoring 37.74 on LLaMA2-7B and 46.92 on Mistral-7B, which exceeds the strongest competing sample-selection baseline by +3.44 and +2.00, respectively. In bottom figures with a fixed 50\% token budget, Q-Tuning also outperforms token-level baselines at the same 12.5\% sample ratio, improving over the best baseline from 33.82 to 37.74 on LLaMA2-7B (+3.92) and from 43.57 to 46.92 on Mistral-7B (+3.35). These results indicate that Q-Tuning remains competitive against stronger baselines under the same constrained budgets and delivers consistent gains across model families. Additional results under more pruning ratios are reported in Appendix~\ref{appendix:add_baseline}. 

\paragraph{Robustness to extreme data budgets.} 
Table~\ref{tab:less_sample} shows that Q-Tuning remains highly data-efficient even when retaining only 3--9\% of the original training tokens on Mistral-7B. Across all four extreme settings, the average score ranges from 46.23 to 46.92, consistently outperforming both the zero-shot baseline (36.98) and the full-dataset SFT baseline (45.84), despite using at most 8.75\% of the data and maintaining strong per-task accuracy. Notably, with just 6.25\% samples and 50\% tokens (\emph{i.e.}, 3.125\% data retention), Q-Tuning already achieves an average of 46.23, exceeding full-data fine-tuning. Increasing the token ratio to 70\% at the same sample budget further improves the average to 46.62, while still operating in a regime where less than 5\% of the original data is used in training epochs.

\begin{table}[tb!]
\centering
\caption{Q-Tuning under extreme data-retention settings on the Wizard dataset using Mistral-7B.}
\vspace{-5pt}
\label{tab:less_sample}
\resizebox{0.99\linewidth}{!}{%
\begin{tabular}{c|c|c|cccccc}
\toprule 
\multicolumn{1}{c|}{Sample} & \multicolumn{1}{c|}{Token} & \multicolumn{1}{c|}{Data} & \multicolumn{6}{c}{\textbf{Mistral-7B}} \\
\multicolumn{1}{c|}{Ratio} & \multicolumn{1}{c|}{Ratio} & \multicolumn{1}{c|}{Retention} & ARC-E & ARC-C & GSM8K & SQuAD & TriviaQA & Avg. \\ 
\midrule

\multicolumn{2}{c|}{Zero-Shot} 
 & - & 66.67 & 46.10 & 18.35 & 10.01 & 43.77 & 36.98 \\
\midrule

6.25\% & 50\% & 3.125\% & 70.72 & 46.10 & \textbf{40.41} & 17.25 & 56.68 & 46.23 \\
6.25\% & 70\% & 4.375\% & 71.60 & \textbf{48.47} & 39.73 & 16.13 & 57.16 & 46.62 \\
12.50\% & 50\% & 6.250\% & 71.60 & 48.14 & 29.34 & 27.75 & \textbf{57.78} & \textbf{46.92} \\
12.50\% & 70\% & 8.750\% & \textbf{71.78} & 47.12 & 26.08 & \textbf{32.79} & 56.17 & 46.79 \\ 
\midrule
\multicolumn{2}{c|}{Full Dataset} 
 & 100\% & 71.25 & 45.76 & 26.68 & 31.81 & 53.67 & 45.84 \\
\midrule
\end{tabular}
}
\vspace{-10pt}
\end{table}

\begin{figure}[htbp]
    \centering
    \includegraphics[width=0.99\linewidth]{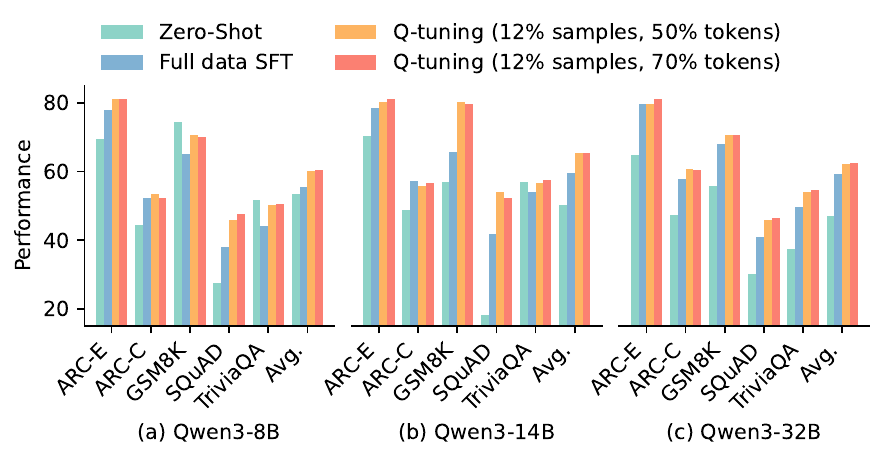} 
    \vspace{-10pt}
    \caption{Comparison of Qwen3-series of varying scales (8B, 14B, 32B) across multiple benchmarks and their average. We report performance of Zero-shot, full dataset, and Q-Tuning with 12.5\% samples under two token ratios (50\% and 70\%).}
    \vspace{-10pt}
    \label{fig:different_scale}
\end{figure}

\paragraph{Generalization across model scales.} 
Figure~\ref{fig:different_scale} studies Q-Tuning on Qwen3-series of increasing scale (8B, 14B, and 32B). Under the same constrained budgets (12.5\% samples with 50\% or 70\% tokens), Q-Tuning consistently matches or exceeds full-dataset SFT across benchmarks, demonstrating strong scalability despite using only a small fraction of the training data. The gains are especially evident on reasoning- and QA-oriented tasks such as GSM8K, ARC-E, and SQuAD. Moreover, for the 32B model, increasing the token budget from 50\% to 70\% typically yields further improvements and delivers the strongest overall averages.

\subsection{Further Analysis}

\paragraph{Effect of quadrant-wise pruning strategies.}
Table~\ref{tab:different_quadrant3} reports a quadrant-wise ablation on Qwen3-8B under two budget settings. Retaining Q1 consistently yields the lowest averages, with pronounced drops on reasoning and QA tasks (\emph{e.g.}, GSM8K and SQuAD), whereas pruning Q1 leads to more stable performance. Token-level pruning of Q4 is generally harmful, indicating that these low-PPL yet diverse tokens carry useful signal and should be preserved. In contrast, pruning Q2 tokens is crucial, as retaining Q2’s high-PPL tokens degrades performance while removing them consistently improves results. Therefore, the best configurations prune Q1 and Q3 at the sample level and prune Q2 at the token level, supporting the design of Q-Tuning. Additional quadrant-wise ablations are provided in Appendix~\ref{appendix:add_quadrant}.

\begin{table}[htbp]
\centering
\caption{Ablation on four quadrants on OpenHermes and Qwen3-8B under different sample and token ratios, where \checkmark indicates that the corresponding quadrant is pruned. Additional quadrant-wise ablations are provided in Appendix~\ref{appendix:add_quadrant}.}
\vspace{-5pt}
\label{tab:different_quadrant3}
\resizebox{0.99\linewidth}{!}{%
\begin{tabular}{cc|cc|cccccc}
\toprule 
\multicolumn{2}{c|}{Sample pruning} & \multicolumn{2}{c|}{Token pruning} & \multicolumn{6}{c}{\textbf{Qwen3-8B}} \\
Q1 & Q3  &  Q2  & Q4 & ARC-E & ARC-C & GSM8K & SQuAD & TriviaQA & Avg. \\ 
\midrule
\rowcolor{gray!5}
\multicolumn{10}{c}{\textit{$12.5\%$ Samples, $50\%$ Tokens}} \\ \midrule
&   \checkmark &  \checkmark &  & 79.01 & 52.20 & \textbf{74.68} & 36.35 & 48.20 & 58.09 \\
\checkmark &   \checkmark &   \checkmark &   \checkmark & 79.37 & 52.20 & 71.80 & 35.55 & 47.23 & 57.23 \\
\checkmark &   \checkmark &     &  & 78.84 & \textbf{53.90} & 68.99 & 44.32 & 49.15 & 59.04 \\
\checkmark &   \checkmark &  \checkmark &    & \textbf{80.95} & 53.56 & 70.51 & \textbf{45.76} & \textbf{50.16} & \textbf{60.19} \\ \midrule
\rowcolor{gray!5}
\multicolumn{10}{c}{\textit{$25\%$ Samples, $50\%$ Tokens}} \\ \midrule
&   \checkmark &     \checkmark &    & 75.31 & 50.17 & 68.08 & 29.47 & 45.90 & 53.79 \\
\checkmark &   \checkmark &   \checkmark &   \checkmark & 76.37 & 50.17 & 65.28 & 31.63 & 45.24 & 53.74 \\
\checkmark &   \checkmark &   \checkmark &    & \textbf{79.89} & \textbf{52.88} & \textbf{69.37} & \textbf{43.54} & \textbf{48.33} & \textbf{58.80} \\ 
\midrule
\end{tabular}
}
\vspace{-5pt}
\end{table}

\begin{figure}[tb]
    \vspace{-5pt}
    \centering
    \includegraphics[width=\linewidth]{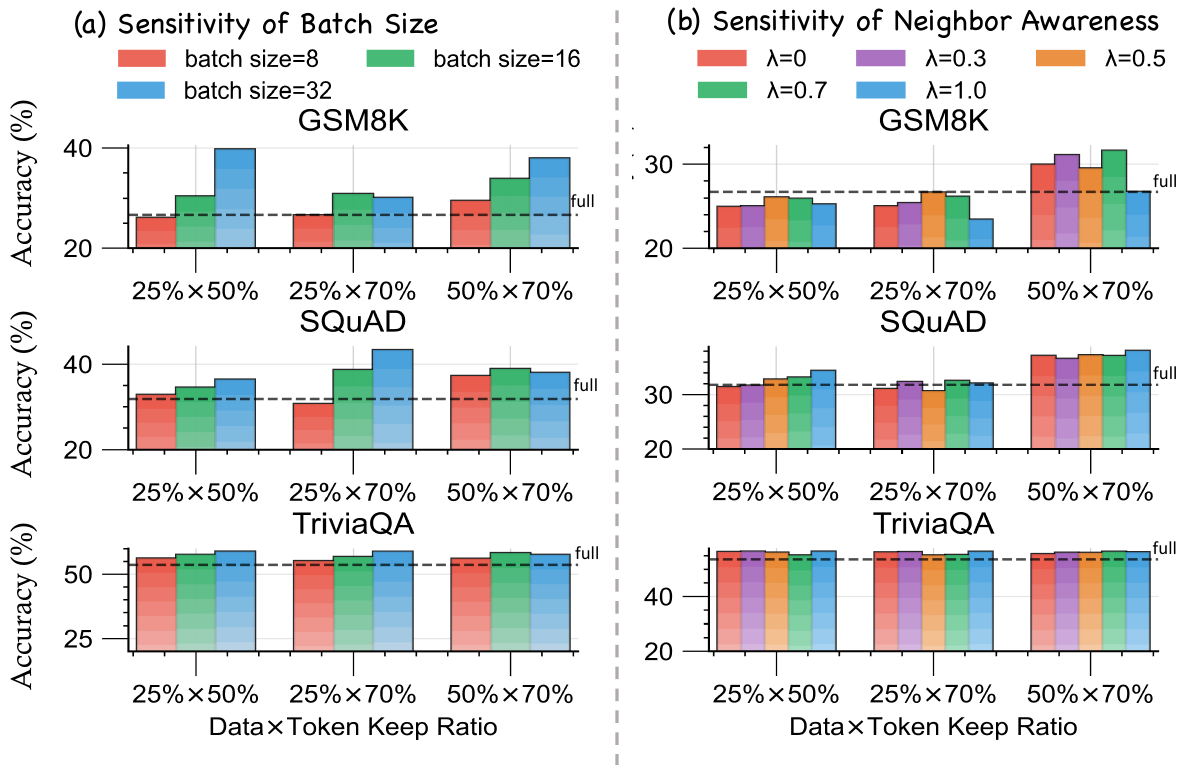}
    \vspace{-15pt}
    \caption{Effect of varying (a) batch size and (b) neighbor awareness for Mistral-7B under three keep ratio configurations. Additional benchmark results (Avg. of five benchmarks, ARC-E, ARC-C) are provided in the Figure~\ref{fig:ablation_supp} in Appendix~\ref{app:hyp}.}
    \vspace{-15pt}
    \label{fig:ablation}
\end{figure}

\paragraph{Sensitivity to hyperparameters.}
As shown in Figure~\ref{fig:ablation}, we vary batch size ($8/16/32$) and neighbor awareness $\lambda$ ($0$--$1.0$) on Mistral-7B under multiple keep-ratio settings, and find Q-Tuning to be largely robust. Larger batches and moderate $\lambda$ tend to improve GSM8K and SQuAD, while TriviaQA is relatively stable (see Figure~\ref{fig:ablation_supp} in Appendix~\ref{app:hyp}). 

\begin{figure*}[htbp]
    \centering
    \includegraphics[width=0.95\textwidth]{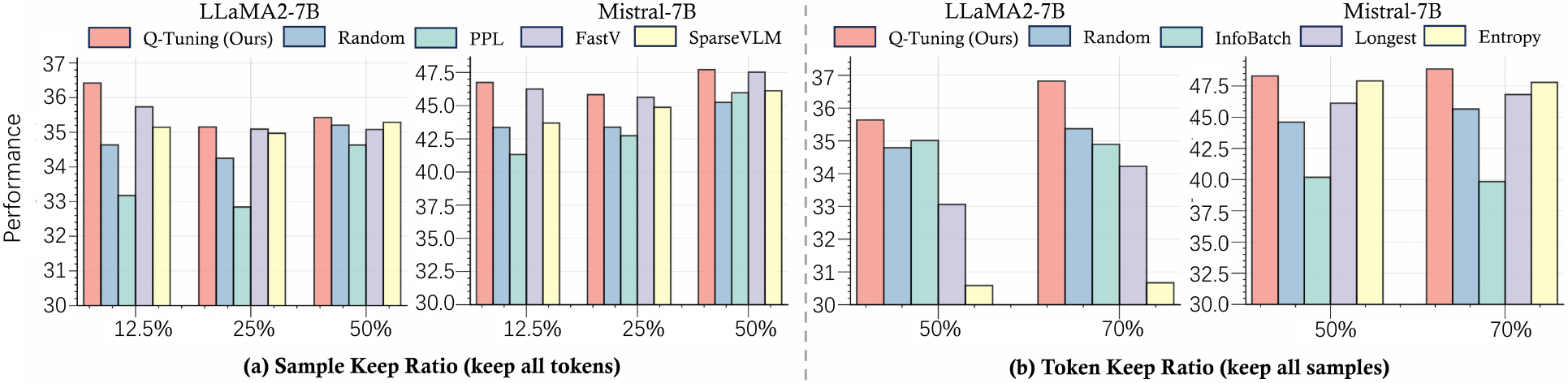} 
    \caption{
    Comparison of independent (a) dynamic sample pruning and (b) dynamic token pruning across different keep ratios for LLaMA2-7B and Mistral-7B.
    }
    \label{fig:ablation_100}
\end{figure*}
\paragraph{Effect of token pruning strategy.}
We compare different token pruning strategies under a fixed $25\%\times50\%$ budget on Mistral-7B, including Rho-1, PPL, and a reversed PPL variant. The neighbor-aware PPL used in Q-Tuning achieves the best average performance, while alternative pruners are less consistent across tasks (see Table~\ref{tab:tokenratio50}). This supports our design choice of locality-aware scoring that preserves salient error-relevant spans.

\begin{table}[htbp]
\centering
\vspace{-5pt}
\caption{Ablation study on token pruner in Q-tuning}
\vspace{-5pt}
\label{tab:tokenratio50}
\resizebox{\linewidth}{!}{%
\begin{tabular}{l|lccccc}
\toprule
 Method &  Avg. &ARC-E & ARC-C & GSM8K & SQuAD & TriviaQA \\
\midrule
 Rho-1   &  45.03 &69.66 & 46.78 & 24.03 & 29.23 & 55.43 \\
 PPL ($\lambda=0$)&  45.92 &71.60 & 46.44 & 25.32 & 29.71 & \textbf{56.54} \\
 Reversed PPL ($\lambda=0.5$) &  44.86 &\textbf{73.02} & 47.12 & 16.68 & 32.01 & 55.47 \\
\rowcolor{cyan!5}
 PPL ($\lambda=0.5$) &  \textbf{46.79} &71.78 & \textbf{47.12} & \textbf{26.08} & \textbf{32.79} & 56.17 \\
\bottomrule
\end{tabular}
}
\vspace{-10pt}
\end{table}

\begin{figure}[htbp]
    \centering
    \includegraphics[width=0.99\linewidth]{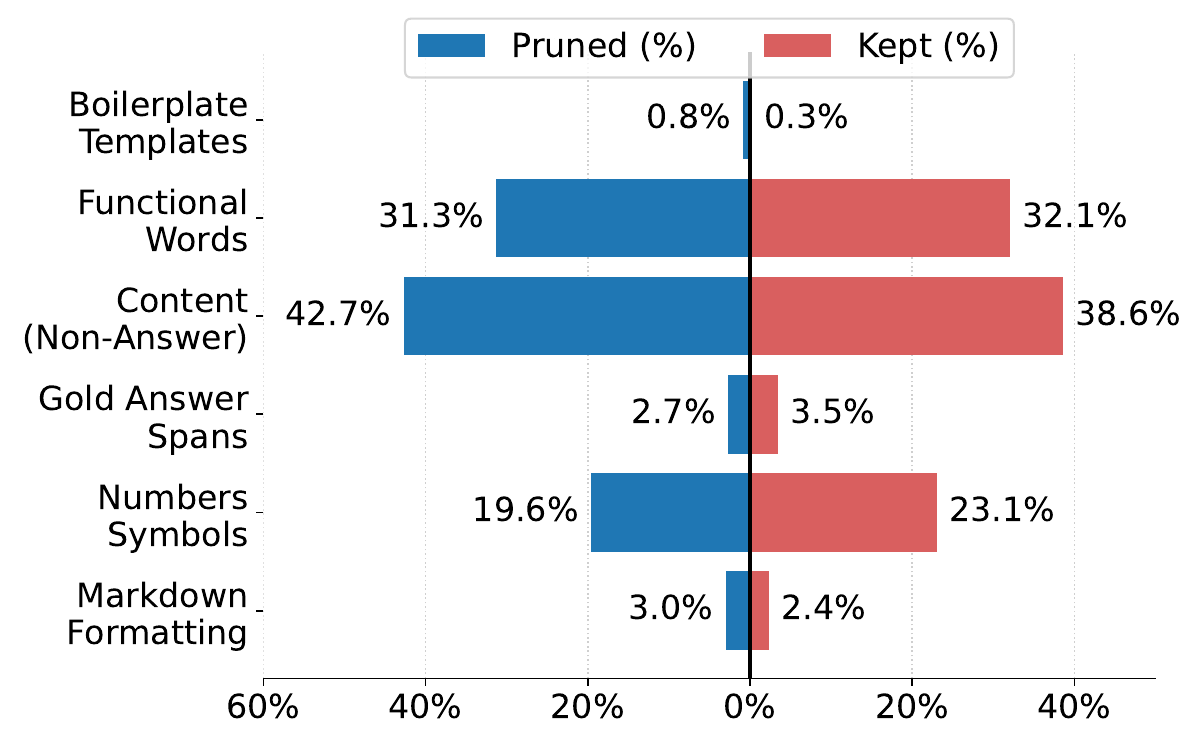}
    \caption{Semantic distribution of pruned vs. retained tokens in the Q2 quadrant, based on 300 instances annotated by GPT-5.}
    \label{fig:semantic_q2}
\end{figure}

\paragraph{Case Study: Which Types of Tokens Are Preferentially Pruned in Q2?} To better characterize the semantic effects of Q2 pruning, we perform a microscopic analysis within the Q2 quadrant. We randomly sample 300 Q2 instances and use GPT-5~\citep{singh2025openai} to annotate functional roles for 44K+ tokens, then compare category proportions between pruned and kept sets. Figure~\ref{fig:semantic_q2} shows that redundant content is over-represented among pruned tokens: boilerplate/templates and formatting increase, and non-answer content shifts toward pruning. In contrast, supervision-critical signals are enriched in kept tokens. Gold answer spans account for a higher proportion among retained tokens, and numbers/symbols are substantially more prevalent. We provide quadrant examples in Appendix~\ref{appendix:example_quadrant} and token-level pruning cases in Appendix~\ref{sec:pruned-case-study}.


\subsection{Efficiency Analysis: Empirical Latency}


Table~\ref{tab:efficiency_combined} quantifies the end-to-end efficiency of Q-Tuning on LLaMA2-7B under our standard three-epoch setup with $8\times$ A100 GPUs. Compared to full-data SFT (90 minutes, 1.0$\times$ FLOPs, 78\% peak memory), Q-Tuning with 12.5\% samples and 50\% tokens reduces training compute to 0.0625$\times$ and completes in 65 minutes, yielding a $\approx$28\% wall-clock reduction while lowering peak memory to 27\%. The screening overhead is modest relative to the overall budget and is amortized by the savings from skipping backward/optimizer updates on pruned data. Even as budget increases (25\% and 50\% samples at 50\% tokens), total time remains below full-data SFT (70–80 minutes) with substantially reduced FLOPs (0.125–0.25$\times$) and peak memory (35–50\%), indicating a favorable practical cost–benefit trade-off. 

\begin{table}[htbp]
\centering
\caption{Efficiency and overhead breakdown on LLaMA2-7B under different pruning settings.} 
\label{tab:efficiency_combined}
\resizebox{0.99\linewidth}{!}{%
\begin{tabular}{cc|cc|cc}
\toprule
\multicolumn{2}{c|}{Pruner} 
& \multicolumn{2}{c|}{Selection Time (mins)} 
& \multicolumn{2}{c}{Training Efficiency} \\
\cmidrule(lr){1-2}\cmidrule(lr){3-4}\cmidrule(lr){5-6}
Sample & Token 
& Sample & Token 
& Total & Peak \\
Ratio (\%)  & Ratio (\%) 
& Selection & Selection 
& Time (mins) & Mem (\%) \\
\midrule
12.5        & 50   & 35.88 & 6.96 & 65 & 27 \\
25          & 50   & 38.64 & 7.49 & 70 & 35 \\
50          & 50   & 44.16 & 8.56 & 80 & 50 \\
100 (Full)  & 100  & 0.00  & 0.00 & 90 & 78 \\
\bottomrule
\end{tabular}
}
\end{table}

\section{Discussion}
\paragraph{Can Q-Tuning outperform independent sample or token pruning?}
To isolate the effect of each pruning dimension, we conduct controlled ablations that apply \emph{only} dynamic sample pruning (retaining all tokens) or \emph{only} dynamic token pruning (retaining all samples). As shown in Figure~\ref{fig:ablation_100}, Q-Tuning consistently outperforms all baseline methods across keep ratios on both LLaMA2-7B and Mistral-7B, indicating that its advantage does not rely on a particular pruning axis but stems from more effective data utility modeling.

\begin{figure}[htbp]
    \centering
    \includegraphics[width=\linewidth]{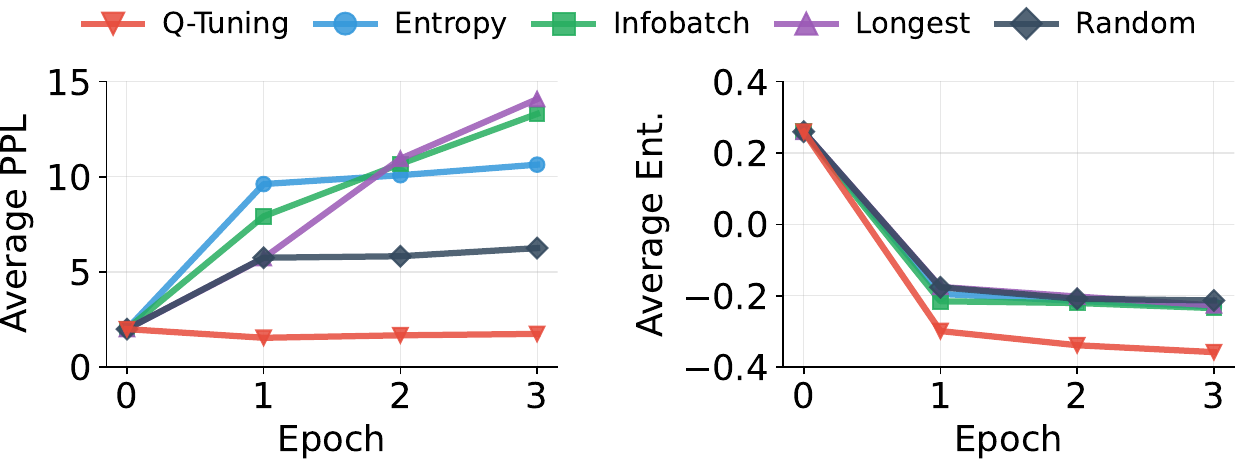}
    \vspace{-10pt}
    \caption{Training dynamics of different sample pruners. Compared to baseline strategies, Q-Tuning consistently reduces both metrics at a faster rate, indicating more efficient learning dynamics.}
    \vspace{-15pt}
    \label{fig:distribution}
\end{figure}

\paragraph{How does the sample distribution evolve during training?}
To characterize training dynamics, we track the average perplexity and token entropy on 100 randomly sampled training instances throughout fine-tuning. As shown in Figure~\ref{fig:distribution}, Q-Tuning reduces both metrics more rapidly than alternative sample-pruning strategies, suggesting faster stabilization of the training signal and correlating with improved downstream performance. 

\section{Conclusion}
This work turns risky dynamic data pruning for LLM fine-tuning into a reliable, high-utility strategy. By analyzing sample modes via the Error–Uncertainty Plane, we expose data’s heterogeneous value and the need for nuance. Our Quadrant-based Tuning (Q-Tuning) uses a two-stage framework to coordinate sample- and token-level pruning, preserving valuable signals while removing noise, boosting efficiency without sacrificing overall stability, and often further improving performance across diverse tasks and models.

\section*{Impact Statement}
Our work makes dynamic data pruning a more reliable and practical tool for LLM post-training. By using simple, interpretable signals (perplexity and entropy) to guide Q-Tuning, we better preserve informative data while removing noise and redundancy, improving efficiency without sacrificing stability. This reduces compute and carbon cost and improves reproducibility, making high-quality post-training more accessible.



\bibliography{icml2026}
\bibliographystyle{icml2026}

\newpage
\appendix
\onecolumn
\appendix

\section{Related Work}

\subsection{Sample Pruning}

A growing body of work explores sample pruning as a means to reduce training cost by selecting smaller yet higher-quality subsets of data without compromising performance~\citep{ivison2025large, yang2024smalltolarge, xia2024less, zhou2024programming}. For example, LIMA~\citep{zhou2023lima} shows that aligning LLMs requires only a small collection of high-quality instruction–response pairs rather than large-scale corpora. Broadly, existing approaches fall into static and dynamic pruning. In the static setting, data subsets are determined in advance using fixed criteria: SVP~\citep{coleman2020selection} leverages proxy models to estimate sample importance, D3~\citep{zhang2025d3} combines diversity, difficulty, and dependability into weighted coresets, and Less is More~\citep{deng2025less} shows that carefully curated preference subsets can outperform full datasets by filtering noisy or redundant examples. Other static approaches target in-context examples: LIMO~\citep{ye2025limo} removes redundant demonstrations while distilling essential reasoning patterns, and methods such as DEFT~\citep{ivison2023data}, Alpagasus~\citep{chen2023alpagasus}, and Data Whisperer~\citep{wang2025datawhisperer} automatically select or reweight demonstrations based on influence estimation, contribution to performance, or few-shot evaluation. 

In contrast, dynamic pruning adapts sample usage throughout training. Early work by~\citep{raju2021accelerating} proposed two methods, $\epsilon$-greedy and UCB, which retain uncertain examples while discarding easier ones. Subsequent approaches explore alternative criteria for adaptive pruning: RHO-LOSS~\citep{mindermann2022prioritized} assigns importance using high-quality reference models, InfoBatch~\citep{qin2024infobatch} removes low-loss examples on the fly, and more recent methods such as MATES~\citep{yu2024mates} and DWM~\citep{yu2025llm} reweight or select samples according to the evolving model state.Beyond single-example pruning, several methods consider structured or representation-based criteria: ActivePrune~\citep{azeemi2024language} selects demonstrations per input via similarity metrics, SCAN~\citep{guo2025scan} iteratively removes ill-matched and redundant data during contrastive pretraining, GDeR~\citep{zhang2024gder} models data as a prototypical graph to prune noisy or redundant samples, and DQ~\citep{zhou2023dataset} clusters data in representation space and replaces each cluster with representative subsets.

\subsection{Token Pruning}

Parallel to sample pruning, recent research~\citep{wang2025beyond, pan2024llmlingua, keith2024optimizing} has increasingly focused on token-level pruning to retain only the most informative portions of input sequences. At the pretraining stage, methods such as Rho-1~\citep{lin2024not} leverage reference models to estimate token importance, discarding uninformative tokens. In fine-tuning and inference, pruning is often performed dynamically: Token Cleaning~\citep{pang2025token} identifies harmful tokens as noisy labels and prunes them based on their estimated influence on parameter updates, while approaches like LazyLLM~\citep{fu2024lazyllm} and SlimInfer~\citep{long2025sliminfer} accelerate long-context inference by selectively dropping tokens or hidden-state blocks with the help of attention signals and cache mechanisms. Along similar lines, TokenSkip~\citep{xia2025tokenskip} removes redundant reasoning steps by skipping semantically less important tokens. Another family of approaches instead performs input compression before inference, exemplified by LLMLingua~\citep{jiang2023llmlingua} and Selective Context~\citep{li2023compressing}, which prune low-information tokens, phrases, or sentences through coarse-to-fine filtering with budget control. Beyond text-only models, multimodal pruning exploits cross-modal attention patterns: FastV~\citep{chen2024image} and SparseVLM~\citep{zhang2024sparsevlm} drop redundant visual tokens guided by attention, while LMTL~\citep{huang2023fewer} dynamically adjusts computation by pruning unnecessary visual features. More recent advances further refine pruning with task-specific objectives, such as GAP~\citep{chien2025grounding}, which preserves spatial grounding by correcting position misalignments during token removal, and DART~\citep{wen2025stop}, which reduces token-level computation through confidence-based early stopping. 

Despite their progress, prior work typically treats sample pruning and token pruning as independent lines of research. Sample pruning methods focus on reducing the number of training examples but leave token-level redundancy unaddressed, while token pruning techniques emphasize sequence-level efficiency without considering redundancy across training samples. Such separation limits the potential for jointly optimizing efficiency and effectiveness. In this work, we bridge these directions by proposing a unified framework that integrates sample-level and token-level pruning, enabling models to simultaneously filter uninformative data and compress redundant inputs.

\section{Experimental Settings}
\label{appendix:exp_setting}
\subsection{Models and Datasets.} 

We conducted experiments on language models of different scales, including \emph{Mistral-7B}~\citep{jiang2023mistral7b}, \emph{LLaMA2-7B}~\citep{touvron2023llama}, \emph{LLaMA3-8B}~\citep{grattafiori2024llama}, \emph{SmolLM2-1.7B}~\citep{allal2025smollm2}, \emph{Qwen3-8B}~\citep{yang2025qwen3} and larger Qwen3-series (\emph{Qwen-14B}, \emph{Qwen-32B}). For Qwen3-series, we fine-tune on the \emph{OpenHermes}~\citep{OpenHermes25}, which contains diverse instruction–response pairs for general-purpose alignment. To fine-tune other models, we considered two datasets that target complementary aspects of alignment: \emph{WizardLM}~\citep{xu2024wizardlm} for alignment, which provides high-quality instruction–response pairs for supervised fine-tuning (SFT), and \emph{MathInstruct}~\citep{yue2023mammoth} for reasoning, which contains mathematically focused instructions designed to assess arithmetic and symbolic problem-solving. 
\vspace{-8pt}
\subsection{Evaluation.} 

For alignment task, we used five standard benchmarks: ARC-E, ARC-C~\citep{clark2018think}, GSM8K~\citep{cobbe2021training}, SQuAD~\citep{rajpurkar2016squad}, and TriviaQA~\citep{joshi2017triviaqa}. These tasks spanned knowledge-intensive question answering, commonsense reasoning, and reading comprehension, offering a broad assessment of model capability. For reasoning task, we evaluated models fine-tuned on MathInstruct using GSM8K and MATH~\citep{hendrycks2021measuring}, two widely used benchmarks for mathematical problem solving.
\vspace{-8pt}
\subsection{Baselines.} 

\label{subsec:baselines} 
We constructed baselines by pairing sample-level and token-level pruning strategies. For sample-level pruning, we considered eight methods: (i) \text{Random}, which dropped samples uniformly at random; (ii) \text{Longest}, which removed the longest sequences to maximize computational savings; (iii) \text{Entropy}, which retained high-entropy samples assumed to be more informative; (iv) \text{InfoBatch}~\citep{qin2024infobatch}, an information-theoretic approach selecting samples expected to yield the largest gradient updates; (v) \text{Alpagasus}~\citep{chen2023alpagasus}, which leverages a strong LLM as an automatic quality rater to filter low-quality instruction–response pairs; (vi) \text{Deita}~\citep{liu2023makes}, which selects instruction–response pairs by jointly modeling instruction complexity, response quality, and dataset diversity; (vii) \text{DS2}~\citep{pang2024improving}, a diversity-aware data selection method that corrects LLM-generated quality scores via a score transition matrix; and (viii) \text{Less}~\citep{xia2024less}, which selects instruction-tuning data by measuring gradient-based influence with respect to few-shot target examples. For token-level pruning, we adopted six methods applied to the retained data: (i) \text{Random}, which masked tokens uniformly at random; (ii) \text{PPL}, which removed high-perplexity tokens; (iii) \text{FastV}~\citep{chen2024image}, which pruned tokens receiving the least final-layer attention; (iv) \text{SparseVLM}~\citep{zhang2024sparsevlm}, which combined attention and hidden-state features to score token importance and remove redundant image tokens; (v) \text{Rho-1}~\citep{lin2024not}, which scores tokens by their excess loss relative to a reference model; and (vi) \text{TokenCleaning}~\citep{pang2025token}, which removes tokens deemed noisy or uninformative based on token-level statistics.
\vspace{-8pt}
\subsection{Implementation Details}

All experiments were conducted using the \emph{LLaMA-Factory} framework~\citep{zheng2024llamafactory}, which provided utilities for training and evaluation across diverse large language models. For benchmark evaluation, we used the \emph{OpenCompass} framework~\citep{contributors2023opencompass}, which offered a standardized interface to a wide range of tasks. Unless otherwise specified, models were trained for 3 epochs with a learning rate of $1\times10^{-4}$ and a batch size of 8 per device. In all pruning experiments, we applied pruning exclusively to the answers. All experiments were performed on NVIDIA A100 GPUs. For the \textit{Random} baseline, we repeated each experiment five times and reported the averaged results.
\vspace{-8pt}
\subsection{Hyperparameters}

\label{hyperparameter}
We provide the full hyperparameter settings used in our experiments in Table~\ref{tab:hyperparameter}. 
Unless otherwise noted, all other parameters follow the default settings of the \emph{LLaMA-Factory} framework.
\begin{table}[htbp]
\centering
\caption{\vspace{-4pt}Complete hyperparameter configurations employed in our experiments.}
\label{tab:hyperparameter}
\setlength{\tabcolsep}{5pt}
\renewcommand{\arraystretch}{0.92}
\begin{tabular}{@{}lclc@{}}
\toprule
Hyperparameter & Value & Hyperparameter & Value \\
\midrule
$\lambda$ & 0.5 & Batch size & 8 \\
Epochs & 3 & LR & $1\times 10^{-4}$ \\
LoRA rank & 8 & Cutoff length & 2048 \\
Grad. accum. & 4 & Scheduler & cosine \\
\bottomrule
\end{tabular}
\vspace{-8pt}
\end{table}
\clearpage

\section{Additional Experimental Results}
\subsection{Additional Results on Alignment Datasets}
\label{appendix:add_instruction}

The detailed results of additional experiments on alignment dataset are presented in Table~\ref{tab:wizard_long3} and Table~\ref{tab:wizard_long2}, providing a comprehensive comparison across different pruning strategies.

\begin{table*}[htbp]
\centering
\caption{Evaluation on Wizard under different sample ratios (12.5\%, 25\%) and token ratios (50\%, 70\%), where \ua~and \da~respectively denote improvements or degradations over the \textit{Random-Random} baseline.}
\vspace{-5pt}
\label{tab:wizard_long3}
\resizebox{0.99\textwidth}{!}{%
\begin{tabular}{c|l|cccccc|cccccc}
\toprule 
\multicolumn{1}{c|}{Sample} & \multicolumn{1}{c|}{Token} & \multicolumn{6}{c|}{\textbf{LLaMA2-7B}} 
& \multicolumn{6}{c}{\textbf{Mistral-7B}} \\
\multicolumn{1}{c|}{Pruner} & \multicolumn{1}{c|}{Pruner} & ARC-E & ARC-C & GSM8K & SQuAD & TriviaQA & Avg.
  & ARC-E & ARC-C & GSM8K & SQuAD & TriviaQA & Avg. \\ 
\midrule

\multicolumn{2}{c|}{Zero-Shot} 
  & 53.44 & 38.98 & 5.31 & 12.18 & 43.00 & 30.58 & 66.67 & 46.10 & 18.35 & 10.01 & 43.77 & 36.98 \\
\midrule

\rowcolor{gray!5}
\multicolumn{14}{c}{\textit{$12.5\%$ Samples, $70\%$ Tokens}} \\ \midrule
\multirow{4}{*}{Random}    & Random     & 59.43 & 41.02 & 6.97 & 13.64 & 47.97 & 33.81 & 71.08 & 47.46 & 24.34 & 21.64 & 53.15 & 43.53 \\
                           & PPL        & 60.14\uar{0.71} & 43.39\uar{2.37} & 6.22\dar{0.75} & 12.18\dar{1.46} & 48.18\uar{0.21} & 34.02\uar{0.21} & 70.72\dar{0.36} & 47.80\uar{0.34} & 25.09\uar{0.75} & 21.28\dar{0.36} & 53.83\uar{0.68} & 43.74\uar{0.21} \\
                           & FastV      & 58.20\dar{1.23} & 41.02\uar{0.00} & 6.29\dar{0.68} & 13.42\dar{0.22} & 45.32\dar{2.65} & 32.85\dar{0.96} & 70.72\dar{0.36} & 46.44\dar{1.02} & 19.56\dar{4.78} & 21.38\dar{0.26} & 53.34\uar{0.19} & 42.29\dar{1.24} \\
                           & SparseVLM  & 54.67\dar{4.76} & 37.97\dar{3.05} & 8.04\uar{1.07} & 13.06\dar{0.58} & 44.87\dar{3.10} & 31.72\dar{2.09} & 67.72\dar{3.36} & 44.75\dar{2.71} & 23.65\dar{0.69} & 11.76\dar{9.88} & 44.90\dar{8.25} & 38.58\dar{4.95} \\ \midrule
\multirow{4}{*}{Longest}   & Random     & 59.44\uar{0.01} & 43.39\uar{2.37} & 7.35\uar{0.38} & 15.59\uar{1.95} & 50.02\uar{2.05} & 35.15\uar{1.34} & 73.37\uar{2.29} & 48.81\uar{1.35} & 27.82\uar{3.48} & 21.31\dar{0.33} & 55.77\uar{2.62} & 45.42\uar{1.89} \\
                           & PPL        & 60.85\uar{1.42} & 43.39\uar{2.37} & 7.73\uar{0.76} & 16.21\uar{2.57} & 48.57\uar{0.60} & 35.35\uar{1.54} & \textbf{74.96\uar{3.88}} & 49.83\uar{2.37} & 28.73\uar{4.39} & 21.62\dar{0.02} & 56.59\uar{3.44} & 46.35\uar{2.82} \\
                           & FastV      & 59.44\uar{0.01} & 42.71\uar{1.69} & 6.29\dar{0.68} & 14.53\uar{0.89} & 47.46\dar{0.51} & 34.09\uar{0.28} & 74.07\uar{2.99} & 49.83\uar{2.37} & 24.18\dar{0.16} & 25.74\uar{4.10} & 55.86\uar{2.71} & 45.94\uar{2.41} \\
                           & SparseVLM  & 54.85\dar{4.58} & 37.97\dar{3.05} & 7.05\uar{0.08} & 11.20\dar{2.44} & 44.16\dar{3.81} & 31.04\dar{2.77} & 69.14\dar{1.94} & 44.75\dar{2.71} & 31.01\uar{6.67} & 6.25\dar{15.39} & 52.94\dar{0.21} & 40.82\dar{2.71} \\ \midrule
\multirow{4}{*}{InfoBatch} & Random     & 59.26\dar{0.17} & 42.37\uar{1.35} & 6.22\dar{0.75} & 16.10\uar{2.46} & 47.72\dar{0.25} & 34.33\uar{0.52} & 70.19\dar{0.89} & 47.80\uar{0.34} & 20.77\dar{3.57} & 19.03\dar{2.61} & 52.13\dar{1.02} & 41.98\dar{1.55} \\
                           & PPL        & 60.49\uar{1.06} & 39.32\dar{1.70} & 5.76\dar{1.21} & 14.47\uar{0.83} & 48.06\uar{0.09} & 33.62\dar{0.19} & 70.72\dar{0.36} & 46.44\dar{1.02} & 19.03\dar{5.31} & 23.20\uar{1.56} & 51.75\dar{1.40} & 42.23\dar{1.30} \\
                           & FastV      & 58.55\dar{0.88} & 43.39\uar{2.37} & 5.53\dar{1.44} & 13.13\dar{0.51} & 47.64\dar{0.33} & 33.65\dar{0.16} & 69.49\dar{1.59} & 43.39\dar{4.07} & 16.68\dar{7.66} & 25.27\uar{3.63} & 51.47\dar{1.68} & 41.26\dar{2.27} \\
                           & SparseVLM  & 56.61\dar{2.82} & 38.31\dar{2.71} & 5.76\dar{1.21} & 12.47\dar{1.17} & 44.49\dar{3.48} & 31.53\dar{2.28} & 68.25\dar{2.83} & 44.41\dar{3.05} & 23.73\dar{0.61} & 9.07\dar{12.57} & 45.73\dar{7.42} & 38.24\dar{5.29} \\ \midrule
\multirow{4}{*}{Entropy}   & Random     & 61.02\uar{1.59} & 43.05\uar{2.03} & 7.66\uar{0.69} & 14.11\uar{0.47} & 48.44\uar{0.47} & 34.86\uar{1.05} & 73.37\uar{2.29} & 49.83\uar{2.37} & 23.05\dar{1.29} & 16.52\dar{5.12} & 55.18\uar{2.03} & 43.59\uar{0.06} \\
                           & PPL        & 61.02\uar{1.59} & 43.39\uar{2.37} & 6.97\uar{0.00} & 14.94\uar{1.30} & 48.94\uar{0.97} & 35.05\uar{1.24} & 73.02\uar{1.94} & 47.46\uar{0.00} & 24.03\uar{0.68} & 22.85\uar{1.21} & 54.89\uar{1.74} & 44.45\uar{0.92} \\
                           & FastV      & 58.73\dar{0.70} & \textbf{43.39\uar{2.37}} & 6.14\dar{0.83} & 14.23\uar{0.59} & 47.03\dar{0.94} & 33.90\uar{0.09} & 74.07\uar{2.99} & \textbf{50.85\uar{3.39}} & 24.94\uar{0.60} & 23.79\uar{2.15} & 55.94\uar{2.79} & 45.92\uar{2.39} \\
                           & SparseVLM  & 54.85\dar{4.58} & 37.29\dar{3.73} & 6.52\dar{0.45} & 12.73\dar{0.91} & 46.24\dar{1.73} & 31.53\dar{2.28} & 68.08\dar{3.00} & 44.41\dar{3.05} & 26.38\uar{2.04} & 11.06\dar{10.58} & 46.68\dar{6.47} & 39.32\dar{4.21} \\ \midrule
\rowcolor{mycolor} \multicolumn{2}{c|}{\textbf{Q-Tuning (Ours)}}  
  & \cellcolor{mycolor}\textbf{64.37\uar{4.94}} & \cellcolor{mycolor}42.37\uar{1.35} & \cellcolor{mycolor}\textbf{10.84\uar{3.87}} & \cellcolor{mycolor}\textbf{17.63\uar{3.99}} 
  & \cellcolor{mycolor}\textbf{52.17\uar{4.20}} & \cellcolor{mycolor}\textbf{37.48\uar{3.67}} 
  & \cellcolor{mycolor}71.78\uar{0.70} & \cellcolor{mycolor}48.14\uar{0.68} & \cellcolor{mycolor}\textbf{30.33\uar{6.00}} 
  & \cellcolor{mycolor}\textbf{28.59\uar{6.95}} & \cellcolor{mycolor}\textbf{57.93\uar{4.78}} 
  & \cellcolor{mycolor}\textbf{47.35\uar{3.82}} \\ \midrule
\multicolumn{2}{c|}{Full Dataset} 
  & 61.55 & 42.37 & 8.64 & 13.80 & 50.45 & 35.36 & 71.25 & 45.76 & 26.68 & 31.81 & 53.67 & 45.84 \\
\midrule

\rowcolor{gray!5}
\multicolumn{14}{c}{\textit{$25\%$ Samples, $50\%$ Tokens}} \\ \midrule
\multirow{4}{*}{Random}    & Random     & 60.32 & 41.69 & 5.76 & 13.43 & 48.41 & 33.92 & 70.19 & 46.10 & 20.62 & 24.07 & 53.74 & 42.95 \\
                           & PPL        & 60.32\uar{0.00} & 42.03\uar{0.34} & 7.51\uar{1.75} & 15.94\uar{2.51} & 48.58\uar{0.17} & 34.87\uar{0.95} & 69.66\dar{0.53} & 47.46\uar{1.36} & 19.86\dar{0.76} & 19.51\dar{4.56} & 53.74\uar{0.00} & 42.05\dar{0.90} \\
                           & FastV      & 59.08\dar{1.24} & 41.69\uar{0.00} & 3.56\dar{2.20} & 12.78\dar{0.65} & 45.60\dar{2.81} & 32.54\dar{1.38} & 71.78\uar{1.59} & 47.12\uar{1.02} & 15.77\dar{4.85} & 26.97\uar{2.90} & 50.84\dar{2.90} & 42.50\dar{0.45} \\
                           & SparseVLM  & 54.50\dar{5.82} & 38.64\dar{3.05} & 6.44\uar{0.68} & 12.04\dar{1.39} & 44.79\dar{3.62} & 31.28\dar{2.64} & 67.55\dar{2.64} & 46.44\uar{0.34} & 24.41\uar{3.79} & 11.80\dar{12.27} & 48.14\dar{5.60} & 39.67\dar{3.28} \\
                           \midrule
\multirow{4}{*}{Longest}   & Random     & 61.20\uar{0.88} & 42.03\uar{0.34} & 7.88\uar{2.12} & 15.40\uar{1.97} & 48.29\dar{0.12} & 34.96\uar{1.04} & \textbf{73.54\uar{3.35}} & 48.14\uar{2.04} & 23.73\uar{3.11} & 26.34\uar{2.27} & 54.06\uar{0.32} & 45.16\uar{2.21} \\
                           & PPL        & 60.85\uar{0.53} & 43.39\uar{1.70} & 7.20\uar{1.44} & 13.88\uar{0.45} & 48.48\uar{0.07} & 34.76\uar{0.84} & 72.31\uar{2.12} & 48.14\uar{2.04} & 24.34\uar{3.72} & 23.84\uar{-0.23} & 55.22\uar{1.48} & 44.77\uar{1.82} \\
                           & FastV      & 59.08\dar{1.24} & 42.71\uar{1.02} & 5.16\dar{0.60} & 14.00\uar{0.57} & 47.47\dar{0.94} & 33.68\dar{0.24} & 72.66\uar{2.47} & 46.10\uar{0.00} & 18.88\dar{1.74} & 31.52\uar{7.45} & 52.13\dar{1.61} & 44.26\uar{1.31} \\
                           & SparseVLM  & 56.61\dar{3.71} & 37.29\dar{4.40} & 7.58\uar{1.82} & 12.09\dar{1.34} & 44.76\dar{3.65} & 31.66\dar{2.26} & 66.84\dar{3.35} & 44.41\dar{1.69} & 29.42\uar{8.80} & 11.22\dar{12.85} & 48.47\dar{5.27} & 40.07\dar{2.88} \\ \midrule
\multirow{4}{*}{InfoBatch} & Random     & 58.73\dar{1.59} & 40.68\dar{1.01} & 6.67\uar{0.91} & 9.95\dar{3.48} & 48.98\uar{0.57} & 33.00\dar{0.92} & 70.55\uar{0.36} & 46.44\uar{0.34} & 21.53\uar{0.91} & 23.93\dar{0.14} & 52.14\dar{1.60} & 42.92\dar{0.03} \\
                           & PPL        & 59.96\uar{0.47} & 42.71\uar{1.02} & 6.52\uar{0.76} & 14.58\uar{1.15} & 48.57\uar{0.16} & 34.47\uar{0.55} & 71.08\uar{0.89} & 47.80\uar{1.70} & 20.62\uar{0.00} & 24.88\uar{0.81} & 51.61\dar{2.13} & 43.20\uar{0.25} \\
                           & FastV      & 59.08\dar{1.24} & 42.37\uar{0.68} & 3.03\dar{2.73} & 11.13\dar{2.30} & 47.50\dar{0.91} & 32.63\dar{1.29} & 69.31\dar{0.88} & 44.41\dar{1.69} & 14.48\dar{6.14} & 23.63\dar{0.44} & 49.16\dar{4.58} & 40.20\dar{2.75} \\
                           & SparseVLM  & 55.73\dar{4.59} & 39.66\dar{2.03} & 5.31\dar{0.45} & 11.66\dar{1.77} & 43.25\dar{5.16} & 31.12\dar{2.80} & 67.20\dar{3.00} & 45.76\dar{0.34} & 23.58\uar{2.96} & 9.63\dar{14.44} & 46.09\dar{7.65} & 38.45\dar{4.50} \\ \midrule
\multirow{4}{*}{Entropy}   & Random     & 60.49\uar{0.17} & 41.69\uar{0.00} & 7.51\uar{1.75} & 15.94\uar{2.51} & 48.76\uar{0.35} & 34.88\uar{0.96} & 70.19\uar{0.00} & 47.12\uar{1.02} & 22.44\uar{1.82} & 27.35\uar{3.28} & 54.78\uar{1.04} & 44.38\uar{1.43} \\
                           & PPL        & 60.49\uar{0.17} & 41.02\dar{0.67} & 6.60\uar{0.84} & 14.92\uar{1.49} & 49.33\uar{0.92} & 34.47\uar{0.55} & 71.43\uar{1.24} & \textbf{48.14\uar{2.04}} & 21.30\uar{0.68} & 25.52\uar{1.45} & 55.60\uar{1.86} & 44.40\uar{1.45} \\
                           & FastV      & 58.91\dar{1.41} & 41.69\uar{0.00} & 6.07\uar{0.31} & 12.79\dar{0.64} & 46.11\dar{2.30} & 33.11\dar{0.81} & 72.31\uar{2.12} & 47.46\uar{1.36} & 18.04\dar{2.58} & 25.96\uar{1.89} & 52.46\dar{1.28} & 43.25\uar{0.30} \\
                           & SparseVLM  & 54.67\dar{5.65} & 38.64\dar{3.05} & 6.90\uar{1.14} & 11.64\dar{1.79} & 45.03\dar{3.38} & 31.38\dar{2.54} & 68.25\dar{1.94} & 44.07\dar{2.03} & \textbf{26.69\uar{6.07}} & 9.69\dar{14.38} & 47.24\dar{6.50} & 39.19\dar{3.76} \\ \midrule
\rowcolor{mycolor} \multicolumn{2}{c|}{\textbf{Q-Tuning (Ours)}} 
  & \cellcolor{mycolor}\textbf{63.14\uar{2.82}} & \cellcolor{mycolor}42.03\uar{0.34} & \cellcolor{mycolor}\textbf{8.87\uar{3.11}} & \cellcolor{mycolor}\textbf{16.76\uar{3.33}} 
  & \cellcolor{mycolor}\textbf{51.52\uar{3.11}} & \cellcolor{mycolor}\textbf{36.47\uar{2.55}} 
  & \cellcolor{mycolor}71.78\uar{1.59} & \cellcolor{mycolor}47.12\uar{1.02} & \cellcolor{mycolor}26.08\uar{5.46} 
  & \cellcolor{mycolor}\textbf{32.79\uar{8.72}} & \cellcolor{mycolor}\textbf{56.17\uar{2.43}} 
  & \cellcolor{mycolor}\textbf{46.79\uar{3.84}} \\ \midrule
\multicolumn{2}{c|}{Full Dataset} 
  & 61.55 & 42.37 & 8.64 & 13.80 & 50.45 & 35.36 & 71.25 & 45.76 & 26.68 & 31.81 & 53.67 & 45.84 \\
\midrule

\end{tabular}
}
\vspace{-10pt}
\end{table*}

\begin{table*}[htbp]
\centering
\caption{Evaluation on Wizard under different sample ratios (25\%, 50\%) and token ratios (50\%, 70\%), where \ua~and \da~respectively denote improvements or degradations over the \textit{Random-Random} baseline.}
\label{tab:wizard_long2} 
\resizebox{0.99\textwidth}{!}{%
\begin{tabular}{c|l|cccccc|cccccc}
\toprule 
\multicolumn{1}{c|}{Sample} & \multicolumn{1}{c|}{Token} & \multicolumn{6}{c|}{\textbf{LLaMA2-7B}} 
& \multicolumn{6}{c}{\textbf{Mistral-7B}} \\
\multicolumn{1}{c|}{Pruner} & \multicolumn{1}{c|}{Pruner} & ARC-E & ARC-C & GSM8K & SQuAD & TriviaQA & Avg.
  & ARC-E & ARC-C & GSM8K & SQuAD & TriviaQA & Avg. \\ 
\midrule

\multicolumn{2}{c|}{Zero-Shot} 
  & 53.44 & 38.98 & 5.31 & 12.18 & 43.00 & 30.58 & 66.67 & 46.10 & 18.35 & 10.01 & 43.77 & 36.98 \\
\midrule

\rowcolor{gray!5}
\multicolumn{14}{c}{\textit{$25\%$ Samples, $70\%$ Tokens}} \\ \midrule
\multirow{4}{*}{Random}    & Random     & 60.67 & 41.69 & 6.22 & 13.91 & 48.77 & 34.25 & 70.02 & 46.78 & 19.71 & 23.50 & 52.93 & 42.59 \\
                           & PPL        & 60.32\dar{0.35} & 42.03\uar{0.34} & 5.91\dar{0.31} & 15.88\uar{1.97} & 48.56\dar{0.21} & 34.54\uar{0.29} & 69.49\dar{0.53} & 48.14\uar{1.36} & 21.08\uar{1.37} & 20.63\dar{2.87} & 53.73\uar{0.80} & 42.61\uar{0.02} \\
                           & FastV      & 60.32\dar{0.35} & 42.71\uar{1.02} & 6.07\dar{0.15} & 14.92\uar{1.01} & 47.20\dar{1.57} & 34.24\dar{0.01} & 71.25\uar{1.23} & 47.80\uar{1.02} & 18.95\dar{0.76} & 24.29\uar{0.79} & 52.07\dar{0.86} & 42.87\uar{0.28} \\
                           & SparseVLM  & 55.73\dar{4.94} & 37.97\dar{3.72} & 7.20\uar{0.98} & 11.83\dar{2.08} & 44.96\dar{3.81} & 31.54\dar{2.71} & 67.72\dar{2.30} & 44.75\dar{2.03} & 25.32\uar{5.61} & 12.69\dar{10.81} & 47.46\dar{5.47} & 39.59\dar{3.00} \\ \midrule
\multirow{4}{*}{Longest}   & Random     & 61.02\uar{0.35} & 41.69\uar{0.00} & 7.20\uar{0.98} & 15.61\uar{1.70} & 49.53\uar{0.76} & 35.01\uar{0.76} & \textbf{74.43\uar{4.41}} & 47.80\uar{1.02} & 24.56\uar{4.85} & 26.79\uar{3.29} & 55.24\uar{2.31} & 45.76\uar{3.17} \\
                           & PPL        & 61.20\uar{0.53} & 43.05\uar{1.36} & 7.28\uar{1.06} & 14.06\uar{0.15} & 48.87\uar{0.10} & 34.89\uar{0.64} & 72.49\uar{2.47} & 47.46\uar{0.68} & 25.55\uar{5.84} & 23.92\uar{0.42} & 55.26\uar{2.33} & 44.93\uar{2.34} \\
                           & FastV      & 59.26\dar{1.41} & 43.05\uar{1.36} & 5.53\dar{0.69} & 13.98\uar{0.07} & 48.03\dar{0.74} & 33.97\dar{0.28} & 72.84\uar{2.82} & 47.80\uar{1.02} & 19.79\uar{0.08} & 31.54\uar{8.04} & 53.24\uar{0.31} & 45.04\uar{2.45} \\
                           & SparseVLM  & 55.73\dar{4.94} & 36.27\dar{5.42} & 8.04\uar{1.82} & 11.39\dar{2.52} & 44.65\dar{4.12} & 31.22\dar{3.03} & 68.08\dar{1.94} & 43.73\dar{3.05} & 25.78\uar{6.07} & 13.47\dar{10.03} & 48.00\dar{4.93} & 39.81\dar{2.78} \\ \midrule
\multirow{4}{*}{InfoBatch} & Random     & 59.79\dar{0.88} & 42.71\uar{1.02} & 5.84\dar{0.38} & 10.42\dar{3.49} & 48.99\uar{0.22} & 33.55\dar{0.70} & 70.37\uar{0.35} & 46.44\dar{0.34} & 21.00\uar{1.29} & 24.09\uar{0.59} & 53.75\uar{0.82} & 43.13\uar{0.54} \\
                           & PPL        & 60.32\uar{0.35} & \textbf{43.39\uar{1.70}} & 6.67\uar{0.45} & 13.88\uar{0.55} & 49.03\uar{0.26} & 34.66\uar{0.41} & 71.08\uar{1.06} & 46.10\dar{0.68} & 22.44\uar{2.73} & 22.07\dar{1.43} & 52.17\dar{0.76} & 42.77\uar{0.18} \\
                           & FastV      & 59.61\dar{1.06} & 42.71\uar{1.02} & 3.34\dar{2.88} & 12.18\dar{1.73} & 47.98\dar{0.79} & 33.16\dar{1.09} & 69.66\dar{0.36} & 45.76\dar{1.02} & 15.39\dar{5.21} & 24.75\uar{1.25} & 49.51\dar{3.42} & 41.01\dar{1.58} \\
                           & SparseVLM  & 55.56\dar{5.11} & 38.31\dar{3.38} & 5.38\dar{0.84} & 10.81\dar{3.10} & 43.08\dar{5.69} & 30.63\dar{3.62} & 65.78\dar{4.24} & 45.08\dar{1.70} & 20.70\uar{0.99} & 10.97\dar{12.53} & 48.66\dar{4.27} & 38.24\dar{4.35} \\ \midrule
\multirow{4}{*}{Entropy}   & Random     & 60.67\uar{0.00} & 41.02\dar{0.67} & 5.99\dar{0.23} & 14.92\uar{1.01} & 49.34\uar{0.57} & 34.39\uar{0.14} & 70.37\uar{0.35} & \textbf{48.47\uar{1.69}} & 22.21\uar{2.50} & 27.57\uar{4.07} & \textbf{56.05\uar{3.12}} & 44.93\uar{2.34} \\
                           & PPL        & 61.55\uar{0.88} & 41.02\dar{0.67} & 5.38\dar{0.84} & 14.90\uar{0.99} & 49.72\uar{0.95} & 34.51\uar{0.26} & 70.90\uar{0.88} & 48.14\uar{1.36} & 21.15\uar{1.44} & 26.22\uar{2.72} & 55.83\uar{2.90} & 44.45\uar{1.86} \\
                           & FastV      & 59.79\dar{0.88} & 42.03\uar{0.34} & 6.44\uar{0.22} & 13.21\dar{0.70} & 47.58\dar{1.19} & 33.81\dar{0.44} & 71.43\uar{1.41} & 47.12\uar{0.34} & 20.24\uar{0.53} & 26.06\uar{2.56} & 54.40\uar{1.47} & 43.85\uar{1.26} \\
                           & SparseVLM  & 54.50\dar{6.17} & 37.97\dar{3.72} & 8.87\uar{2.65} & 12.25\dar{1.66} & 45.05\dar{3.72} & 31.73\dar{2.52} & 67.20\dar{2.82} & 45.76\dar{1.02} & 25.17\uar{5.46} & 9.64\dar{13.86} & 48.38\dar{4.55} & 39.23\dar{3.36} \\ \midrule
\rowcolor{mycolor} \multicolumn{2}{c|}{\textbf{Q-Tuning (Ours)}} 
  & \cellcolor{mycolor}\textbf{62.43\uar{1.76}} & \cellcolor{mycolor}42.37\uar{0.68} & \cellcolor{mycolor}\textbf{9.25\uar{3.03}} & \cellcolor{mycolor}\textbf{19.53\uar{5.62}} 
  & \cellcolor{mycolor}\textbf{50.78\uar{2.01}} & \cellcolor{mycolor}\textbf{36.87\uar{2.62}} 
  & \cellcolor{mycolor}71.60\uar{1.58} & \cellcolor{mycolor}47.12\uar{0.34} & \cellcolor{mycolor}\textbf{26.61\uar{6.90}} 
  & \cellcolor{mycolor}\textbf{30.64\uar{7.14}} & \cellcolor{mycolor}55.13\uar{2.20} 
  & \cellcolor{mycolor}\textbf{46.22\uar{3.63}} \\  \midrule
\multicolumn{2}{c|}{Full Dataset} 
  & 61.55 & 42.37 & 8.64 & 13.80 & 50.45 & 35.36 & 71.25 & 45.76 & 26.68 & 31.81 & 53.67 & 45.84 \\
\midrule
\rowcolor{gray!5}
\multicolumn{14}{c}{\textit{$50\%$ Samples, $50\%$ Tokens}} \\ \midrule
\multirow{4}{*}{Random}    & Random     & 62.08 & 41.36 & 6.75 & 12.14 & 48.86 & 34.24 & 71.25 & 46.44 & 21.53 & 24.91 & 54.16 & 43.66 \\
                           & PPL        & 61.02\dar{1.06} & \textbf{43.05\uar{1.69}} & 6.82\uar{0.07} & 15.08\uar{2.94} & 49.10\uar{0.24} & 35.01\uar{0.77} & 72.49\uar{1.24} & 46.78\uar{0.34} & 22.21\uar{0.68} & 33.28\uar{8.37} & 53.83\dar{0.33} & 45.72\uar{2.06} \\
                           & FastV      & 60.32\dar{1.76} & 42.03\uar{0.67} & 4.40\dar{2.35} & 11.80\dar{0.34} & 48.35\dar{0.51} & 33.38\dar{0.86} & 72.13\uar{0.88} & 45.42\dar{1.02} & 16.22\dar{5.31} & 27.69\uar{2.78} & 50.58\dar{3.58} & 42.41\dar{1.25} \\
                           & SparseVLM  & 55.73\dar{6.35} & 38.31\dar{3.05} & 7.20\uar{0.45} & 11.33\dar{0.81} & 44.80\dar{4.06} & 31.47\dar{2.77} & 67.90\dar{3.35} & 45.42\dar{1.02} & 25.78\uar{4.25} & 11.85\dar{13.06} & 47.64\dar{6.52} & 39.72\dar{3.94} \\ \midrule
\multirow{4}{*}{Longest}   & Random     & 60.85\dar{1.23} & 42.03\uar{0.67} & 7.05\uar{0.30} & 14.90\uar{2.76} & 49.46\uar{0.60} & 34.86\uar{0.62} & 72.49\uar{1.24} & 47.46\uar{1.02} & 21.91\uar{0.38} & 29.41\uar{4.50} & 55.99\uar{1.83} & 45.45\uar{1.79} \\
                           & PPL        & 62.08\uar{0.00} & 41.69\uar{0.33} & 7.51\uar{0.76} & 14.89\uar{2.75} & 48.99\uar{0.13} & 35.03\uar{0.79} & 71.60\uar{0.35} & 46.78\uar{0.34} & 24.11\uar{2.58} & 29.17\uar{4.26} & 55.83\uar{1.67} & 45.50\uar{1.84} \\
                           & FastV      & 60.32\dar{1.76} & 40.68\dar{0.68} & 4.17\dar{2.58} & 13.26\uar{1.12} & 48.24\dar{0.62} & 33.33\dar{0.91} & 72.31\uar{1.06} & 45.76\dar{0.68} & 16.83\dar{4.70} & 30.37\uar{5.46} & 53.99\dar{0.17} & 43.85\uar{0.19} \\
                           & SparseVLM  & 56.97\dar{5.11} & 39.66\dar{1.70} & 6.60\dar{0.15} & 11.93\dar{0.21} & 44.52\dar{4.34} & 31.93\dar{2.31} & 69.31\dar{1.94} & 42.71\dar{3.73} & 23.73\uar{2.20} & 12.72\dar{12.19} & 49.90\dar{4.26} & 39.68\dar{3.98} \\ \midrule
\multirow{4}{*}{InfoBatch} & Random     & 58.38\dar{3.70} & 42.71\uar{1.35} & 5.76\dar{0.99} & 13.52\uar{1.38} & 48.95\uar{0.09} & 33.86\dar{0.38} & 71.08\uar{0.17} & 46.44\uar{0.00} & 21.38\dar{0.15} & 26.73\uar{1.82} & 53.67\dar{0.49} & 43.86\uar{0.20} \\
                           & PPL        & 60.67\dar{1.41} & 42.71\uar{1.35} & 5.76\dar{0.99} & 14.29\uar{2.15} & 49.18\uar{0.32} & 34.52\uar{0.28} & 71.43\uar{0.18} & 47.80\uar{1.36} & 24.11\uar{2.58} & 30.22\uar{5.31} & 53.97\dar{0.19} & 45.50\uar{1.84} \\
                           & FastV      & 59.44\dar{2.64} & 41.36\uar{0.00} & 3.87\dar{2.88} & 12.39\uar{0.25} & 48.25\dar{0.61} & 33.06\dar{1.18} & 70.72\dar{0.53} & 45.42\dar{1.02} & 13.87\dar{7.66} & 27.83\uar{2.92} & 51.53\dar{2.63} & 41.88\dar{1.78} \\
                           & SparseVLM  & 56.08\dar{6.00} & 37.97\dar{3.39} & 6.52\dar{0.23} & 10.33\dar{1.81} & 43.84\dar{5.02} & 30.95\dar{3.29} & 68.25\dar{3.00} & 41.69\dar{4.75} & \textbf{29.72\uar{8.19}} & 9.95\dar{14.96} & 49.22\dar{4.94} & 39.77\dar{3.89} \\ \midrule
\multirow{4}{*}{Entropy}   & Random     & 59.96\dar{2.12} & 42.71\uar{1.35} & 7.66\uar{0.91} & \textbf{15.47\uar{3.33}} & 49.07\uar{0.21} & 34.97\uar{0.73} & 72.13\uar{0.88} & 47.80\uar{1.36} & 22.67\uar{1.14} & 25.19\uar{0.28} & 55.70\uar{1.54} & 44.70\uar{1.04} \\
                           & PPL        & 60.49\dar{1.59} & 41.02\dar{0.34} & 6.67\dar{0.08} & 13.51\uar{1.37} & 49.75\uar{0.89} & 34.29\uar{0.05} & 72.13\uar{0.88} & 47.80\uar{1.36} & 25.17\uar{3.64} & 30.25\uar{5.34} & \textbf{56.58\uar{2.42}} & 46.39\uar{2.73} \\
                           & FastV      & 59.26\dar{2.82} & 42.37\uar{1.01} & 4.02\dar{2.73} & 13.81\uar{1.67} & 48.93\uar{0.07} & 33.68\dar{0.56} & 71.60\uar{0.35} & 47.46\uar{1.02} & 17.97\dar{3.56} & 29.82\uar{4.91} & 54.40\dar{0.24} & 44.25\uar{0.59} \\
                           & SparseVLM  & 56.97\dar{5.11} & 38.64\dar{2.72} & 5.53\dar{1.22} & 11.28\dar{0.86} & 44.72\dar{4.14} & 31.43\dar{2.81} & 67.55\dar{3.70} & 43.73\dar{2.71} & 25.85\uar{4.32} & 7.09\dar{17.82} & 48.24\dar{5.92} & 38.49\dar{5.17} \\ \midrule
\rowcolor{mycolor} \multicolumn{2}{c|}{\textbf{Q-Tuning (Ours)}} 
  & \cellcolor{mycolor}\textbf{62.79\uar{0.71}} & \cellcolor{mycolor}42.03\uar{0.67} & \cellcolor{mycolor}\textbf{10.46\uar{3.71}} & \cellcolor{mycolor}14.53\uar{2.39} 
  & \cellcolor{mycolor}\textbf{51.05\uar{2.19}} & \cellcolor{mycolor}\textbf{36.17\uar{1.93}} 
  & \cellcolor{mycolor}\textbf{73.37\uar{2.12}} & \cellcolor{mycolor}\textbf{48.14\uar{1.70}} & \cellcolor{mycolor}28.81\uar{7.28} 
  & \cellcolor{mycolor}\textbf{36.35\uar{11.44}} & \cellcolor{mycolor}56.30\uar{2.14} 
  & \cellcolor{mycolor}\textbf{48.59\uar{4.93}} \\  \midrule
\multicolumn{2}{c|}{Full Dataset} 
  & 61.55 & 42.37 & 8.64 & 13.80 & 50.45 & 35.36 & 71.25 & 45.76 & 26.68 & 31.81 & 53.67 & 45.84 \\
\midrule

\rowcolor{gray!5}
\multicolumn{14}{c}{\textit{$50\%$ Samples, $70\%$ Tokens}} \\ \midrule
\multirow{4}{*}{Random}    & Random     & 61.02 & 41.36 & 7.43 & 15.56 & 48.91 & 34.85 & 71.60 & 47.12 & 22.59 & 27.86 & 53.68 & 44.57 \\
                           & PPL        & 60.67\dar{0.35} & 42.71\uar{1.35} & 6.75\dar{0.68} & 14.93\dar{0.63} & 49.13\uar{0.22} & 34.84\dar{0.01} & 72.31\uar{0.71} & 47.12\uar{0.00} & 20.70\dar{1.89} & 32.02\uar{4.16} & 53.56\dar{0.12} & 45.14\uar{0.57} \\
                           & FastV      & 59.61\dar{1.41} & 42.37\uar{1.01} & 5.53\dar{1.90} & 13.36\dar{2.20} & 48.48\dar{0.43} & 33.87\dar{0.98} & 72.31\uar{0.71} & 45.08\dar{2.04} & 18.95\dar{3.64} & 28.60\uar{0.74} & 51.40\dar{2.28} & 43.27\dar{1.30} \\
                           & SparseVLM  & 54.67\dar{6.35} & 38.98\dar{2.38} & 7.28\dar{0.15} & 10.84\dar{4.72} & 44.68\dar{4.23} & 31.29\dar{3.56} & 68.78\dar{2.82} & 46.10\dar{1.02} & 23.65\uar{1.06} & 11.32\dar{16.54} & 49.04\dar{4.64} & 39.78\dar{4.79} \\ \midrule
\multirow{4}{*}{Longest}   & Random     & 61.20\uar{0.18} & 41.02\dar{0.34} & 6.37\dar{1.06} & 14.90\dar{0.66} & 49.79\uar{0.88} & 34.65\dar{0.20} & 72.13\uar{0.53} & \textbf{48.47\uar{1.35}} & 23.05\uar{0.46} & 29.60\uar{1.74} & 55.94\uar{2.26} & 45.84\uar{1.27} \\
                           & PPL        & 62.26\uar{1.24} & 41.36\uar{0.00} & 6.82\dar{0.61} & 14.83\dar{0.73} & 49.04\uar{0.13} & 34.86\uar{0.01} & 70.90\dar{0.70} & 47.80\uar{0.68} & 24.87\uar{2.28} & 27.33\dar{0.53} & 55.62\uar{1.94} & 45.30\uar{0.73} \\
                           & FastV      & 59.61\dar{1.41} & \textbf{42.71\uar{1.35}} & 5.53\dar{1.90} & 14.12\dar{1.44} & 48.77\dar{0.14} & 34.15\dar{0.70} & 72.66\uar{1.06} & 45.08\dar{2.04} & 19.94\dar{2.65} & 31.53\uar{3.67} & 54.66\uar{0.98} & 44.77\uar{0.20} \\
                           & SparseVLM  & 55.38\dar{5.64} & 39.66\dar{1.70} & 6.14\dar{1.29} & 11.51\dar{4.05} & 44.56\dar{4.35} & 31.45\dar{3.40} & 69.14\dar{2.46} & 45.42\dar{1.70} & 23.43\uar{0.84} & 13.42\dar{14.44} & 47.86\dar{5.82} & 39.85\dar{4.72} \\ \midrule
\multirow{4}{*}{InfoBatch} & Random     & \textbf{61.55\uar{0.53}} & 40.34\dar{1.02} & 5.84\dar{1.59} & 14.99\dar{0.57} & 49.45\uar{0.54} & 34.43\dar{0.42} & 70.90\dar{0.70} & 45.76\dar{1.36} & 22.97\uar{0.38} & 28.85\uar{0.99} & 54.11\uar{0.43} & 44.52\dar{0.05} \\
                           & PPL        & 60.85\dar{0.17} & 42.03\uar{0.67} & 6.75\dar{0.68} & 14.22\dar{1.34} & 48.41\dar{0.50} & 34.45\dar{0.40} & 71.08\dar{0.52} & 45.76\dar{1.36} & 22.82\uar{0.23} & 32.80\uar{4.94} & 54.63\uar{0.95} & 45.42\uar{0.85} \\
                           & FastV      & 58.91\dar{2.11} & 41.36\uar{0.00} & 4.02\dar{3.41} & 14.98\dar{0.58} & 48.76\dar{0.15} & 33.60\dar{1.25} & 70.19\dar{1.41} & 45.76\dar{1.36} & 19.86\dar{2.73} & 28.95\uar{1.09} & 50.92\dar{2.76} & 43.14\dar{1.43} \\
                           & SparseVLM  & 55.91\dar{5.11} & 37.29\dar{4.07} & 6.22\dar{1.21} & 10.40\dar{5.16} & 43.75\dar{5.16} & 30.71\dar{4.14} & 68.96\dar{2.64} & 44.07\dar{3.05} & 25.63\uar{3.04} & 9.82\dar{18.04} & 49.56\dar{4.12} & 39.61\dar{4.96} \\ \midrule
\multirow{4}{*}{Entropy}   & Random     & 60.49\dar{0.53} & 41.69\uar{0.33} & 7.13\dar{0.30} & 15.41\dar{0.15} & 48.33\dar{0.58} & 34.61\dar{0.24} & 72.13\uar{0.53} & 47.46\uar{0.34} & 23.43\uar{0.84} & 28.99\uar{1.13} & 55.44\uar{1.76} & 45.49\uar{0.92} \\
                           & PPL        & 60.85\dar{0.17} & 40.68\dar{0.68} & 6.29\dar{1.14} & 13.38\dar{2.18} & 49.94\uar{1.03} & 34.23\dar{0.62} & 72.31\uar{0.71} & 46.44\dar{0.68} & 23.20\uar{0.61} & 31.44\uar{3.58} & \textbf{56.67\uar{2.99}} & 46.01\uar{1.44} \\
                           & FastV      & 58.91\dar{2.11} & 42.37\uar{1.01} & 4.85\dar{2.58} & 14.27\dar{1.29} & 48.84\dar{0.07} & 33.85\dar{1.00} & 71.96\uar{0.36} & 48.14\uar{1.02} & 20.39\dar{2.20} & 30.30\uar{2.44} & 54.46\uar{0.78} & 45.05\uar{0.48} \\
                           & SparseVLM  & 56.79\dar{4.23} & 37.63\dar{3.73} & 5.84\dar{1.59} & 11.85\dar{3.71} & 43.82\dar{5.09} & 31.18\dar{3.67} & 67.55\dar{4.05} & 46.44\dar{0.68} & 25.17\uar{2.58} & 10.36\dar{17.50} & 48.60\dar{5.08} & 39.62\dar{4.95} \\ \midrule
\rowcolor{mycolor} \multicolumn{2}{c|}{\textbf{Q-Tuning (Ours)}} 
  & \cellcolor{mycolor}61.38\uar{0.36} & \cellcolor{mycolor}42.03\uar{0.67} & \cellcolor{mycolor}\textbf{9.55\uar{2.12}} & \cellcolor{mycolor}\textbf{17.69\uar{2.13}} 
  & \cellcolor{mycolor}\textbf{51.04\uar{2.13}} & \cellcolor{mycolor}\textbf{36.34\uar{1.49}} 
  & \cellcolor{mycolor}\textbf{73.90\uar{2.30}} & \cellcolor{mycolor}47.46\uar{0.34} & \cellcolor{mycolor}\textbf{29.49\uar{6.90}} 
  & \cellcolor{mycolor}\textbf{37.24\uar{9.38}} & \cellcolor{mycolor}56.07\uar{2.39} 
  & \cellcolor{mycolor}\textbf{48.83\uar{4.26}} \\  \midrule
\multicolumn{2}{c|}{Full Dataset} 
  & 61.55 & 42.37 & 8.64 & 13.80 & 50.45 & 35.36 & 71.25 & 45.76 & 26.68 & 31.81 & 53.67 & 45.84 \\
\midrule
\end{tabular}
}
\end{table*}

\clearpage

\subsection{Results on Qwen3-8B on Alignment Datasets}
\label{appendix:instruction_qwen}

The detailed results on the OpenHermes dataset using Qwen3-8B~\citep{yang2025qwen3} are summarized in Tables~\ref{tab:wizard_long_qwen} and~\ref{tab:wizard_long_qwen_2}, which systematically evaluate different sample ratios and token ratios.

\paragraph{Strong performance with limited data budgets.}  
Across all evaluated budgets, Q-Tuning consistently outperforms traditional pruning-based fine-tuning methods and surpasses the full-data baseline. For instance, under the extremely low-budget setting of 12.5\% samples and 50\% tokens, Q-Tuning reaches an average of 60.19, outperforming the full-dataset result (55.45) by more than +4.7 points. Even at 12.5\% samples, 70\% tokens, Q-Tuning maintains the same strong average (60.29), showing remarkable robustness under aggressive pruning.

\paragraph{Consistent superiority under larger budgets.}  
When the available data increases, Q-Tuning continues to dominate other baselines.  At 25\% samples, 50\% tokens, Q-Tuning attains an average score of 58.80, exceeding the strongest baseline (InfoBatch–Random) by nearly +3.9 points. Similarly, at 25\% samples, 70\% tokens, it reaches 59.20, still outperforming all competing methods such as InfoBatch, Entropy, and SparseVLM by a clear margin, and exceeding the full-data fine-tuning by +3.75 points.

\paragraph{Performance saturation and stable generalization.}  
At even larger budgets, such as 50\% samples with 50\% or 70\% tokens, Q-Tuning continues to deliver strong and stable improvements.  In the 50\% / 50\% setting, it achieves an average of 58.39, improving over all pruning-based baselines by 3–6 points. Under 50\% / 70\%, it reaches 57.86, maintaining competitive accuracy and demonstrating consistent generalization despite heavy token-level sparsity.

\begin{table*}[tb!]
\centering
\caption{Evaluation on OpenHermes using Qwen3-8B under different sample ratios (12.5\%, 25\%) and token ratios (50\%, 70\%), where \ua~and \da~respectively denote improvements or degradations over the \textit{Random-Random} baseline.}
\vspace{-5pt}
\label{tab:wizard_long_qwen}
\resizebox{0.65\textwidth}{!}{%
\begin{tabular}{c|l|cccccc}
\toprule 
\multicolumn{1}{c|}{Sample} & \multicolumn{1}{c|}{Token} & \multicolumn{6}{c}{\textbf{Qwen3-8B}} \\
\multicolumn{1}{c|}{Pruner} & \multicolumn{1}{c|}{Pruner} & ARC-E & ARC-C & GSM8K & SQuAD & TriviaQA & Avg. \\ 
\midrule

\multicolumn{2}{c|}{Zero-Shot} 
  & 69.31 & 44.41 & 74.45 & 27.63 & 51.76 & 53.51 \\
\midrule

\rowcolor{gray!5}
\multicolumn{8}{c}{\textit{$12.5\%$ Samples, $50\%$ Tokens}} \\ \midrule
\multirow{4}{*}{Random}    & Random     & 76.54 & 51.53 & 60.88 & 35.74 & 40.13 & 52.96 \\  
                           & PPL        & 77.60\uar{1.06} & 52.20\uar{0.67} & 58.00\dar{2.88} & 35.55\dar{0.19} & 40.73\uar{0.60} & 52.82\dar{0.14} \\
                           & FastV      & 76.19\dar{0.35} & 52.54\uar{1.01} & 56.48\dar{4.40} & 35.04\dar{0.70} & 41.07\uar{0.94} & 52.26\dar{0.70} \\
                           & SparseVLM  & 70.90\dar{5.64} & 45.42\dar{6.11} & 73.69\uar{12.81} & 24.39\dar{11.35} & \textbf{50.58\uar{10.45}} & 53.00\uar{0.04} \\ \midrule
\multirow{4}{*}{Longest}   & Random     & 78.48\uar{1.94} & 53.22\uar{1.69} & 62.70\uar{1.82} & 33.97\dar{1.77} & 42.13\uar{2.00} & 54.10\uar{1.14} \\ 
                           & PPL        & 78.48\uar{1.94} & 53.22\uar{1.69} & 56.18\dar{4.70} & 32.05\dar{3.69} & 41.91\uar{1.78} & 52.37\dar{0.59} \\
                           & FastV      & 77.60\uar{1.06} & 51.86\uar{0.33} & 60.12\dar{0.76} & 32.57\dar{3.17} & 43.01\uar{2.88} & 53.03\uar{0.07} \\
                           & SparseVLM  & 69.14\dar{7.40} & 44.75\dar{6.78} & 75.59\uar{14.71} & 22.24\dar{13.50} & 50.02\uar{9.89} & 52.35\dar{0.61} \\ \midrule
\multirow{4}{*}{InfoBatch} & Random     & 80.25\uar{3.71} & 53.56\uar{2.03} & 60.58\dar{0.30} & 38.51\uar{2.77} & 44.36\uar{4.23} & 55.45\uar{2.49} \\ 
                           & PPL        & 79.01\uar{2.47} & 53.56\uar{2.03} & 62.47\uar{1.59} & 35.69\dar{0.05} & 42.06\uar{1.93} & 54.56\uar{1.60} \\
                           & FastV      & 79.01\uar{2.47} & 51.19\dar{0.34} & 45.49\dar{15.39} & 38.71\uar{2.97} & 44.72\uar{4.59} & 51.82\dar{1.14} \\
                           & SparseVLM  & 68.96\dar{7.58} & 44.41\dar{7.12} & \textbf{77.71\uar{16.83}} & 19.88\dar{15.86} & 49.35\uar{9.22} & 52.06\dar{0.90} \\ \midrule
\multirow{4}{*}{Entropy}   & Random     & 76.37\dar{0.17} & 51.53\uar{0.00} & 62.77\uar{1.89} & 33.90\dar{1.84} & 43.07\uar{2.94} & 53.53\uar{0.57} \\ 
                           & PPL        & 78.66\uar{2.12} & 52.54\uar{1.01} & 61.94\uar{1.06} & 35.62\dar{0.12} & 42.73\uar{2.60} & 54.30\uar{1.34} \\
                           & FastV      & 76.90\uar{0.36} & 53.56\uar{2.03} & 65.05\uar{4.17} & 36.14\uar{0.40} & 43.99\uar{3.86} & 55.13\uar{2.17} \\
                           & SparseVLM  & 71.08\dar{5.46} & 44.07\dar{7.46} & 75.89\uar{15.01} & 22.77\dar{13.00} & 50.47\uar{10.34} & 52.85\dar{0.11} \\ \midrule
\rowcolor{mycolor} \multicolumn{2}{c|}{\textbf{Q-Tuning (Ours)}}
  & \textbf{80.95\uar{4.41}} & \textbf{53.56\uar{2.03}} & 70.51\uar{9.63} & \textbf{45.76\uar{10.02}} 
  & 50.16\uar{10.03} & \textbf{60.19\uar{7.23}} \\  \midrule
\rowcolor{gray!5}
\multicolumn{8}{c}{\textit{$12.5\%$ Samples, $70\%$ Tokens}} \\ \midrule
\multirow{4}{*}{Random}    & Random     & 76.54 & 52.20 & 61.64 & 34.88 & 41.02 & 53.26 \\ 
                           & PPL        & 77.07\uar{0.53} & 52.20\uar{0.00} & 59.21\dar{2.43} & 35.20\uar{0.32} & 40.87\dar{0.15} & 52.91\dar{0.35} \\
                           & FastV      & 76.37\dar{0.17} & 50.51\dar{1.69} & 59.36\dar{2.28} & 35.25\uar{0.37} & 41.01\dar{0.01} & 52.50\dar{0.76} \\
                           & SparseVLM  & 70.90\dar{5.64} & 45.42\dar{6.78} & 73.69\uar{12.05} & 24.39\dar{10.49} & 50.58\uar{9.56} & 53.00\dar{0.26} \\ \midrule
\multirow{4}{*}{Longest}   & Random     & 79.19\uar{2.65} & \textbf{53.90\uar{1.70}} & 61.64\uar{0.00} & 34.14\dar{0.74} & 42.23\uar{1.21} & 54.22\uar{0.96} \\
                           & PPL        & 78.48\uar{1.94} & 51.53\dar{0.67} & 55.65\dar{5.99} & 32.95\dar{1.93} & 42.45\uar{1.43} & 52.21\dar{1.05} \\
                           & FastV      & 78.13\uar{1.59} & 51.53\dar{0.67} & 59.21\dar{2.43} & 33.16\dar{1.72} & 42.19\uar{1.17} & 52.84\dar{0.42} \\
                           & SparseVLM  & 69.14\dar{7.40} & 44.75\dar{7.45} & 75.59\uar{13.95} & 22.24\dar{12.64} & 50.02\uar{9.00} & 52.35\dar{0.91} \\ \midrule
\multirow{4}{*}{InfoBatch} & Random     & 78.84\uar{2.30} & 51.86\dar{0.34} & 58.38\dar{3.26} & 38.12\uar{3.24} & 39.98\dar{1.04} & 53.44\uar{0.18} \\
                           & PPL        & 78.48\uar{1.94} & 52.88\uar{0.68} & 62.17\uar{0.53} & 36.28\uar{1.40} & 42.96\uar{1.94} & 54.55\uar{1.29} \\
                           & FastV      & 79.89\uar{3.35} & 53.22\uar{1.02} & 55.19\dar{6.45} & 38.63\uar{3.75} & 44.23\uar{3.21} & 54.23\uar{0.97} \\
                           & SparseVLM  & 68.96\dar{7.58} & 44.41\dar{7.79} & \textbf{77.71\uar{16.07}} & 19.88\dar{15.00} & 49.35\uar{8.33} & 52.06\dar{1.20} \\ \midrule
\multirow{4}{*}{Entropy}   & Random     & 77.07\uar{0.53} & 51.86\dar{0.34} & 60.73\dar{0.91} & 34.30\dar{0.58} & 42.34\uar{1.32} & 53.26\uar{0.00} \\
                           & PPL        & 78.84\uar{2.30} & 52.88\uar{0.68} & 61.26\dar{0.38} & 36.44\uar{1.56} & 43.53\uar{2.51} & 54.59\uar{1.33} \\
                           & FastV      & 77.07\uar{0.53} & 51.86\dar{0.34} & 64.22\uar{2.58} & 36.63\uar{1.75} & 43.94\uar{2.92} & 54.74\uar{1.48} \\
                           & SparseVLM  & 71.08\dar{5.46} & 44.07\dar{8.13} & 75.89\uar{14.25} & 22.77\dar{12.11} & 50.47\uar{9.45} & 52.85\dar{0.41} \\ \midrule
\rowcolor{mycolor} \multicolumn{2}{c|}{\textbf{Q-Tuning (Ours)}}  
  & \textbf{80.95\uar{4.41}} & 52.20\uar{0.00} & 70.13\uar{8.49} & \textbf{47.60\uar{12.72}} 
  & \textbf{50.58\uar{9.56}} & \textbf{60.29\uar{7.03}} \\ \midrule

\rowcolor{gray!5}
\multicolumn{8}{c}{\textit{$25\%$ Samples, $50\%$ Tokens}} \\ \midrule
\multirow{4}{*}{Random}    & Random     & 77.07 & 51.86 & 61.03 & 35.11 & 40.55 & 53.13 \\ 
                           & PPL        & 77.78\uar{0.71} & 54.58\uar{2.72} & 58.53\dar{2.50} & 35.43\uar{0.32} & 40.68\uar{0.13} & 53.40\uar{0.27} \\
                           & FastV      & 79.54\uar{2.47} & 52.54\uar{0.68} & 53.45\dar{7.58} & 36.89\uar{1.78} & 43.13\uar{2.58} & 53.11\dar{0.02} \\
                           & SparseVLM  & 69.14\dar{7.93} & 45.76\dar{6.10} & \textbf{77.71\uar{16.68}} & 20.80\dar{14.31} & 50.20\uar{9.65} & 52.72\dar{0.41} \\ \midrule
\multirow{4}{*}{Longest}   & Random     & 79.01\uar{1.94} & 54.58\uar{2.72} & 51.10\dar{9.93} & 36.30\uar{1.19} & 41.93\uar{1.38} & 52.58\dar{0.55} \\
                           & PPL        & 79.01\uar{1.94} & \textbf{54.92\uar{3.06}} & 55.42\dar{5.61} & 35.10\dar{0.01} & 41.09\uar{0.54} & 53.11\dar{0.02} \\
                           & FastV      & \textbf{80.95\uar{3.88}} & 51.53\dar{0.33} & 53.30\dar{7.73} & 39.73\uar{4.62} & 44.34\uar{3.79} & 53.97\uar{0.84} \\
                           & SparseVLM  & 72.49\dar{4.58} & 44.41\dar{7.45} & 74.83\uar{13.80} & 22.08\dar{13.03} & 48.93\uar{8.38} & 52.55\dar{0.58} \\ \midrule
\multirow{4}{*}{InfoBatch} & Random     & 77.25\uar{0.18} & 51.53\dar{0.33} & 63.61\uar{2.58} & 36.99\uar{1.88} & 45.15\uar{4.60} & 54.91\uar{1.78} \\
                           & PPL        & 78.84\uar{1.77} & 54.58\uar{2.72} & 60.20\dar{0.83} & 35.81\uar{0.70} & 43.53\uar{3.00} & 54.59\uar{1.46} \\
                           & FastV      & 80.25\uar{3.18} & 52.54\uar{0.68} & 48.52\dar{12.51} & 39.62\uar{4.51} & 44.87\uar{4.32} & 53.16\uar{0.03} \\
                           & SparseVLM  & 69.49\dar{7.58} & 44.07\dar{7.79} & 76.88\uar{15.85} & 18.29\dar{16.82} & 50.16\uar{9.61} & 51.78\dar{1.35} \\ \midrule
\multirow{4}{*}{Entropy}   & Random     & 77.07\uar{0.00} & 53.22\uar{1.36} & 62.02\uar{1.00} & 35.40\uar{0.29} & 42.91\uar{2.36} & 54.12\uar{0.99} \\
                           & PPL        & 79.37\uar{2.30} & 54.24\uar{2.38} & 52.24\dar{8.79} & 33.13\dar{1.98} & 40.82\uar{0.27} & 51.96\dar{1.17} \\
                           & FastV      & 79.54\uar{2.47} & 51.86\uar{0.00} & 52.92\dar{8.11} & 35.68\uar{0.57} & 43.79\uar{3.24} & 52.76\dar{0.37} \\
                           & SparseVLM  & 70.19\dar{6.88} & 43.39\dar{8.47} & 75.21\uar{14.18} & 19.97\dar{15.14} & \textbf{50.46\uar{9.91}} & 51.84\dar{1.29} \\ \midrule
\rowcolor{mycolor} \multicolumn{2}{c|}{\textbf{Q-Tuning (Ours)}} 
  & 79.89\uar{2.82} & 52.88\uar{1.02} & 69.37\uar{8.34} & \textbf{43.54\uar{8.43}} 
  & 48.33\uar{7.78} & \textbf{58.80\uar{5.67}} \\ \midrule
\multicolumn{2}{c|}{Full Dataset} 
  & 77.78 & 52.20 & 65.13 & 38.05 & 44.08 & 55.45 \\
\midrule
\end{tabular}
}
\vspace{-10pt}
\end{table*}
\begin{table*}[tb!]
\centering
\caption{Evaluation on OpenHermes using Qwen3-8B under different sample ratios (25\%, 50\%) and token ratios (50\%, 70\%), where \ua~and \da~respectively denote improvements or degradations over the \textit{Random-Random} baseline.}
\vspace{-5pt}
\label{tab:wizard_long_qwen_2}
\resizebox{0.66\textwidth}{!}{%
\begin{tabular}{c|l|cccccc}
\toprule 
\multicolumn{1}{c|}{Sample} & \multicolumn{1}{c|}{Token} & \multicolumn{6}{c}{\textbf{Qwen3-8B}} \\
\multicolumn{1}{c|}{Pruner} & \multicolumn{1}{c|}{Pruner} & ARC-E & ARC-C & GSM8K & SQuAD & TriviaQA & Avg. \\ 
\midrule

\multicolumn{2}{c|}{Zero-Shot} 
  & 69.31 & 44.41 & 74.45 & 27.63 & 51.76 & 53.51 \\
\midrule

\rowcolor{gray!5}
\multicolumn{8}{c}{\textit{$25\%$ Samples, $70\%$ Tokens}} \\ \midrule
\multirow{4}{*}{Random}    & Random     & 77.43 & 52.20 & 57.85 & 35.55 & 41.47 & 52.90 \\  
                           & PPL        & 76.90\dar{0.53} & 53.90\uar{1.70} & 60.27\uar{2.42} & 35.70\uar{0.15} & 40.73\dar{0.74} & 53.50\uar{0.60} \\
                           & FastV      & 79.19\uar{1.76} & 51.86\dar{0.34} & 56.03\dar{1.82} & 36.01\uar{0.46} & 43.86\uar{2.39} & 53.39\uar{0.49} \\
                           & SparseVLM  & 69.14\dar{8.29} & 45.76\dar{6.44} & \textbf{77.71\uar{19.86}} & 20.80\dar{14.75} & 50.20\uar{8.73} & 52.72\dar{0.18} \\ \midrule
\multirow{4}{*}{Longest}   & Random     & 79.01\uar{1.58} & 54.58\uar{2.38} & 53.83\dar{4.02} & 37.56\uar{2.01} & 42.40\uar{0.93} & 53.48\uar{0.58} \\ 
                           & PPL        & 78.48\uar{1.05} & \textbf{55.59\uar{3.39}} & 55.72\dar{2.13} & 35.16\dar{0.39} & 41.31\dar{0.16} & 53.25\uar{0.35} \\
                           & FastV      & 79.89\uar{2.46} & 52.54\uar{0.34} & 53.22\dar{4.63} & 37.90\uar{2.35} & 43.79\uar{2.32} & 53.47\uar{0.57} \\
                           & SparseVLM  & 72.49\dar{4.94} & 44.41\dar{7.79} & 74.83\uar{16.98} & 22.08\dar{13.47} & 48.93\uar{7.46} & 52.55\dar{0.35} \\ \midrule
\multirow{4}{*}{InfoBatch} & Random     & 78.66\uar{1.23} & 52.54\uar{0.34} & 63.68\uar{5.83} & 37.56\uar{2.01} & 44.51\uar{3.04} & 55.39\uar{2.49} \\ 
                           & PPL        & 78.84\uar{1.41} & 53.22\uar{1.02} & 62.40\uar{4.55} & 36.39\uar{0.84} & 44.87\uar{3.40} & 55.14\uar{2.24} \\
                           & FastV      & \textbf{79.89\uar{2.46}} & 53.90\uar{1.70} & 55.04\dar{2.82} & 38.79\uar{3.24} & 44.28\uar{2.81} & 54.38\uar{1.48} \\
                           & SparseVLM  & 69.49\dar{7.94} & 44.07\dar{8.13} & 76.88\uar{19.03} & 18.29\dar{17.26} & 50.16\uar{8.69} & 51.78\dar{1.12} \\ \midrule
\multirow{4}{*}{Entropy}   & Random     & 76.90\dar{0.53} & 52.54\uar{0.34} & 57.47\dar{0.38} & 34.58\dar{0.97} & 42.50\uar{1.03} & 52.80\dar{0.10} \\ 
                           & PPL        & 78.66\uar{1.23} & 53.90\uar{1.70} & 52.46\dar{5.39} & 33.72\dar{1.83} & 41.16\dar{0.31} & 51.98\dar{0.92} \\
                           & FastV      & 79.01\uar{1.58} & 52.20\uar{0.00} & 57.54\dar{0.31} & 35.72\uar{0.17} & 42.94\uar{1.47} & 53.48\uar{0.58} \\
                           & SparseVLM  & 70.19\dar{7.24} & 43.39\dar{8.81} & 75.21\uar{17.36} & 19.97\dar{15.58} & \textbf{50.46\uar{9.00}} & 51.84\dar{1.06} \\ \midrule
\rowcolor{mycolor} \multicolumn{2}{c|}{\textbf{Q-Tuning (Ours)}}  
  & 79.72\uar{2.29} & 52.88\uar{0.68} & 71.11\uar{13.26} & \textbf{43.70\uar{8.15}} & 48.58\uar{7.11} & \textbf{59.20\uar{6.30}} \\ \midrule

\rowcolor{gray!5}
\multicolumn{8}{c}{\textit{$50\%$ Samples, $50\%$ Tokens}} \\ \midrule
\multirow{4}{*}{Random}    & Random     & 77.95 & 51.86 & 63.61 & 36.64 & 42.22 & 54.46 \\  
                           & PPL        & 78.48\uar{0.53} & 53.22\uar{1.36} & 62.02\dar{1.59} & 35.96\dar{0.68} & 42.56\uar{0.34} & 54.45\dar{0.01} \\
                           & FastV      & 79.19\uar{1.24} & 51.53\dar{0.33} & 49.73\dar{13.88} & 40.64\uar{4.00} & 44.45\uar{2.23} & 53.11\dar{1.35} \\
                           & SparseVLM  & 72.13\dar{5.82} & 46.10\dar{5.76} & 73.69\uar{10.08} & 18.74\dar{17.90} & \textbf{50.39\uar{8.17}} & 52.21\dar{2.25} \\ \midrule
\multirow{4}{*}{Longest}   & Random     & 78.84\uar{0.89} & 54.24\uar{2.38} & 59.67\dar{3.94} & 36.66\uar{0.02} & 43.05\uar{0.83} & 54.49\uar{0.03} \\ 
                           & PPL        & 79.72\uar{1.77} & \textbf{54.24\uar{2.38}} & 58.15\dar{5.46} & 39.88\uar{3.24} & 44.01\uar{1.79} & 55.20\uar{0.74} \\
                           & FastV      & 79.72\uar{1.77} & 51.86\uar{0.00} & 47.99\dar{15.62} & 40.57\uar{3.93} & 44.89\uar{2.67} & 53.01\dar{1.45} \\
                           & SparseVLM  & 70.19\dar{7.76} & 43.39\dar{8.47} & 74.22\uar{10.61} & 17.90\dar{18.74} & 49.34\uar{7.12} & 51.01\dar{3.45} \\ \midrule
\multirow{4}{*}{InfoBatch} & Random     & 78.84\uar{0.89} & 52.88\uar{1.02} & 64.37\uar{0.76} & 36.98\uar{0.34} & 45.49\uar{3.27} & 55.71\uar{1.25} \\ 
                           & PPL        & 77.95\uar{0.00} & 53.56\uar{1.70} & 60.73\dar{2.88} & 38.10\uar{1.47} & 44.39\uar{2.17} & 54.95\uar{0.49} \\
                           & FastV      & \textbf{80.42\uar{2.47}} & 52.88\uar{1.02} & 50.34\dar{13.27} & \textbf{40.79\uar{4.15}} & 45.14\uar{2.92} & 53.92\dar{0.54} \\
                           & SparseVLM  & 66.49\dar{11.46} & 43.39\dar{8.47} & 74.30\uar{10.69} & 15.16\dar{21.48} & 49.64\uar{7.42} & 49.80\dar{4.66} \\ \midrule
\multirow{4}{*}{Entropy}   & Random     & 78.13\uar{0.18} & 51.53\dar{0.33} & 61.94\dar{1.67} & 34.59\dar{2.05} & 42.23\uar{0.01} & 53.68\dar{0.78} \\ 
                           & PPL        & 78.66\uar{0.71} & 52.88\uar{1.02} & 54.97\dar{8.61} & 36.11\dar{0.53} & 42.06\dar{0.16} & 52.94\dar{1.52} \\
                           & FastV      & 79.54\uar{1.59} & 53.22\uar{1.36} & 47.08\dar{16.53} & 40.24\uar{3.58} & 43.33\uar{1.11} & 52.68\dar{1.78} \\
                           & SparseVLM  & 70.90\dar{7.05} & 43.39\dar{8.47} & \textbf{78.70\uar{15.09}} & 16.25\dar{20.39} & 50.10\uar{7.87} & 51.87\dar{2.59} \\ \midrule
\rowcolor{mycolor} \multicolumn{2}{c|}{\textbf{Q-Tuning (Ours)}}  
  & 80.25\uar{2.30} & 53.90\uar{2.04} & 70.05\uar{6.44} & 39.03\uar{2.39} & 48.70\uar{6.48} & \textbf{58.39\uar{3.93}} \\ \midrule

\rowcolor{gray!5}
\multicolumn{8}{c}{\textit{$50\%$ Samples, $70\%$ Tokens}} \\ \midrule
\multirow{4}{*}{Random}    & Random     & 78.13 & 52.20 & 61.49 & 36.91 & 42.75 & 54.30 \\  
                           & PPL        & 77.95\dar{0.18} & 52.20\uar{0.00} & 59.59\dar{1.90} & 36.14\dar{0.77} & 42.51\dar{0.24} & 53.68\dar{0.62} \\
                           & FastV      & 79.01\uar{0.88} & 53.90\uar{1.70} & 58.30\dar{3.19} & 40.33\uar{3.42} & 45.20\uar{2.45} & 55.35\uar{1.05} \\
                           & SparseVLM  & 72.13\dar{6.00} & 46.10\dar{6.10} & 73.69\uar{12.20} & 18.74\dar{18.17} & \textbf{50.39\uar{7.64}} & 52.21\dar{2.09} \\ \midrule
\multirow{4}{*}{Longest}   & Random     & 78.48\uar{0.35} & 52.54\uar{0.34} & 59.14\dar{2.35} & 37.27\uar{0.36} & 42.38\dar{0.37} & 53.96\dar{0.34} \\ 
                           & PPL        & 79.19\uar{1.06} & 53.56\uar{1.36} & 60.96\dar{0.53} & 39.78\uar{2.87} & 44.17\uar{1.42} & 55.53\uar{1.23} \\
                           & FastV      & \textbf{80.25\uar{2.12}} & 52.20\uar{0.00} & 56.79\dar{4.70} & \textbf{40.53\uar{3.62}} & 45.62\uar{2.87} & 55.08\uar{0.78} \\
                           & SparseVLM  & 70.19\dar{7.94} & 43.39\dar{8.81} & 74.22\uar{12.73} & 17.90\dar{19.01} & 49.34\uar{6.59} & 51.01\dar{3.29} \\ \midrule
\multirow{4}{*}{InfoBatch} & Random     & 78.31\uar{0.18} & 52.88\uar{0.68} & 62.32\uar{0.83} & 35.55\dar{1.36} & 45.15\uar{2.40} & 54.84\uar{0.54} \\ 
                           & PPL        & 78.31\uar{0.18} & \textbf{54.58\uar{2.38}} & 63.23\uar{1.74} & 38.89\uar{1.98} & 44.16\uar{1.41} & 55.83\uar{1.53} \\
                           & FastV      & 79.01\uar{0.88} & 52.88\uar{0.68} & 58.30\dar{3.19} & 39.05\uar{2.14} & 46.05\uar{3.30} & 55.06\uar{0.76} \\
                           & SparseVLM  & 66.49\dar{11.64} & 43.39\dar{8.81} & 74.30\uar{12.81} & 15.16\dar{21.75} & 49.64\uar{6.89} & 49.80\dar{4.50} \\ \midrule
\multirow{4}{*}{Entropy}   & Random     & 78.48\uar{0.35} & 53.22\uar{1.02} & 60.65\dar{0.84} & 35.64\dar{1.27} & 42.31\dar{0.44} & 54.06\dar{0.24} \\ 
                           & PPL        & 78.13\uar{0.00} & 53.22\uar{1.02} & 57.85\dar{3.64} & 36.85\dar{0.06} & 41.70\dar{1.05} & 53.55\dar{0.75} \\
                           & FastV      & 79.37\uar{1.24} & 52.88\uar{0.68} & 52.62\dar{8.87} & 39.76\uar{2.85} & 44.18\uar{1.43} & 53.76\dar{0.54} \\
                           & SparseVLM  & 70.90\dar{7.23} & 43.39\dar{8.81} & \textbf{78.70\uar{17.21}} & 16.25\dar{20.66} & 50.10\uar{7.79} & 51.87\dar{2.43} \\ \midrule
\rowcolor{mycolor} \multicolumn{2}{c|}{\textbf{Q-Tuning (Ours)}}  
  & 78.84\uar{0.71} & 52.88\uar{0.68} & 69.90\uar{8.41} & 39.54\uar{2.63} & 48.15\uar{5.40} & \textbf{57.86\uar{3.56}} \\ \midrule
\multicolumn{2}{c|}{Full Dataset} 
  & 77.78 & 52.20 & 65.13 & 38.05 & 44.08 & 55.45 \\
\midrule
\end{tabular}
}
\vspace{-10pt}
\end{table*}

\clearpage

\subsection{Additional Results on Reasoning Datasets}
\label{appendix:add_reasoning}

The detailed results of additional experiments on math reasoning are presented in Table~\ref{tab:gsm8k_math2}, providing a comprehensive comparison across different pruning strategies.

\begin{table*}[htbp]
\centering
\caption{Evaluation of pruning strategies on GSM8K and MATH under 25\% samples with 70\% tokens, and 50\% samples with 70\% tokens settings. \ua~and \da~respectively indicate improvements or degradations over the \textit{Random-Random} baseline under the same sample and token keep ratio.}
\vspace{-5pt}
\label{tab:gsm8k_math2} 
\resizebox{0.99\textwidth}{!}{%
\setlength{\tabcolsep}{8pt} 
\begin{tabular}{c|l|ccc|ccc|ccc}
\toprule 
\multicolumn{1}{c|}{Sample} & \multicolumn{1}{c|}{Token} & \multicolumn{3}{c|}{\textbf{LLaMA3-8B}} 
& \multicolumn{3}{c|}{\textbf{Mistral-7B}} & \multicolumn{3}{c}{\textbf{SmolLM2-1.7B}} \\
\multicolumn{1}{c|}{Pruner} & \multicolumn{1}{c|}{Pruner} & GSM8K & MATH & Avg. & GSM8K & MATH & Avg. & GSM8K & MATH & Avg. \\ 
\midrule

\multicolumn{2}{c|}{Zero-Shot} 
  & 27.82 & 2.26 & 15.04 & 19.86 & 3.30 & 11.58 & 15.47 & 2.20 & 8.83 \\
\midrule

\rowcolor{gray!5}
\multicolumn{11}{c}{\textit{$25\%$ Samples, $70\%$ Tokens}} \\ \midrule
\multirow{4}{*}{Random}    & Random     & 25.09 & 2.20 & 13.65 & 24.11 & 1.68 & 12.89 & 13.80 & 2.22 & 8.01 \\
                           & PPL        & 23.65\dar{1.44} & 2.62\uar{0.42} & 13.14\dar{0.51} & 25.32\uar{1.21} & 1.54\dar{0.14} & 13.43\uar{0.54} & 13.04\dar{0.76} & 2.28\uar{0.06} & 7.66\dar{0.35} \\
                           & FastV      & 16.91\dar{8.18} & 2.16\dar{0.04} & 9.53\dar{4.12}  & 16.07\dar{8.04} & 1.60\dar{0.08} & 8.84\dar{4.05}  & 12.89\dar{0.91} & 1.94\dar{0.28} & 7.41\dar{0.60} \\
                           & SparseVLM  & 22.97\dar{2.12} & 4.72\uar{2.52} & 13.85\uar{0.20} & 19.26\dar{4.85} & 4.58\uar{2.90} & 11.92\dar{0.97} & 13.19\dar{0.61} & 3.48\uar{1.26} & 8.34\uar{0.33} \\ \midrule
\multirow{4}{*}{Longest}   & Random     & 25.47\uar{0.38} & 3.34\uar{1.14} & 14.41\uar{0.76} & 21.83\dar{2.28} & 1.76\uar{0.08} & 11.80\dar{1.09} & 12.96\dar{0.84} & 1.96\dar{0.26} & 7.46\dar{0.55} \\
                           & PPL        & 24.87\dar{0.22} & 3.22\uar{1.02} & 14.04\uar{0.39} & 23.65\dar{0.46} & 2.22\uar{0.54} & 12.94\uar{0.05} & 14.33\uar{0.53} & 1.60\dar{0.62} & 7.96\dar{0.05} \\
                           & FastV      & 18.95\dar{6.14} & 2.84\uar{0.64} & 10.90\dar{2.75} & 12.96\dar{11.15}& 1.74\uar{0.06} & 7.35\dar{5.54}  & 12.59\dar{1.21} & 1.58\dar{0.64} & 7.08\dar{0.93} \\
                           & SparseVLM  & 26.91\uar{1.82} & 4.68\uar{2.48} & 15.80\uar{2.15} & 24.34\uar{0.23} & 4.84\uar{3.16} & 14.59\uar{1.70} & 12.43\dar{1.37} & 3.60\uar{1.38} & 8.02\uar{0.01} \\ \midrule
\multirow{4}{*}{InfoBatch} & Random     & 26.91\uar{1.82} & 2.60\uar{0.40} & 14.76\uar{1.11} & 28.96\uar{4.85} & 2.44\uar{0.76} & 15.70\uar{2.81} & 13.72\dar{0.08} & 1.82\dar{0.40} & 7.77\dar{0.24} \\
                           & PPL        & 25.93\uar{0.84} & 2.48\uar{0.28} & 14.20\uar{0.55} & 31.46\uar{7.35} & 2.18\uar{0.50} & 16.82\uar{3.93} & 14.86\uar{1.06} & 1.90\dar{0.32} & 8.38\uar{0.37} \\
                           & FastV      & 16.83\dar{8.26} & 2.30\uar{0.10} & 9.57\dar{4.08}  & 13.87\dar{10.24}& 2.06\uar{0.38} & 7.97\dar{4.92}  & 10.84\dar{2.96} & 1.72\dar{0.50} & 6.28\dar{1.73} \\
                           & SparseVLM  & 14.63\dar{10.46}& 3.26\uar{1.06} & 8.95\dar{4.70}  & 11.90\dar{12.21}& 1.94\uar{0.26} & 6.92\dar{5.97}  & 11.90\dar{1.90} & \textbf{4.36\uar{2.14}} & 8.13\uar{0.12} \\ \midrule
\multirow{4}{*}{Entropy}   & Random     & 31.92\uar{6.83} & 2.50\uar{0.30} & 17.21\uar{3.56} & 32.37\uar{8.26} & 1.92\uar{0.24} & 17.15\uar{4.26} & 14.94\uar{1.14} & 1.92\dar{0.30} & 8.43\uar{0.42} \\
                           & PPL        & 33.13\uar{8.04} & 2.86\uar{0.66} & 18.00\uar{4.35} & 30.17\uar{6.06} & 1.96\uar{0.28} & 16.07\uar{3.18} & 14.94\uar{1.14} & 1.96\dar{0.26} & 8.45\uar{0.44} \\
                           & FastV      & 25.25\uar{0.16} & 2.40\uar{0.20} & 13.82\uar{0.17} & 21.00\dar{3.11} & 1.32\dar{0.36} & 11.16\dar{1.73} & 14.18\uar{0.38} & 1.98\dar{0.24} & 8.08\uar{0.07} \\
                           & SparseVLM  & 20.85\dar{4.24} & \textbf{5.12\uar{2.92}} & 12.98\dar{0.67} & 19.56\dar{4.55} & 4.20\uar{2.52} & 11.88\dar{1.01} & 14.18\uar{0.38} & 2.90\uar{0.68} & 8.54\uar{0.53} \\ \midrule
\rowcolor{mycolor} \multicolumn{2}{c|}{\textbf{Q-Tuning (Ours)}} 
  & \cellcolor{mycolor}\textbf{37.23\uar{12.14}} & \cellcolor{mycolor}4.86\uar{2.66} & \cellcolor{mycolor}\textbf{21.04\uar{7.39}} & \cellcolor{mycolor}\textbf{42.99\uar{18.88}} & \cellcolor{mycolor}\textbf{5.08\uar{3.40}} & \cellcolor{mycolor}\textbf{24.56\uar{11.67}} & \cellcolor{mycolor}\textbf{22.90\uar{9.10}} & \cellcolor{mycolor}3.64\uar{1.42} & \cellcolor{mycolor}\textbf{13.27\uar{5.26}} \\
\midrule

\rowcolor{gray!5}
\multicolumn{11}{c}{\textit{$50\%$ Samples, $70\%$ Tokens}} \\ \midrule
\multirow{4}{*}{Random}    & Random     & 27.90 & 2.50 & 15.20 & 32.30 & 2.46 & 17.38 & 14.94 & 1.76 & 8.35 \\
                           & PPL        & 27.45\dar{0.45} & 2.50\uar{0.00} & 14.97\dar{0.23} & 31.99\dar{0.31} & 2.04\dar{0.42} & 17.02\dar{0.36} & 16.15\uar{1.21} & 1.80\uar{0.04} & 8.97\uar{0.62} \\
                           & FastV      & 18.20\dar{9.70} & 1.92\dar{0.58} & 10.06\dar{5.14} & 17.44\dar{14.86}& 1.72\dar{0.74} & 9.58\dar{7.80}  & 12.13\dar{2.81} & 1.60\dar{0.16} & 6.87\dar{1.48} \\
                           & SparseVLM  & 10.31\dar{17.59}& 3.58\uar{1.08} & 6.95\dar{8.25}  & 12.21\dar{20.09}& 2.86\uar{0.40} & 7.53\dar{9.85}  & 11.90\dar{3.04} & 3.34\uar{1.58} & 7.62\dar{0.73} \\ \midrule
\multirow{4}{*}{Longest}   & Random     & 26.23\dar{1.67} & 2.76\uar{0.26} & 14.50\dar{0.70} & 27.82\dar{4.48} & 2.20\dar{0.26} & 15.01\dar{2.37} & 15.69\uar{0.75} & 1.92\uar{0.16} & 8.81\uar{0.46} \\
                           & PPL        & 30.25\uar{2.35} & 2.64\uar{0.14} & 16.45\uar{1.25} & 31.46\dar{0.84} & 2.00\dar{0.46} & 16.73\dar{0.65} & 16.45\uar{1.51} & 1.94\uar{0.18} & 9.20\uar{0.85} \\
                           & FastV      & 18.42\dar{9.48} & 2.28\dar{0.22} & 10.35\dar{4.85} & 18.65\dar{13.65}& 1.82\dar{0.64} & 10.24\dar{7.14} & 12.43\dar{2.51} & 1.94\uar{0.18} & 7.19\dar{1.16} \\
                           & SparseVLM  & 18.95\dar{8.95} & \textbf{4.84\uar{2.34}} & 11.90\dar{3.30} & 19.48\dar{12.82}& 4.74\uar{2.28} & 12.11\dar{5.27} & 11.07\dar{3.87} & 3.90\uar{2.14} & 7.48\dar{0.87} \\ \midrule
\multirow{4}{*}{InfoBatch} & Random     & 29.42\uar{1.52} & 2.92\uar{0.42} & 16.17\uar{0.97} & 35.03\uar{2.73} & 2.50\uar{0.04} & 18.76\uar{1.38} & 16.00\uar{1.06} & 1.78\uar{0.02} & 8.89\uar{0.54} \\
                           & PPL        & 28.73\uar{0.83} & 2.78\uar{0.28} & 15.76\uar{0.56} & 38.82\uar{6.52} & 2.98\uar{0.52} & 20.90\uar{3.52} & 15.39\uar{0.45} & 2.20\uar{0.44} & 8.80\uar{0.45} \\
                           & FastV      & 19.33\dar{8.57} & 1.76\dar{0.74} & 10.55\dar{4.65} & 18.04\dar{14.26}& 1.94\dar{0.52} & 9.99\dar{7.39}  & 12.05\dar{2.89} & 1.86\uar{0.10} & 6.96\dar{1.39} \\
                           & SparseVLM  &  9.10\dar{18.80}& 3.12\uar{0.62} & 6.11\dar{9.09}  & 12.05\dar{20.25}& 2.32\dar{0.14} & 7.19\dar{10.19} & 10.84\dar{4.10} & \textbf{3.96\uar{2.20}} & 7.40\dar{0.95} \\ \midrule
\multirow{4}{*}{Entropy}   & Random     & 32.60\uar{4.70} & 2.14\dar{0.36} & 17.37\uar{2.17} & 38.67\uar{6.37} & 2.10\dar{0.36} & 20.38\uar{3.00} & 18.88\uar{3.94} & 2.18\uar{0.42} & 10.53\uar{2.18}\\
                           & PPL        & 33.97\uar{6.07} & 3.10\uar{0.60} & 18.53\uar{3.33} & 40.18\uar{7.88} & 2.30\dar{0.16} & 21.24\uar{3.86} & 17.21\uar{2.27} & 2.06\uar{0.30} & 9.64\uar{1.29} \\
                           & FastV      & 24.94\dar{2.96} & 2.26\dar{0.24} & 13.60\dar{1.60} & 21.53\dar{10.77}& 1.56\dar{0.90} & 11.55\dar{5.83} & 16.22\uar{1.28} & 1.78\uar{0.02} & 9.00\uar{0.65} \\
                           & SparseVLM  & 10.31\dar{17.59}& 4.56\uar{2.06} & 7.44\dar{7.76}  &  9.40\dar{22.90}& 4.40\uar{1.94} & 6.90\dar{10.48} & 13.95\dar{0.99} & 3.68\uar{1.92} & 8.82\uar{0.47} \\ \midrule
\rowcolor{mycolor} \multicolumn{2}{c|}{\textbf{Q-Tuning (Ours)}} 
  & \cellcolor{mycolor}\textbf{38.21\uar{10.31}} & \cellcolor{mycolor}4.30\uar{1.80} & \cellcolor{mycolor}\textbf{21.26\uar{6.06}} & \cellcolor{mycolor}\textbf{48.07\uar{15.77}} & \cellcolor{mycolor}\textbf{6.14\uar{3.68}} & \cellcolor{mycolor}\textbf{26.57\uar{9.19}} & \cellcolor{mycolor}\textbf{20.47\uar{5.53}} & \cellcolor{mycolor}3.20\uar{1.44} & \cellcolor{mycolor}\textbf{11.84\uar{3.49}} \\  \midrule
\multicolumn{2}{c|}{Full Dataset} 
  & 32.90 & 3.02 & 17.96 & 42.08 & 3.08 & 22.58 & 16.53 & 2.10 & 9.31 \\
\midrule
\end{tabular}
}
\vspace{-10pt}
\end{table*}

\clearpage

\subsection{Additional Results on additional baselines}
\label{appendix:add_baseline}


Tables~\ref{tab:additional_sample_baseline} and~\ref{tab:additional_token_baseline} extend our evaluation to a broader set of pruning strategies under matched sample–token budgets. Table~\ref{tab:additional_sample_baseline} varies sample ratios (12.5\%, 25\%) and token ratios (50\%, 100\%) under a fixed random token pruner, while Table~\ref{tab:additional_token_baseline} fixes the token ratio at 50\% and varies the sample ratio (12.5\%, 25\%, 100\%) under a random sample pruner.




\begin{table*}[htbp]
\centering
\caption{Evaluation of stronger task-relevant baselines under different sample ratios (12.5\%, 25\%) and token ratios (50\%, 100\%) with the same token pruner (\textit{Random}), where \ua~and \da~respectively denote improvements or degradations over the \textit{Random} baseline.}
\vspace{-5pt}
\label{tab:additional_sample_baseline}
\resizebox{0.99\textwidth}{!}{%
\begin{tabular}{c|c|cccccc|cccccc}
\toprule 
\multicolumn{1}{c|}{Sample} &\multicolumn{1}{c|}{Token} & \multicolumn{6}{c|}{\textbf{LLaMA2-7B}} 
& \multicolumn{6}{c}{\textbf{Mistral-7B}} \\
\multicolumn{1}{c|}{Pruner} &\multicolumn{1}{c|}{Ratio} & ARC-E & ARC-C & GSM8K & SQuAD & TriviaQA & Avg.
  & ARC-E & ARC-C & GSM8K & SQuAD & TriviaQA & Avg. \\ 
\midrule

\multicolumn{2}{c|}{Zero-Shot} 
  & 53.44 & 38.98 & 5.31 & 12.18 & 43.00 & 30.58 & 66.67 & 46.10 & 18.35 & 10.01 & 43.77 & 36.98 \\
\midrule

\rowcolor{gray!5}
\multicolumn{14}{c}{\textit{$12.5\%$ Samples}} \\ \midrule
\multirow{1}{*}{Random} & 50\% & 59.25 & 41.02 & 8.11 & 12.75 & 48.75 & 33.98 & 70.55 & 48.14 & 22.74 & 19.57 & 52.63 & 42.73 \\ \midrule
\multirow{2}{*}{Alpagasus} 
& 50\%  
& 58.91\dar{0.34} & 41.69\uar{0.67} & 6.67\dar{1.44} & 12.81\uar{0.06} & 46.28\dar{2.47} & 33.27\dar{0.71}
& 72.84\uar{2.29} & 48.47\uar{0.33} & 20.55\dar{2.19} & 20.41\uar{0.84} & 52.21\dar{0.42} & 42.90\uar{0.17} \\
& 100\%
& 58.91\dar{0.34} & 41.36\uar{0.34} & 7.58\dar{0.53} & 13.45\uar{0.70} & 46.96\dar{1.79} & 33.65\dar{0.33}
& 70.72\uar{0.17} & 48.81\uar{0.67} & 23.35\uar{0.61} & 24.30\uar{4.73} & 52.71\uar{0.08} & 43.98\uar{1.25} \\ \midrule
\multirow{2}{*}{Deita}
& 50\%
& 58.20\dar{1.05} & 42.37\uar{1.35} & 7.43\dar{0.68} & 13.85\uar{1.10} & 45.59\dar{3.16} & 33.49\dar{0.49}
& 72.66\uar{2.11} & \textbf{50.51\uar{2.37}} & 22.97\dar{0.23} & 20.91\uar{1.34} & 54.17\uar{1.54} & 44.25\uar{1.52} \\
& 100\%
& 58.38\dar{0.87} & 42.37\uar{1.35} & 5.46\dar{2.65} & 13.99\uar{1.24} & 45.63\dar{3.12} & 33.17\dar{0.81}
& 72.13\uar{1.58} & 50.17\uar{2.03} & 22.51\dar{0.23} & 24.46\uar{4.89} & 55.28\uar{2.65} & 44.92\uar{2.19} \\ \midrule
\multirow{2}{*}{DS2}
& 50\%
& 58.73\dar{0.52} & 42.37\uar{1.35} & 5.99\dar{2.12} & 15.05\uar{2.30} & 45.51\dar{3.24} & 33.53\dar{0.45}
& 71.25\dar{0.70} & 47.46\dar{0.68} & 21.23\dar{1.51} & 21.57\uar{2.00} & 51.23\dar{1.40} & 42.55\dar{0.18} \\
& 100\%
& 59.08\dar{0.17} & 41.36\uar{0.34} & 6.67\dar{1.44} & 15.14\uar{2.39} & 45.69\dar{3.06} & 33.59\dar{0.39}
& 72.13\uar{1.58} & 49.83\uar{1.69} & 23.35\uar{0.61} & 23.99\uar{4.42} & 52.85\uar{0.22} & 44.43\uar{1.70} \\ \midrule
\multirow{2}{*}{LESS}
& 50\%
& 58.20\dar{1.05} & 42.37\uar{1.35} & 6.67\dar{1.44} & 14.56\uar{1.81} & 46.62\dar{2.13} & 33.69\dar{0.29}
& 73.19\uar{2.64} & 49.49\uar{1.35} & 22.90\dar{0.16} & 19.61\uar{0.04} & 53.14\uar{0.51} & 43.67\uar{0.94} \\
& 100
& 58.38\dar{0.87} & \textbf{43.39\uar{2.37}} & 6.82\dar{1.29} & 16.16\uar{3.41} & 46.72\dar{2.03} & 34.30\uar{0.32}
& \textbf{73.90\uar{3.35}} & 48.47\uar{0.33} & 23.12\uar{0.38} & 21.67\uar{2.10} & 56.00\uar{3.37} & 44.63\uar{1.90} \\ \midrule
\rowcolor{mycolor} \multicolumn{1}{c|}{\textbf{Q-Tuning (Ours)}}
  & \cellcolor{mycolor}50\% & \cellcolor{mycolor}\textbf{64.20\uar{4.95}} & \cellcolor{mycolor}42.03\uar{1.01} & \cellcolor{mycolor}\textbf{10.54\uar{2.43}} & \cellcolor{mycolor}\textbf{18.79\uar{6.04}} 
  & \cellcolor{mycolor}\textbf{53.12\uar{4.37}} & \cellcolor{mycolor}\textbf{37.74\uar{3.76}} 
  & \cellcolor{mycolor}71.60\uar{1.05} & \cellcolor{mycolor}48.14\uar{0.00} & \cellcolor{mycolor}\textbf{29.34\uar{6.60}} 
  & \cellcolor{mycolor}\textbf{27.75\uar{8.18}} & \cellcolor{mycolor}\textbf{57.78\uar{5.15}} 
  & \cellcolor{mycolor}\textbf{46.92\uar{4.19}} \\  
\midrule
\rowcolor{gray!5}
\multicolumn{14}{c}{\textit{$25\%$ Samples}} \\ \midrule
\multirow{1}{*}{Random} & 50\% 
& 60.32 & 41.69 & 5.76 & 13.43 & 48.41 & 33.92
& 70.19 & 46.10 & 20.62 & 24.07 & 53.74 & 42.95 \\ \midrule

\multirow{2}{*}{Alpagasus}
& 50\%
& 59.61\dar{0.71} & 42.71\uar{1.02} & 5.91\uar{0.15} & 14.12\uar{0.69} & 47.30\dar{1.11} & 33.93\uar{0.01}
& 70.37\uar{0.18} & 47.80\uar{1.70} & 20.85\uar{0.23} & 22.38\dar{1.69} & 52.87\dar{0.87} & 42.78\dar{0.17} \\
& 100\%
& 60.49\uar{0.17} & 43.39\uar{1.70} & 7.51\uar{1.75} & 13.16\dar{0.27} & 48.41\uar{0.00} & 34.59\uar{0.67}
& 71.60\uar{1.41} & \textbf{50.17\uar{4.07}} & 23.35\uar{2.73} & 22.38\dar{1.69} & 55.73\uar{1.99} & 44.65\uar{1.70} \\ \midrule
\multirow{2}{*}{Deita}
& 50\%
& 59.08\dar{1.24} & 41.69\uar{0.00} & 7.66\uar{1.90} & 13.23\dar{0.20} & 47.55\dar{0.86} & 33.84\dar{0.08}
& 72.13\uar{1.94} & 47.80\uar{1.70} & 21.00\uar{0.38} & 23.16\dar{0.91} & 53.97\uar{0.23} & 43.61\uar{0.66} \\
& 100\%
& 58.73\dar{1.59} & 42.71\uar{1.02} & 4.92\dar{0.84} & 14.88\uar{1.45} & 48.24\dar{0.17} & 33.90\dar{0.02}
& \textbf{72.66\uar{2.47}} & 47.80\uar{1.70} & 22.29\uar{1.67} & 25.87\uar{1.80} & 54.40\uar{0.66} & 44.60\uar{1.65} \\ \midrule
\multirow{2}{*}{DS2}
& 50\%
& 59.44\dar{0.88} & 42.03\uar{0.34} & 6.44\uar{0.68} & 15.59\uar{2.16} & 47.34\dar{1.07} & 34.17\uar{0.25}
& 70.72\uar{0.53} & 48.47\uar{2.37} & 22.74\uar{2.12} & 21.15\dar{2.92} & 53.49\dar{0.25} & 43.32\uar{0.37} \\
& 100\%
& 59.26\dar{1.06} & 41.36\dar{0.33} & 7.05\uar{1.29} & 15.20\uar{1.77} & 48.76\uar{0.35} & 34.33\uar{0.41}
& 71.43\uar{1.24} & 48.47\uar{2.37} & 22.44\uar{1.82} & 21.72\dar{2.35} & 53.76\uar{0.02} & 43.57\uar{0.62} \\ \midrule
\multirow{2}{*}{LESS}
& 50\%
& 59.44\dar{0.88} & \textbf{43.39\uar{1.70}} & 7.05\uar{1.29} & 15.89\uar{2.46} & 46.60\dar{1.81} & 34.47\uar{0.55}
& 71.78\uar{1.59} & 47.80\uar{1.70} & 22.14\uar{1.52} & 24.51\uar{0.44} & 53.04\dar{0.70} & 43.85\uar{0.90} \\
& 100
& 59.79\dar{0.53} & 41.02\dar{0.67} & 6.37\uar{0.61} & 14.44\uar{1.01} & 48.44\uar{0.03} & 34.01\uar{0.09}
& 71.60\uar{1.41} & 49.49\uar{3.39} & 21.68\uar{1.06} & 30.85\uar{6.78} & 53.40\dar{0.34} & 45.41\uar{2.46} \\ \midrule
\rowcolor{mycolor} \multicolumn{1}{c|}{\textbf{Q-Tuning (Ours)}} & 50\% 
  & \cellcolor{mycolor}\textbf{63.14\uar{2.82}} & \cellcolor{mycolor}42.03\uar{0.34} & \cellcolor{mycolor}\textbf{8.87\uar{3.11}} & \cellcolor{mycolor}\textbf{16.76\uar{3.33}} 
  & \cellcolor{mycolor}\textbf{51.52\uar{3.11}} & \cellcolor{mycolor}\textbf{36.47\uar{2.55}} 
  & \cellcolor{mycolor}71.78\uar{1.59} & \cellcolor{mycolor}47.12\uar{1.02} & \cellcolor{mycolor}\textbf{26.08\uar{5.46}} 
  & \cellcolor{mycolor}\textbf{32.79\uar{8.72}} & \cellcolor{mycolor}\textbf{56.17\uar{2.43}} 
  & \cellcolor{mycolor}\textbf{46.79\uar{3.84}} \\ \midrule
\multicolumn{2}{c|}{Full Dataset} 
  & 61.55 & 42.37 & 8.64 & 13.80 & 50.45 & 35.36 & 71.25 & 45.76 & 26.68 & 31.81 & 53.67 & 45.84 \\
\midrule
\end{tabular}
}
\vspace{-10pt}
\end{table*}

\begin{table*}[htbp]
\centering
\caption{Evaluation of stronger task-relevant baselines under different sample ratios (12.5\%, 25\%, 100\%) and fixed token ratio (50\%) with the same sample pruner (\textit{Random}), where \ua~and \da~respectively denote improvements or degradations over the \textit{Random} baseline.}
\vspace{-5pt}
\label{tab:additional_token_baseline}
\resizebox{0.99\textwidth}{!}{%
\begin{tabular}{c|c|cccccc|cccccc}
\toprule 
\multicolumn{1}{c|}{Token} &\multicolumn{1}{c|}{Sample} & \multicolumn{6}{c|}{\textbf{LLaMA2-7B}} 
& \multicolumn{6}{c}{\textbf{Mistral-7B}} \\
\multicolumn{1}{c|}{Pruner} &\multicolumn{1}{c|}{Ratio} & ARC-E & ARC-C & GSM8K & SQuAD & TriviaQA & Avg.
  & ARC-E & ARC-C & GSM8K & SQuAD & TriviaQA & Avg. \\ 
\midrule

\multicolumn{2}{c|}{Zero-Shot} 
  & 53.44 & 38.98 & 5.31 & 12.18 & 43.00 & 30.58 & 66.67 & 46.10 & 18.35 & 10.01 & 43.77 & 36.98 \\
\midrule

\rowcolor{gray!5}
\multicolumn{14}{c}{\textit{$50\%$ Tokens}} \\ \midrule

\multirow{2}{*}{Random}    
& 12.5\% & 59.25 & 41.02 & 8.11 & 12.75 & 48.75 & 33.98 & 70.55 & 48.14 & 22.74 & 19.57 & 52.63 & 42.73 \\
& 25\% & 60.32\uar{1.07} & 41.69\uar{0.67} & 5.76\dar{2.35} & 13.43\uar{0.68} & 48.41\dar{0.34} & 33.92\dar{0.06} & 70.19\dar{0.36} & 46.10\dar{2.04} & 20.62\dar{2.12} & 24.07\uar{4.50} & 53.74\uar{1.11} & 42.95\uar{0.22} \\ \midrule

\multirow{3}{*}{Rho-1} 
& 12.5\% & 53.97\dar{5.28} & 40.34\dar{0.68} & 6.67\dar{1.44} & 13.27\uar{0.52} & 46.62\dar{2.13} & 32.17\dar{1.81} & 71.60\uar{1.05} & 47.46\dar{0.68} & 20.62\dar{2.12} & 23.74\uar{4.17} & 53.71\uar{1.08} & 43.43\uar{0.70} \\
& 25\% & 59.08\dar{0.17} & 41.02\uar{0.00} & 6.67\dar{1.44} & 13.99\uar{1.24} & 47.19\dar{1.56} & 33.59\dar{0.39} & 70.37\dar{0.18} & 46.78\dar{1.36} & 21.76\dar{0.98} & 20.85\uar{1.28} & 52.22\dar{0.41} & 42.40\dar{0.33} \\
& 100\% & 60.67\uar{1.42} & 41.36\uar{0.34} & 7.28\dar{0.83} & 14.97\uar{2.22} & 49.82\uar{1.07} & 34.82\uar{0.84} & 63.67\dar{6.88} & 43.39\dar{4.75} & 11.68\dar{11.06} & 15.46\dar{4.11} & 31.84\dar{20.79} & 33.21\dar{9.52} \\ \midrule

\multirow{3}{*}{TokenCleaning}     
& 12.5\% & 58.73\dar{0.52} & \textbf{43.73\uar{2.71}} & 7.43\dar{0.68} & 12.84\uar{0.09} & 46.37\dar{2.38} & 33.82\dar{0.16} & 71.60\uar{1.05} & \textbf{49.49\uar{1.35}} & 21.00\dar{1.74} & 22.73\uar{3.16} & 53.03\uar{0.40} & 43.57\uar{0.84} \\
& 25\% & 59.96\uar{0.71} & 42.37\uar{1.35} & 5.99\dar{2.12} & 13.62\uar{0.87} & 46.14\dar{2.61} & 33.62\dar{0.36} & 70.19\dar{0.36} & 47.80\dar{0.34} & 19.94\dar{2.80} & 24.17\uar{4.60} & 52.02\dar{0.61} & 42.82\uar{0.09} \\
& 100\% & 61.20\uar{1.95} & 41.02\uar{0.00} & 7.73\dar{0.38} & 15.41\uar{2.66} & 48.29\dar{0.46} & 34.73\uar{0.75} & \textbf{74.07\uar{3.52}} & 46.44\dar{1.70} & 24.79\uar{2.05} & 33.26\uar{13.69} & 54.32\uar{1.69} & 46.58\uar{3.85} \\ \midrule

\rowcolor{mycolor} 
& \cellcolor{mycolor}12.5\% & \cellcolor{mycolor}\textbf{64.20\uar{4.95}} & \cellcolor{mycolor}42.03\uar{1.01} & \cellcolor{mycolor}\textbf{10.54\uar{2.43}} & \cellcolor{mycolor}\textbf{18.79\uar{6.04}} & \cellcolor{mycolor}\textbf{53.12\uar{4.37}} & \cellcolor{mycolor}\textbf{37.74\uar{3.76}} & \cellcolor{mycolor}71.60\uar{1.05} & \cellcolor{mycolor}48.14\uar{0.00} & \cellcolor{mycolor}\textbf{29.34\uar{6.60}} & \cellcolor{mycolor}27.75\uar{8.18} & \cellcolor{mycolor}\textbf{57.78\uar{5.15}} & \cellcolor{mycolor}\textbf{46.92\uar{4.19}} \\
\rowcolor{mycolor}
\multirow{-2}{*}{\cellcolor{mycolor}\textbf{Q-Tuning (Ours)}}
& \cellcolor{mycolor}25\% & \cellcolor{mycolor}63.14\uar{3.89} & \cellcolor{mycolor}42.03\uar{1.01} & \cellcolor{mycolor}8.87\uar{0.76} & \cellcolor{mycolor}16.76\uar{4.01} & \cellcolor{mycolor}51.52\uar{2.77} & \cellcolor{mycolor}36.47\uar{2.49} & \cellcolor{mycolor}71.78\uar{1.23} & \cellcolor{mycolor}47.12\dar{1.02} & \cellcolor{mycolor}26.08\uar{3.34} & \cellcolor{mycolor}\textbf{32.79\uar{13.22}} & \cellcolor{mycolor}56.17\uar{3.54} & \cellcolor{mycolor}46.79\uar{4.06} \\ \midrule

\multicolumn{2}{c|}{Full Dataset} 
  & 61.55 & 42.37 & 8.64 & 13.80 & 50.45 & 35.36 & 71.25 & 45.76 & 26.68 & 31.81 & 53.67 & 45.84 \\
\midrule
\end{tabular}
}
\vspace{-10pt}
\end{table*}

\clearpage

\subsection{Further Ablation study}

\subsubsection{Ablation with different hyperparameters}
\label{app:hyp}

\begin{figure}[htbp!]
    \centering
    \includegraphics[width=0.9\textwidth]{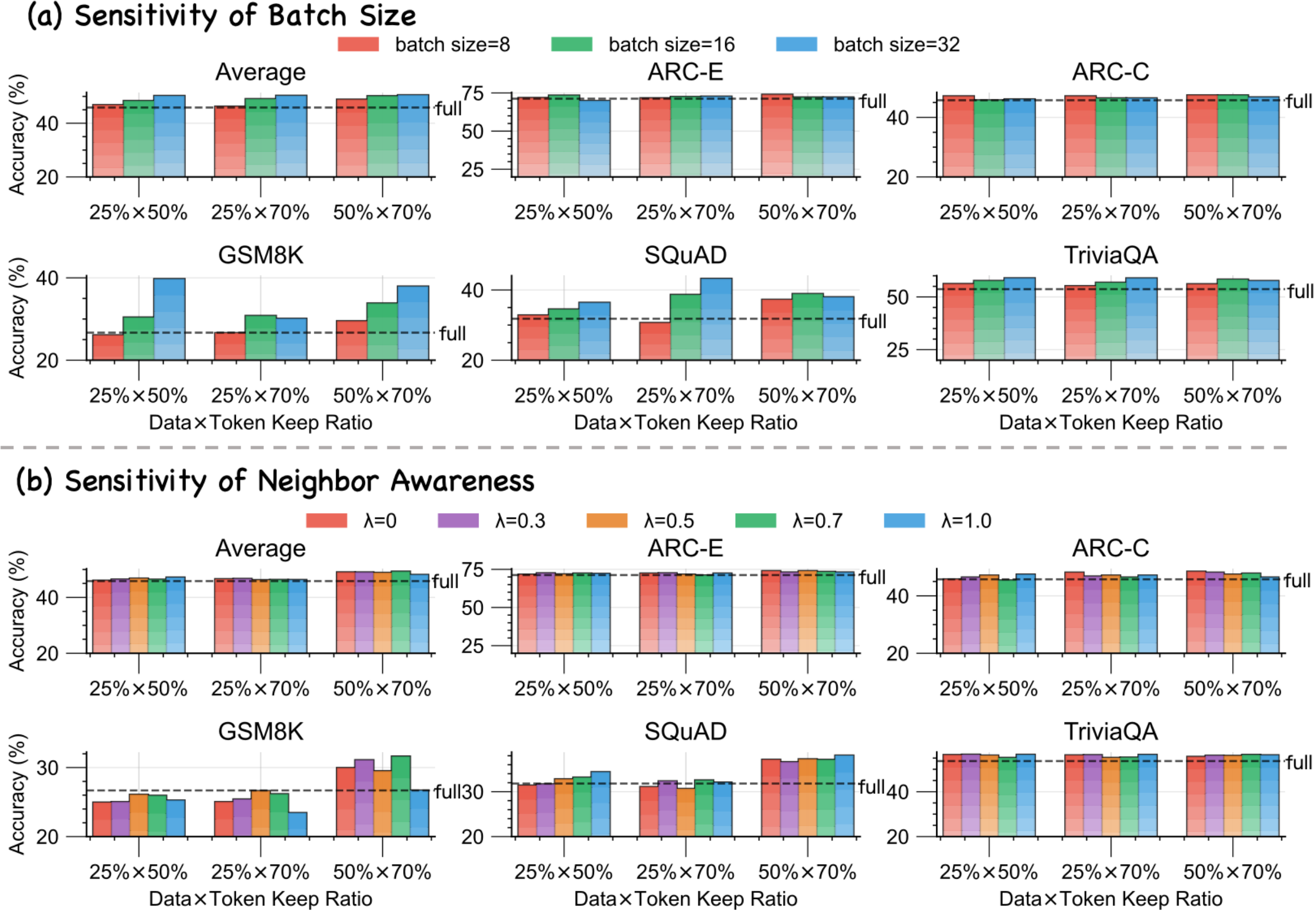} 
    \vspace{-5pt}
    \caption{
    Effect of varying (a) batch size (8, 16, 32) and (b) neighbor awareness $\lambda$ ($0,0.3,0.5,0.7,1.0$) for Mistral-7B under three data--token keep ratio configurations ($25\%\times50\%$, $25\%\times70\%$, $50\%\times70\%$). 
    Dashed lines marked ``full'' denote models trained on the full data without pruning. 
    }
    \vspace{-10pt}
    \label{fig:ablation_supp}
\end{figure}

\paragraph{Sensitivity of batch size in dynamic sample pruning.}
An important question is how batch size influences the performance of Q-Tuning. To study this, we varied the batch size across $\{8,16,32\}$ and evaluated ARC-E, ARC-C, GSM8K, SQuAD, TriviaQA and Average of five benchmarks. As shown in the upper part of Figure~\ref{fig:ablation_supp}, larger batch sizes generally improve performance, especially on GSM8K and SQuAD, where the gains are more pronounced. In contrast, TriviaQA remains relatively stable across all settings, suggesting limited sensitivity to batch size. Overall, these results indicate that Q-Tuning is robust to the choice of batch size, with moderate to larger batches offering additional benefits in certain tasks.

\paragraph{Effectiveness of context awareness $\lambda$.}
\label{ablation:lambda}
We also examined the impact of neighbor awareness, controlled by the coefficient $\lambda$. As shown in the lower part of Figure~\ref{fig:ablation_supp}, moderate values of $\lambda$ improve performance on GSM8K and SQuAD, whereas extreme values yield diminishing or unstable gains. In contrast, TriviaQA shows little sensitivity to $\lambda$, with performance remaining stable across all settings. These results suggest that Q-Tuning benefits from incorporating neighbor awareness, but only up to a moderate level, beyond which gains are marginal.

\clearpage


\subsubsection{Additional ablation across different quadrants}
\label{appendix:add_quadrant}

Table~\ref{tab:different_quadrant_combined} presents additional quadrant-wise ablation results on the Wizard dataset, evaluated on LLaMA2-7B and Mistral-7B, examining how pruning Q1--Q4 at both the sample and token levels impacts downstream performance.

\begin{table*}[htbp]
\centering
\caption{Ablation on four quadrants under different sample and token ratios, where \checkmark indicates that the corresponding quadrant is pruned.}
\vspace{-5pt}
\label{tab:different_quadrant_combined}

\setlength{\tabcolsep}{3.2pt}
\renewcommand{\arraystretch}{0.95}
\scriptsize
\resizebox{\linewidth}{!}{%
\begin{tabular}{@{}cc|cc|cccccc|cccccc@{}}
\toprule
\multicolumn{2}{c|}{Sample pruning} 
& \multicolumn{2}{c|}{Token pruning} 
& \multicolumn{6}{c|}{\textbf{LLaMA2-7B}} 
& \multicolumn{6}{c}{\textbf{Mistral-7B}} \\
Q1 & Q3 & Q2 & Q4
& ARC-E & ARC-C & GSM8K & SQuAD & TriviaQA & Avg.
& ARC-E & ARC-C & GSM8K & SQuAD & TriviaQA & Avg. \\
\midrule

\rowcolor{gray!5}
\multicolumn{16}{c}{\textit{$12.5\%$ Samples, $50\%$ Tokens}} \\
\midrule
 & \checkmark & \checkmark &
& 60.14 & 41.02 & 8.04 & 13.11 & 50.23 & 34.51
& 71.43 & 47.80 & 28.13 & \textbf{31.09} & 54.27 & 46.54 \\

\checkmark & \checkmark & \checkmark & \checkmark
& 64.02 & 41.02 & 9.86 & \textbf{19.26} & 52.00 & 37.23
& 71.60 & \textbf{48.47} & 28.89 & 26.46 & \textbf{57.88} & 46.66 \\

\checkmark & \checkmark & \checkmark &
& \textbf{64.20} & \textbf{42.03} & \textbf{10.54} & 18.79 & \textbf{53.12} & \textbf{37.74}
& \textbf{71.60} & 48.14 & \textbf{29.34} & 27.75 & 57.78 & \textbf{46.92} \\

\midrule
\rowcolor{gray!5}
\multicolumn{16}{c}{\textit{$25\%$ Samples, $50\%$ Tokens}} \\
\midrule
 & \checkmark & \checkmark &
& 59.61 & 41.02 & 7.51 & 13.74 & 50.25 & 34.43
& 71.25 & 46.10 & \textbf{29.04} & 31.55 & 54.84 & 46.56 \\

\checkmark & \checkmark & \checkmark & \checkmark
& 61.02 & 40.68 & \textbf{9.10} & \textbf{18.66} & \textbf{51.68} & 36.23
& 70.72 & 47.12 & 28.58 & 31.85 & \textbf{56.33} & \textbf{46.92} \\

\checkmark & \checkmark & \checkmark &
& \textbf{63.14} & \textbf{42.03} & 8.87 & 16.76 & 51.52 & \textbf{36.47}
& \textbf{71.78} & \textbf{47.12} & 26.08 & \textbf{32.79} & 56.17 & 46.79 \\

\bottomrule
\end{tabular}
}
\vspace{-10pt}
\end{table*}

\clearpage
\section{Pseudocode of the proposed method}
\label{app:pseudo-code}
To facilitate clarity and reproducibility, we summarize the proposed Q-Tuning algorithm in Algorithm~\ref{alg:qtuning}. The procedure unfolds within a single training iteration and consists of two tightly coupled stages: (i) \textit{sample pruning}, where instances are dynamically selected based on their position in the error–uncertainty (EU) plane through an efficient bisection-based search of quantile thresholds, and (ii) \textit{token pruning}, where retained samples undergo finer-grained filtering to preserve only the most informative subset of tokens. 
\begin{algorithm}[htbp!]
\caption{Q-Tuning: dynamic data pruning in one iteration}
\label{alg:qtuning}

\begin{algorithmic}[1]
\STATE \textbf{Input:} Mini-batch $\mathcal{B}_t$, model $f_{\theta_t}$, retention ratios $r_{\textrm{sample}}, r_{\textrm{token}}$, smoothing $\lambda$.
\STATE \textbf{Output:} Pruned mini-batch $\tilde{\mathcal{B}}_t$.

\STATE \textcolor{gray}{\textit{// Stage 1: Sample Pruning via EU Plane}}
\STATE Compute $(\mathrm{PPL}(x,y;f_\theta), \mathrm{Ent}(x,y;f_\theta))$ for each $x \in \mathcal{B}_t$.
\STATE Initialize ranges $\alpha_{\text{low}}=0$, $\alpha_{\text{high}}=0.49$, 
       $\beta_{\text{low}}=0$, $\beta_{\text{high}}=0.49$.

\FOR[Bisection iterations on both axes]{$k=1$ to $K_{\max}$} 
    \STATE $\alpha \gets (\alpha_{\text{low}}+\alpha_{\text{high}})/2$,
           $\beta \gets (\beta_{\text{low}}+\beta_{\text{high}})/2$.
    \STATE Derive thresholds 
           $ppl_{\text{hi}}=Q_{1-\alpha}(\mathrm{PPL}),\; ppl_{\text{lo}}=Q_{\alpha}(\mathrm{PPL})$,
           $ent_{\text{lo}}=Q_{\beta}(\mathrm{Ent}),\; ent_{\text{hi}}=Q_{1-\beta}(\mathrm{Ent})$.
    \STATE Partition samples into quadrants $Q_1$--$Q_4$.
    \STATE $r \gets \frac{|Q_2|+|Q_4|}{|\mathcal{B}_t|}$.
    \IF{$r < r_{\textrm{sample}}$}
        \STATE $\alpha_{\text{low}} \gets \alpha,\;\beta_{\text{low}} \gets \beta$ \COMMENT{Too few kept, relax thresholds}
    \ELSE
        \STATE $\alpha_{\text{high}} \gets \alpha,\;\beta_{\text{high}} \gets \beta$ \COMMENT{Too many kept, tighten thresholds}
    \ENDIF
\ENDFOR

\STATE $\mathcal{B}'_t \gets Q_2 \cup Q_4$.
\IF[Ensure target sample ratio]{$|\mathcal{B}'_t| < r_{\textrm{sample}} \cdot |\mathcal{B}_t|$}
    \STATE Compute \textbf{supp-score} for each $x \in Q_1 \cup Q_3$:  
          $\mathrm{supp}(x) = \max\{\hat{\mathrm{PPL}}(x,y;f_\theta) - \hat{\mathrm{Ent}}(x,y;f_\theta),\; 
          \hat{\mathrm{Ent}}(x,y;f_\theta) - \hat{\mathrm{PPL}}(x,y;f_\theta)\}$,
          where $\hat{\cdot}$ denotes min–max normalization.
    \STATE Select top-scoring samples to augment $\mathcal{B}'_t$ until target size is met.
\ENDIF

\STATE \textcolor{gray}{\textit{// Stage 2: Token Pruning}}
\STATE $\tilde{\mathcal{B}}_t \gets \emptyset$.
\FOR{each $x \in \mathcal{B}'_t$}
    \IF{$x \in Q_4$}
        \STATE Keep all tokens of $x$. \COMMENT{Calibration samples intact}
    \ELSIF{$x \in Q_2$}
        \STATE FOR each token $i \in T(x)$, compute
        \[
        s_i(x,y;f_\theta) = (1-\lambda)\,\mathrm{PPL}_i(x,y;f_\theta)
        + \lambda\,[\mathrm{PPL}_{i-1}(x,y;f_\theta) + \mathrm{PPL}_{i+1}(x,y;f_\theta)].
        \]
        \STATE Keep top-$r_{\textrm{token}}$ fraction of tokens ranked by $s_i$.
    \ENDIF
    \STATE $\tilde{\mathcal{B}}_t \gets \tilde{\mathcal{B}}_t \cup \{m(x) \odot x\}$.
\ENDFOR

\STATE \textbf{return} $\tilde{\mathcal{B}}_t$.
\end{algorithmic}
\end{algorithm}

\clearpage

\section{Further Case Study}
\label{app:semantic_case_study}

\subsection{Example Samples from Each Quadrant}
\label{appendix:example_quadrant}

We report representative samples from each quadrant (Q1--Q4) at the beginning of training, including their perplexity (PPL), entropy, and the corresponding instruction–output pairs.

\renewcommand{\arraystretch}{1.2}
\setlength{\tabcolsep}{6pt}

\begin{longtable}{p{0.1\textwidth} p{0.1\textwidth} p{0.75\textwidth}}
\toprule
PPL & Entropy & Example (Instruction / Output) \\
\midrule
\endfirsthead
\toprule
PPL & Entropy & Example (Instruction / Output) \\
\midrule
\endhead
\bottomrule
\endlastfoot

\multicolumn{3}{l}{\textbf{Q1 Examples}} \\
8.33 & 2.12 & \texttt{Instruction: What are the three primary colors? \ Output: The three primary colors are red, blue, and yellow.} \\
8.04 & 1.73 & \texttt{Instruction: How can we reduce air pollution? \ Output: There are a number of ways to reduce air pollution, such as shifting to renewable energy sources...} \\
91.24 & 1.91 & \texttt{Instruction: Explain why the fraction 4/16 is equivalent to 1/4. \ Output: The fraction 4/16 is equivalent to 1/4 because both numerator and denominator are divisible by 4.} \\
\midrule

\multicolumn{3}{l}{\textbf{Q2 Examples}} \\
8.65 & 1.54 & \texttt{Instruction: Write a short story in third person narration about a protagonist making an important career decision. \ Output: John was at a crossroads in his life. He had just graduate...} \\
11.02 & 1.48 & \texttt{Instruction: Write a short story in third person narration about a protagonist making an important career decision. \ Output: John was at a crossroads in his life. He had just graduate...} \\
15.74 & 1.50 & \texttt{Instruction: Identify the odd one out.\textbackslash nTwitter, Instagram, Telegram \ Output: Telegram} \\
\midrule

\multicolumn{3}{l}{\textbf{Q3 Examples}} \\
4.80 & 1.45 & \texttt{Instruction: Describe the structure of an atom. \ Output: An atom is made up of a nucleus containing protons and neutrons, surrounded by electrons...} \\
2.41 & 1.09 & \texttt{Instruction: How can we reduce air pollution? \ Output: There are a number of ways to reduce air pollution, such as shifting to renewable energy sources...} \\
4.73 & 1.49 & \texttt{Instruction: Identify the odd one out (Twitter, Instagram, Telegram). \ Output: Telegram} \\
\midrule

\multicolumn{3}{l}{\textbf{Q4 Examples}} \\
6.03 & 1.83 & \texttt{Instruction: Give three tips for staying healthy. \ Output: 1. Eat a balanced diet... 2. Exercise regularly...} \\
8.04 & 1.71 & \texttt{Instruction: What are the three primary colors? \ Output: The three primary colors are red, blue, and yellow.} \\
5.37 & 1.79 & \texttt{Instruction: Write a short story in third person narration about a protagonist making an important career decision. \ Output: John was at a crossroads in his life. He had just graduate...} \\
\end{longtable}

\clearpage

\subsection{Pruned Case}
\label{sec:pruned-case-study}
We present representative token-pruning cases from \textsc{MathInstruct} using the \textsc{Mistral-7B} model. For each example, we show the model's original \texttt{output} and highlight in \textcolor{red}{red} the fine-grained segments that are \emph{absent} in the corresponding \texttt{pruned\_output}. This visualization makes the pruning behavior transparent at the character/substring level and helps diagnose which semantic components are preferentially removed (e.g., boilerplate, redundant explanations, or unstable numeric derivations).

Unless otherwise stated, we keep the original formatting and line breaks of the model outputs to preserve exact alignment. The highlighting is produced by an LCS-based diff between \texttt{output} and \texttt{pruned\_output}, ensuring that only truly deleted spans are marked while retained text remains uncolored. We do \emph{not} introduce any additional edits beyond necessary \LaTeX\ escaping.

These qualitative cases complement the aggregate statistics reported in the main text by revealing typical pruning patterns on mathematical reasoning tasks (definitions and restatements vs.\ computational steps), and by surfacing edge cases where pruning removes numerically relevant content. Together, they provide an interpretable view of how token-level pruning affects the fidelity and brevity of generated solutions.
\begin{tcolorbox}[
  breakable,
  colback=gray!3,      
  colframe=gray!55,    
  boxrule=0.6pt,       
  arc=2pt,             
  left=6pt,right=6pt,top=6pt,bottom=6pt,
  boxsep=2pt
]
\subsubsection*{Sample 0}
\paragraph{Instruction} sandy is younger than molly by 20 years . if the ratio of their ages is 7 : 9 , how old is sandy ? Let's write a program.

\paragraph{Output}
\begin{flushleft}\ttfamily
n0 = 20.0\\
n1 \textcolor{red}{= }7.\textcolor{red}{0}\\
n\textcolor{red}{2 = 9.0}\\
t0 = n2 - n\textcolor{red}{1}\textcolor{red}{\texttt{\textbackslash n}}
\\
\textcolor{red}{t1 =} n1 / t0\\
answer =\textcolor{red}{ n}0 * t\textcolor{red}{1}\\
print(answer\textcolor{red}{)}
\normalfont\end{flushleft}
\end{tcolorbox}
\begin{tcolorbox}[
  breakable,
  colback=gray!3,      
  colframe=gray!55,    
  boxrule=0.6pt,       
  arc=2pt,             
  left=6pt,right=6pt,top=6pt,bottom=6pt,
  boxsep=2pt
]
\subsubsection*{Sample 1}
\paragraph{Instruction} A manufacturing process produces 5000 units of a particular product. The probability of a certain type of defect occurring in a unit is 0.02. What is the probability that exactly 100 units will have this defect?
\paragraph{Output (deletions highlighted)}
\begin{flushleft}\ttfamily
To solve this problem, we can use the binomial probability formula, which is:\\

P(X = k) = (nCk) * (p\^{}k) * (1-p)\^{}(n\textcolor{red}{-}k)\\

where:\\
- P(X = k) is the probability of exactly k successes (defective units in this case) in n trials (total units produced)\\
- nCk (read as "n \textcolor{red}{choose }k") is the number of combinations of choosing k successes from n trials, calculated as n! / (k! * \textcolor{red}{(}n-k\textcolor{red}{)}!)\\
- p is \textcolor{red}{the probability of }success (defect occurring) in a single trial\\
- 1-p \textcolor{red}{is the probability of failure (no defect occurring) in a single trial}\\
- n is the total number \textcolor{red}{of }trials (units produced)\\
- \textcolor{red}{k is the number of successes }(defective units) we want to find the probability for\\

In this problem, we have:\\
- n =\textcolor{red}{ 500}0 (total units produced\textcolor{red}{)}\\
- \textcolor{red}{k = 100 (defective units)}\\
- \textcolor{red}{p = 0.02 (probability of a defect occurring in a unit)}\\

Now, we can plug these values into the binomial probability formula:\\

P\textcolor{red}{(X}=1\textcolor{red}{00)=}\textcolor{red}{(}5\textcolor{red}{000}C\textcolor{red}{100}\textcolor{red}{)}\(\cdot\)
(0.02)\(^{100}\)\(\cdot\) (0.\textcolor{red}{9}8)\(^{49\textcolor{red}{00}}\)\textcolor{red}{)}

First, let's calculate 50\textcolor{red}{00C100}:\\

500\textcolor{red}{0}C1\textcolor{red}{00} = 5\textcolor{red}{000!} / \textcolor{red}{(100! }* 4\textcolor{red}{900}!)\\
= 2.031759e+134 / (9.3\textcolor{red}{32}{622e+157 * 1.929713e\textcolor{red}{+}126)}\\
\textcolor{red}{= }2.0317\textcolor{red}{59}e+134 / (1.801713e+284)\\
= 1.\textcolor{red}{1}27271e-150\\

Now, let's calculate \textcolor{red}{0.02\^{}1}00 and \textcolor{red}{0.98\^{}4900}:\\

\textcolor{red}{0.02\^{}100 }= 1.267651e-70\\
\textcolor{red}{0.98\^{}4900 = }2.372472e-44\\

Finally, let's plug these values back into the formula:\\

\textcolor{red}{P(X = 100) = 1}\textcolor{red}{.127271e-150 }* 1\textcolor{red}{.267651e-70 * 2.372472e-44}\\
\textcolor{red}{= 3.394e-264}\\

The probability that exactly \textcolor{red}{100 }units will have this defect is approximately \textcolor{red}{3.394e-\textcolor{red}{26}4}, which is an extremely small probability.
\normalfont\end{flushleft}

\end{tcolorbox}

\clearpage

\end{document}